%% file: Inference.tex
\documentclass[11pt,oneside,open=any,numbers=noenddot]{scrbook}

\usepackage{amsmath}
\usepackage{fontspec}
\usepackage{unicode-math}
\usepackage[paperwidth=7in,paperheight=10in,left=1.2in,right=1.2in,top=0.95in,bottom=1.05in,headsep=16pt,footskip=30pt,heightrounded]{geometry}
\usepackage{microtype}
\usepackage{needspace}

\usepackage{xcolor}
\definecolor{ink}{HTML}{1f2937}\definecolor{accent}{HTML}{601878}
\definecolor{accentlight}{HTML}{9048A8}
\definecolor{rust}{HTML}{b91c1c}\definecolor{coderule}{HTML}{d8dce2}\definecolor{codebg}{HTML}{f6f8fa}
\definecolor{grey}{HTML}{6B7280}\definecolor{rule}{HTML}{D1D5DB}\definecolor{cnum}{HTML}{9CA3AF}
\usepackage{colortbl} 
\definecolor{accentdark}{HTML}{2F0047}
\definecolor{tablehead}{HTML}{E7F3F1}
\definecolor{tablerow}{HTML}{F3F7FA}
\definecolor{rowgood}{HTML}{F1F8F4}\definecolor{rowgoodalt}{HTML}{F7FCF9}
\definecolor{rowwarn}{HTML}{FFF8E8}\definecolor{rowwarnalt}{HTML}{FFFCF2}
\definecolor{rowbad}{HTML}{FFF1F2}\definecolor{rowbadalt}{HTML}{FFF7F8}
\definecolor{sigrow}{HTML}{FDECEC}\definecolor{sigrowalt}{HTML}{FCF3F3}
\definecolor{fixrow}{HTML}{EAF6EE}\definecolor{fixrowalt}{HTML}{F2FAF5}
\definecolor{rowinfo}{HTML}{EEF5FF}\definecolor{rowinfoalt}{HTML}{F6FAFF}
\definecolor{notebg}{HTML}{F6F8FA}
\arrayrulecolor{rule}

\usepackage{seqsplit}
\newcommand{\code}[1]{{\ttfamily\fontsize{8.8}{10.4}\selectfont\color{accentdark}#1}}

\newcommand{\TableCaption}[1]{\par\vspace{3pt}{\usekomafont{caption}#1}\par\vspace{1pt}}

\usepackage{tabularx}
\newcolumntype{L}{>{\RaggedRight\sffamily\footnotesize}X}

\usepackage{booktabs}\usepackage{longtable}\usepackage{array}\usepackage{multirow}\usepackage{calc}\usepackage{etoolbox}\usepackage{ragged2e}\usepackage{enumitem}
\AtBeginEnvironment{longtable}{\footnotesize\sffamily}
\counterwithin{table}{chapter}
\makeatletter
\newcommand{\numcap}[1]{\protected@edef\@currentlabel{\thetable}\TableCaption{\textbf{Table~\thetable.}~#1}}
\makeatother
\newcommand{\numcaptab}[1]{\refstepcounter{table}\TableCaption{\textbf{Table~\thetable.}~#1}}
\newenvironment{notationtable}{%
  \par\medskip\begingroup
  \linespread{1}\selectfont
  \renewcommand{\arraystretch}{1.22}%
  \noindent\begin{tabular}{@{}%
    >{\RaggedRight\arraybackslash\fontsize{9.5}{12}\selectfont}p{1.55in}%
    @{\hspace{14pt}}%
    >{\RaggedRight\arraybackslash\sffamily\fontsize{9}{12}\selectfont}p{2.85in}@{}}
  \arrayrulecolor{accent}\toprule
  {\sffamily\bfseries\footnotesize\color{ink}Symbol} & {\sffamily\bfseries\footnotesize\color{ink}Meaning}\\
  \arrayrulecolor{rule}\midrule
}{%
  \arrayrulecolor{accent}\bottomrule
  \end{tabular}\endgroup\medskip\par
}

\usepackage{graphicx}\usepackage{float}
\graphicspath{{assembled_book/figures/}}
\setkeys{Gin}{width=0.86\linewidth,keepaspectratio}
\usepackage[font={small,color=ink},labelfont={bf,color=accent},labelsep=period,justification=raggedright,singlelinecheck=false]{caption}

\usepackage{fvextra}
\RecustomVerbatimEnvironment{verbatim}{Verbatim}{fontsize=\footnotesize,breaklines=true,breakanywhere=true,breakafter=/-_.frame=single,framesep=6pt,rulecolor=\color{coderule},xleftmargin=10pt,xrightmargin=12pt,breaksymbolleft={}}

\usepackage[most]{tcolorbox}
\tcbuselibrary{skins,breakable,minted}
\usepackage{minted}

\definecolor{codetitlebg}{HTML}{EEF2F7}
\definecolor{codefg}{HTML}{111827}
\definecolor{configbg}{HTML}{FBFBF7}
\definecolor{configrule}{HTML}{DDD6C5}
\definecolor{configtitlebg}{HTML}{F1EDE2}
\definecolor{logbg}{HTML}{111827}
\definecolor{logrule}{HTML}{374151}
\definecolor{logfg}{HTML}{E5E7EB}
\definecolor{logtitlebg}{HTML}{1F2937}

\setminted{fontsize=\footnotesize,breaklines=true,breakanywhere=true,autogobble=true,tabsize=4,obeytabs=true,encoding=utf8}

\newtcblisting{bookcode}[2][]{%
 enhanced,breakable,listing only,listing engine=minted,minted language=#2,
 minted options={fontsize=\footnotesize,breaklines,breakanywhere,autogobble,tabsize=4},
 colback=codebg,colframe=coderule,coltext=codefg,
 boxrule=0.45pt,arc=2pt,left=7pt,right=7pt,top=6pt,bottom=6pt,
 before skip=0.85\baselineskip,after skip=0.85\baselineskip,
 title={#1},coltitle=codefg,colbacktitle=codetitlebg,
 fonttitle=\sffamily\bfseries\footnotesize,
 attach boxed title to top left={xshift=7pt,yshift=-2pt},
 boxed title style={boxrule=0pt,arc=2pt,left=4pt,right=4pt,top=2pt,bottom=2pt}}

\newtcblisting{bookconfig}[2][]{%
 enhanced,breakable,listing only,listing engine=minted,minted language=#2,
 minted options={fontsize=\footnotesize,breaklines,breakanywhere,autogobble,tabsize=2},
 colback=configbg,colframe=configrule,coltext=codefg,
 boxrule=0.45pt,arc=2pt,left=7pt,right=7pt,top=6pt,bottom=6pt,
 before skip=0.85\baselineskip,after skip=0.85\baselineskip,
 title={#1},coltitle=codefg,colbacktitle=configtitlebg,
 fonttitle=\sffamily\bfseries\footnotesize,
 attach boxed title to top left={xshift=7pt,yshift=-2pt},
 boxed title style={boxrule=0pt,arc=2pt,left=4pt,right=4pt,top=2pt,bottom=2pt}}

\newtcblisting{booklog}[1][]{%
 enhanced,breakable,listing only,listing engine=minted,minted language=text,
 minted options={fontsize=\scriptsize,breaklines,breakanywhere,autogobble,tabsize=2},
 colback=logbg,colframe=logrule,coltext=logfg,
 boxrule=0.45pt,arc=2pt,left=7pt,right=7pt,top=6pt,bottom=6pt,
 before skip=0.85\baselineskip,after skip=0.85\baselineskip,
 title={#1},coltitle=logfg,colbacktitle=logtitlebg,
 fonttitle=\sffamily\bfseries\footnotesize,
 attach boxed title to top left={xshift=7pt,yshift=-2pt},
 boxed title style={boxrule=0pt,arc=2pt,left=4pt,right=4pt,top=2pt,bottom=2pt}}

\usepackage{mdframed}
\newtcolorbox{takeawaybox}{breakable,enhanced,colback=accent!5,colframe=accent!45,boxrule=0pt,leftrule=2.5pt,arc=2pt,left=11pt,right=11pt,top=8pt,bottom=8pt,fonttitle=\sffamily\bfseries\footnotesize\color{accent},title=TAKEAWAY}
\newtcolorbox{failurebox}{breakable,enhanced,colback=rust!3,colframe=rust!50!black,boxrule=0.5pt,arc=2pt,left=9pt,right=9pt,top=6pt,bottom=6pt,fonttitle=\sffamily\bfseries\small,title=Failure signature}
\newtcolorbox{tiercard}[1]{enhanced,colback=accent!4,colframe=accent!55!black,boxrule=0.6pt,arc=2pt,left=6pt,right=6pt,top=4pt,bottom=4pt,fonttitle=\sffamily\bfseries\small,coltitle=ink,colbacktitle=accent!14,title=#1}
\newmdenv[backgroundcolor=codebg,linecolor=coderule,linewidth=0.5pt,roundcorner=2pt,innerleftmargin=7pt,innerrightmargin=7pt,innertopmargin=5pt,innerbottommargin=5pt,skipabove=8pt,skipbelow=8pt]{codebox}

\usepackage{pdfpages}
\usepackage{hyperref}
\hypersetup{
  colorlinks=true,
  linkcolor=accent,
  citecolor=accent,
  urlcolor=accent,
  filecolor=accent,
  linktoc=all,
  pdftitle={Inference for Reasoning Models: Practical Lessons from Evaluation and Serving},
  pdfauthor={Luv Verma, PhD},
  pdfsubject={First edition, version 1.0, August 2026. DOI: 10.5281/zenodo.21754285. ORCID: 0000-0002-3835-7836}
}

\usepackage[automark,headsepline]{scrlayer-scrpage}
\clearpairofpagestyles\automark[chapter]{part}
\ihead{\headmark}\ohead*{\pagemark}
\setkomafont{pagehead}{\normalfont\sffamily\footnotesize\color{grey}}
\setkomafont{pagenumber}{\normalfont\sffamily\footnotesize\color{grey}}

\addtokomafont{disposition}{\sffamily}
\setkomafont{chapter}{\sffamily\bfseries\fontsize{25}{29}\selectfont\color{ink}}
\setkomafont{section}{\sffamily\bfseries\large\color{ink}}
\setkomafont{subsection}{\sffamily\bfseries\normalsize\color{ink}}
\KOMAoptions{chapterprefix=true}
\setkomafont{chapterprefix}{\normalfont\sffamily\bfseries\fontsize{11}{13}\selectfont\color{cnum}}

\setkomafont{partentry}{\normalfont\sffamily\bfseries\fontsize{10.2}{12.6}\selectfont\color{accent}}
\setkomafont{partentrypagenumber}{\normalfont\sffamily\bfseries\fontsize{9.4}{12.6}\selectfont\color{accent}}
\setkomafont{chapterentry}{\normalfont\sffamily\fontsize{9.35}{11.7}\selectfont\color{ink}}
\setkomafont{chapterentrypagenumber}{\normalfont\sffamily\fontsize{8.9}{11.7}\selectfont\color{grey}}
\setkomafont{chapterentrydots}{\color{rule}}
\KOMAoptions{chapterentrydots=true}
\DeclareTOCStyleEntry[
  beforeskip=.42em plus .08em,
  indent=0pt,
  numwidth=2.65em,
  rightindent=2.35em,
  pagenumberwidth=2.05em,
  raggedentrytext=true
]{chapter}{chapter}
\DeclareTOCStyleEntry[
  beforeskip=1.15em plus .12em,
  indent=0pt,
  numwidth=2.65em,
  rightindent=2.35em,
  pagenumberwidth=2.05em,
  raggedentrytext=true
]{part}{part}

\newcounter{bookmap}

\newenvironment{partopener}{\begin{quote}\small\sffamily}{\end{quote}}

\newcommand{\nextlesson}[1]{\par\vspace{0.7em}\noindent{\small\sffamily\color{ink!72}\clubpenalty10000\widowpenalty10000\interlinepenalty9999\textit{Next: #1}}\par}

\newtcolorbox{findingbox}{breakable,enhanced,colback=accent!5,colframe=accent!45,boxrule=0pt,leftrule=2.5pt,arc=2pt,left=11pt,right=11pt,top=8pt,bottom=8pt,fonttitle=\sffamily\bfseries\footnotesize\color{accent},title=Chapter finding}
\newcommand{\measuredq}[1]{\par\noindent{\sffamily\bfseries\footnotesize\color{accent}Measured question.}\enspace{\sffamily\small\color{ink}#1}\par\vspace{0.35em}}
\newcommand{\TableFontNine}{\sffamily\fontsize{9}{11.6}\selectfont\RaggedRight}
\newcommand{\TableFontEightSix}{\sffamily\fontsize{8.6}{10.8}\selectfont\RaggedRight}
\newtcolorbox{navbox}{enhanced,colback=notebg,colframe=rule,boxrule=0.5pt,arc=2pt,left=11pt,right=11pt,top=9pt,bottom=9pt}
\newenvironment{readerlist}
  {\begin{itemize}[leftmargin=1.25em,itemsep=2.5pt,parsep=0pt,partopsep=0pt,topsep=3pt,label=\textcolor{accent}{\small\textbullet}]}
  {\end{itemize}}

\title{Inference for Reasoning Models: Practical Lessons from Evaluation and Serving}\author{Luv Verma, PhD}

\newcommand{\thd}[1]{{\sffamily\bfseries\footnotesize #1}}

\newcommand{\TableTitle}[1]{%
  \par\addvspace{0.7em}\needspace{5\baselineskip}%
  {\sffamily\bfseries\fontsize{11.5}{14}\selectfont\color{ink}#1}\par\vspace{0.15em}}

\newcommand{\TableDeck}[1]{%
  {\sffamily\fontsize{9}{12}\selectfont\color{grey}#1}\par\vspace{0.45em}}

\newcommand{\TableNote}[1]{%
  \par\vspace{0.3em}{\sffamily\fontsize{8}{10.5}\selectfont\color{grey}%
  \textit{Notes:} #1}\par\addvspace{0.3em}}

\newcolumntype{R}{>{\RaggedRight\arraybackslash}X}        
\newcolumntype{d}{>{\raggedleft\arraybackslash}p{0.62in}} 

\usepackage[
  backend=biber,
  citestyle=authoryear,
  bibstyle=authoryear,
  maxcitenames=2,
  maxbibnames=99,
  giveninits=true,
  uniquename=false,
  doi=true,
  url=true,
  isbn=true,
  date=year
]{biblatex}
\DeclareDelimFormat{nameyeardelim}{\addcomma\space}

\usepackage{siunitx}
\newcommand{\thu}[1]{{\sffamily\fontsize{7.2}{8.6}\selectfont\color{grey}#1}}

\newcolumntype{Q}[1]{>{\RaggedRight\arraybackslash}p{#1}}

\newcommand{\TableRowAir}{\renewcommand{\arraystretch}{1.18}}

\newcolumntype{C}{>{\centering\arraybackslash}X}

\newcommand{\widefig}[2][5.6in]{%
  \noindent\makebox[\linewidth][c]{\includegraphics[width=#1]{#2}}}

\newcommand{\zrow}{\rowcolor{tablerow}}

\newcommand{\pmv}[2]{#1\,\textpm\,#2}

\renewtcolorbox{findingbox}{enhanced,colback=accent!5,colframe=accent!45,%
  boxrule=0pt,leftrule=2.5pt,arc=2pt,left=11pt,right=11pt,top=8pt,bottom=8pt,%
  fonttitle=\sffamily\bfseries\footnotesize\color{accent},title=Chapter finding}

\newenvironment{statictable}
  {\par\addvspace{\intextsep}\noindent\begin{minipage}{\linewidth}\centering}
  {\end{minipage}\par\addvspace{\intextsep}}

\newenvironment{staticfigure}
  {\par\addvspace{\intextsep}\noindent\begin{minipage}{\linewidth}\centering}
  {\end{minipage}\par\addvspace{\intextsep}}

\newcommand{\StaticCaption}[1]{\par\vspace{4pt}{\usekomafont{caption}#1}\par}

\begin{document}
\frontmatter

\includepdf[pages=1,fitpaper=true]{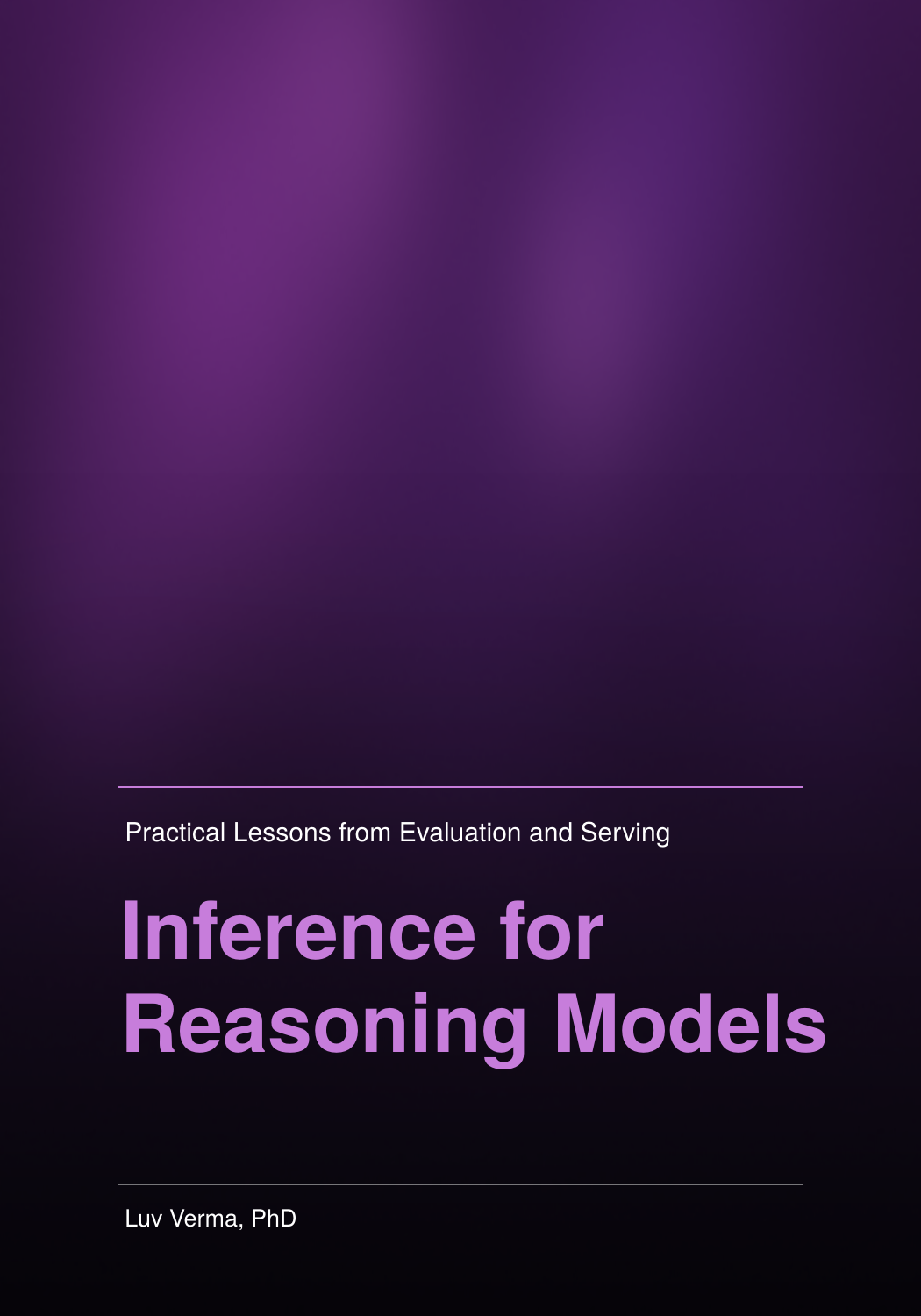}

\thispagestyle{empty}\vspace*{\fill}\noindent
{\small\sffamily\RaggedRight
\textcopyright\ 2026 Luv Verma. First edition, version 1.0, August 2026.\\[6pt]
This work is licensed under the
\href{https://creativecommons.org/licenses/by/4.0/}{Creative Commons
Attribution 4.0 International License (CC\,BY\,4.0)}. It may be shared and
adapted for any purpose, provided appropriate credit is given, the license is
linked, and changes are indicated.\\[6pt]
\href{https://orcid.org/0000-0002-3835-7836}{ORCID: 0000-0002-3835-7836}.\\[6pt]
Set in Source Serif Pro, TeX Gyre Heros, and TeX Gyre Cursor. Typeset with
Xe\LaTeX{}.\\[10pt]
{\color{grey}Also by Luv Verma: \textit{What Broke \& Why: Practical Lessons
from Reinforcement Learning Post-Training.} Zenodo, 2026.
DOI: 10.5281/zenodo.21115798.}\par}
\vspace*{0.06\textheight}\clearpage


\begingroup
\hypersetup{linktoc=all,colorlinks=true,linkcolor=.}
\tableofcontents
\clearpage
\endgroup
\hypersetup{linktoc=all,colorlinks=true,linkcolor=accent}

\addchap{Preface}
\markboth{Preface}{Preface}

An inference result can misrepresent a model even when its weights never change.

One experiment in this book used the same checkpoint, mathematics benchmark,
prompts, decoding mode, and scoring rule in two runs. With an eight-token generation
limit, strict accuracy was 0.08 percent. With a 1{,}024-token limit, it was 93.18
percent. The model had not improved between the runs. The first run simply stopped
almost every response before it reached an answer.

The size of that change is unusual; the cause is not. A reported score comes from a
complete inference pipeline. The prompt template determines what the model sees. The
generation limit determines whether a response can finish. The extractor selects the
answer, and the scoring path evaluates it. Kernels, attention backends, batch
composition, cache reuse, and precision determine how the token sequence is computed.
The scheduler, workload, and memory policy determine the latency, capacity,
throughput, and cost attached to the result.

Inference is not a transparent channel between a checkpoint and a number. It is part
of the measurement.

\Needspace{9\baselineskip}
The book develops five questions in sequence:

\begin{enumerate}[leftmargin=1.6em,itemsep=1pt,parsep=0pt,partopsep=0pt,topsep=2pt]
\item Does the score reflect model capability or its measurement path?
\item Does changed output also change the task result?
\item What sets inference performance and cost?
\item What can controls, context limits, and checkpoint comparisons show?
\item What record supports audit and reconstruction?
\end{enumerate}

\section*{Scope}

Every measurement belongs to a specific checkpoint, workload, software version,
precision mode, hardware configuration, and traffic shape. The printed values are not
constants of all models or future engines. The primary focus is open-weight inference;
the method can apply more broadly, but the claims remain attached to the tested systems.
Evidence and Scope explains the reporting conventions used throughout the book.

\section*{Intended readers}

The book is for researchers, engineers, and technical leaders who evaluate, compare,
deploy, or review language models and their capability or cost claims.

No prior volume is required. Familiarity with language-model generation and serving
will help, but each chapter introduces the distinction it uses. Parts~I--V develop the
argument; Part~VI and the appendices form the reporting and audit reference.

\section*{The central proposition}

An inference result is best read as a sentence, not a scalar:

\begin{quote}
\itshape This checkpoint, under this measurement path, execution path, and workload,
produced this output, score, latency, capacity, or cost.
\end{quote}

\clearpage

\input{why_this_book_the_gap.tex}

\addchap{How to Read This Book}
\markboth{How to Read This Book}{How to Read This Book}

The book supports two reading modes.

The first is a linear argument. Parts~I through~V follow an inference claim from
measurement validity to execution behavior, operation, controls, and architecture. Later
chapters assume distinctions established earlier: a score is separated from its extractor;
text divergence is separated from task impact; a measured quantity is separated from a
derived bound; an operational checkpoint comparison is separated from a causal
architecture claim. Reading the first five parts in order therefore gives the clearest
path through the argument.

The second mode is reference use. Part~VI and the appendices are designed for return
visits. They provide the run-record fields, evaluation preflight, capacity and cost
calculations, evidence-status framework, checkpoint and workload registries, supplemental
results, and operational failure classification used throughout the book.

The best entry point depends on the question.

A score that is unexpectedly low, has moved between harnesses, or disagrees with another
report begins in \textbf{Part~I: Evaluation Validity}. The relevant checks are generation
completion, extraction, prompt and scoring path, decoding, and task coverage.

Identical prompts that produce different text under a changed batch, backend, or cache
condition begin in \textbf{Part~II: Output Stability and Reproducibility}. The central
distinction is whether the requirement is byte-identical replay, stable task outcomes, or
stable aggregate statistics.

Questions about throughput, first-token latency, tail latency, concurrency, memory
pressure, or cost begin in \textbf{Part~III: Serving Efficiency and Capacity}. Performance
is interpreted under a stated workload rather than as a checkpoint constant.

\Needspace{5\baselineskip}
Questions about JSON constraints, tool calls, prompted reasoning budgets, long-context
tasks, hybrid attention, or mixture-of-experts comparisons begin in \textbf{Parts~IV
and~V}. These chapters ask what each control or comparison actually establishes.

Questions about reporting, audit, replication, or release criteria begin in
\textbf{Part~VI: The Field Guide} and the appendices.

The detailed numerical lookup tables do not appear in the front matter. They belong in the
Claim and Evidence Index, after the concepts needed to interpret them have been introduced.

\begin{figure}[htbp]
\centering
\widefig{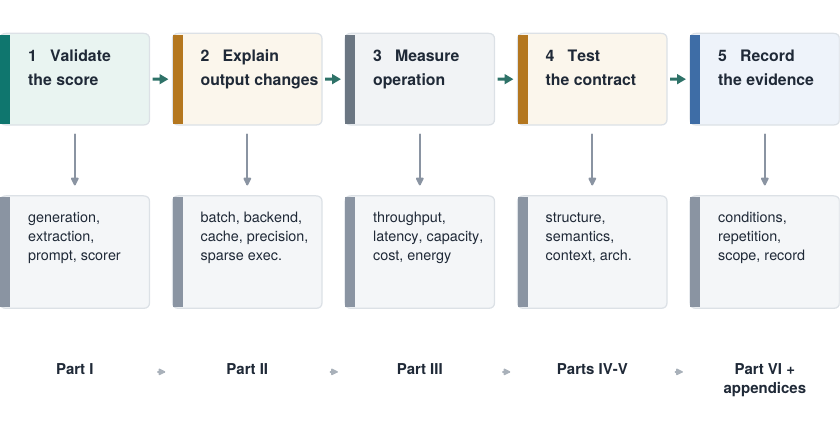}
\caption{\textbf{The life of an inference claim.} The book follows a result from score
validity through execution, operation, contract, and evidence record. This is a reading
map, not a measured result.}
\end{figure}

\clearpage

\addchap{Evidence and Scope}
\markboth{Evidence and Scope}{Evidence and Scope}

Every quantitative result in this book is conditional. The condition includes the
checkpoint, task or workload, prompt and scoring path, decoding settings, inference engine
and version, numerical precision, relevant hardware and topology, and the method used to
aggregate or compare the measurements.

Those conditions are printed with the result. A model family name alone is not sufficient
when tokenizer revisions, chat templates, quantized derivatives, engine versions, or
thinking-mode defaults can change the path being measured. A throughput number is
incomplete without workload shape and concurrency. A latency number is incomplete without
the offered load and percentile. A cost estimate is incomplete without its denominator and
included resources.

\section*{Types of evidence}

The book distinguishes direct measurements from derived quantities. A measured KV-token
pool is a direct observation from the engine; a maximum number of equal-length sequences
calculated from that pool is a derived planning bound. A directly measured p99 latency is
different from a qualitative expectation about queueing. A same-engine checkpoint
comparison establishes an operational difference between the tested checkpoints;
attribution to one architectural component requires a matched ablation.

Repeated results are identified when the study was run across seeds, repetitions, slices,
or model families. A result measured once remains usable when its scope is stated, but it
is not presented with the precision or generality of a repeated result. Boundary results
carry the limiting condition in the same paragraph, caption, or table note as the number.

The complete editorial history---whether a value was preliminary, superseded, unresolved,
or selected for the final text---belongs in Chapter~29 and the Claim and Evidence Index. It
is not necessary to learn that taxonomy before reading Chapter~1.

\section*{Original measurements and external sources}

External citations identify prior methods, benchmark definitions, model reports, standards,
and software designs. They do not establish the numerical effects measured in this book.
Those effects are established by the comparisons reported in the relevant chapters, together
with their printed conditions and uncertainty.

Study identifiers may appear in figure or table notes and are collected in the Claim and
Evidence Index. They are provenance aids, not substitutes for a complete description of the
condition.

Where public supplementary material is provided, it supplements rather than replaces the
book. No primary conclusion should require access to an unpublished file, local path,
manifest, or repository to be interpreted.

\section*{Terminology}

A \textbf{checkpoint} is the exact model object under test, including its revision or
quantized derivative. An \textbf{engine} is the inference runtime and version. A
\textbf{node} is a physical compute host; a \textbf{GPU} or \textbf{accelerator} is a device
within it. A \textbf{service} or \textbf{server} is a running endpoint. A \textbf{run record}
describes the condition that produced a result; Chapter~26 shows how to implement and
validate it.

The book uses percentage points for differences between percentages, percent for relative
changes or rates, and multiplicative notation for ratios. Context length states whether it
means prompt, active-sequence, or prompt-plus-reserved-output length. Tokens per second
states whether it counts generated output or all processed tokens.

These conventions make each result readable on its own without turning the narrative into
repository documentation.

\clearpage

\mainmatter

\part{Evaluation Validity}

\begin{partopener}
Before a benchmark score can be interpreted as model capability, the measurement path must be specified and checked.
\end{partopener}

The chapters in this part examine five places where that path can dominate the reported
result. A generation limit can prevent an otherwise valid response from reaching its
answer. An extractor can assign different scores to identical generated text. A
multiple-choice scoring path can interact with a model's prompt template and tokenization.
A quantized checkpoint can behave differently across tasks and toolchains. A generation
budget can remain too short to measure capability or become so long that additional tokens
add no measurable benefit.

Each chapter holds the model and data fixed while changing one primary condition. The
objective is not to dismiss benchmarks or to argue that one universal harness is correct.
It is to identify the condition under which the reported score becomes interpretable for
the intended comparison.

By the end of Part~I, a benchmark result should state the generation budget and stopping
behavior, the extraction and normalization rule, the prompt and scoring path, the decoding
settings, the exact checkpoint and task slice, and the uncertainty appropriate to the
comparison.

Chapter~1 begins with the most direct failure: a model that appears unable to solve a task
because the generation limit does not allow the response to finish.

\chapter{Generation Limits Can Masquerade as Model Failure}

\measuredq{How much can a generation cap change a benchmark score when the model, prompts, decoding, and scoring are held fixed?}

\begin{figure}
\centering
\includegraphics[width=\linewidth]{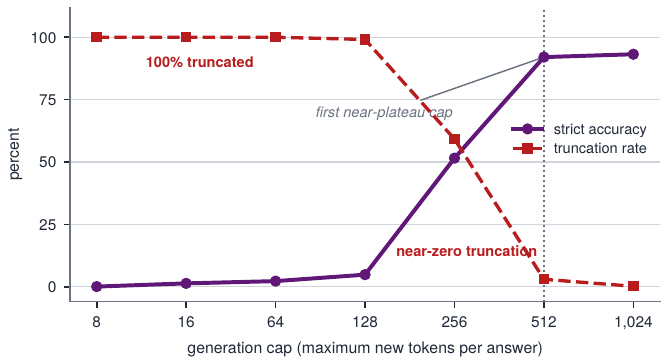}
\caption{\textbf{Accuracy and truncation across generation caps.} Qwen3-8B on the GSM8K test set (n = 1,319), vLLM 0.11.0, greedy decoding, temperature = 0. Strict accuracy increased from 0.08\% at a cap of 8 new tokens to 93.18\% at 1,024, while truncation fell from 100.0\% to 0.2\%. The main transition occurred between 256 and 512 tokens. Source: E1.}
\end{figure}

\medskip{}

A benchmark score can reflect the generation limit rather than the capability intended for
measurement. In E1, Qwen3-8B \autocite{qwen2025qwen3} was evaluated on
GSM8K \autocite{cobbe2021gsm8k} while a single setting was swept from 8 to 1,024: the
\emph{generation cap}, the largest number of new tokens the engine is allowed to produce
for one answer. The cap is a harness setting rather than a property of the model; it
appears as \code{max\_new\_tokens} in the Hugging Face \code{generate} interface and as
\code{max\_tokens} in vLLM sampling parameters. Model weights, prompts, decoding, and
extraction rules were held fixed.

At a cap of eight tokens, strict accuracy was 0.08\% and every generation reached the cap. At 1,024 tokens, strict accuracy was 93.18\% and only 0.2\% of generations were truncated. The 93.10-percentage-point movement came from the inference configuration, not from a change in the model.

\section{1.1 Why short caps fail on long-form tasks}

Generation caps originally suited tasks whose outputs were short labels or spans. A classification answer such as a single letter can fit within a small limit. A mathematical solution may require a sequence of intermediate steps before the final number appears. When the cap is reached first, the extractor receives an incomplete response and records a failure even when the preceding reasoning is useful.

This does not imply that longer output is always better. It means that the configured cap must be long enough for the task's normal completion distribution. The relevant quantity is therefore not the cap alone, but the relationship among the cap, actual generation length, stopping reason, and final-answer position.

\section{1.2 Diagnosing truncation and locating the transition}

The primary diagnostic is \emph{truncation rate}: the fraction of generations that stop because the configured token limit is reached. A score accompanied by high truncation is not a clean capability estimate. The cap is still constraining the measured behavior.

In the E1 sweep, caps of 8, 16, and 64 tokens truncated every item. At 256 tokens, 59.2\% of generations were still truncated and strict accuracy was 51.55\%. At 512 tokens, truncation fell to 3.0\% and accuracy reached 92.04\%. Increasing the cap to 1,024 reduced truncation to 0.2\% and added 1.14 percentage points. For this model--task pair, 512 tokens was the first near-plateau setting and 1,024 was the plateau reference used in the later ceiling study.

\begin{figure}
\centering
\includegraphics{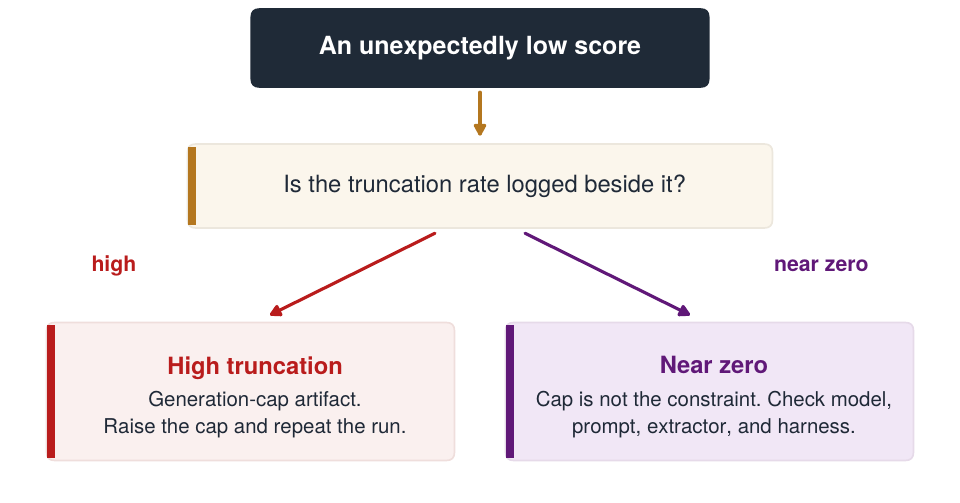}
\caption{\textbf{Truncation diagnostic.} When an unexpectedly low score is accompanied by substantial truncation, increase the cap and repeat the measurement. When truncation is near zero, investigate the model, prompt, extractor, and harness rather than the generation limit. This diagram is a diagnostic synthesis of E1, not a separate experiment.}
\end{figure}

\section{1.3 Reasoning-output models require separate length accounting}

Models that emit an explicit reasoning block can spend most of the generation budget before producing a final-answer field. A separate E6 run on Qwen3-8B with thinking enabled illustrates the boundary. At a total cap of 256 tokens, mean reasoning length was 256 tokens, mean answer length was zero, truncation was 100\%, and accuracy was 8.4\%. The result should not be compared with a non-thinking run at the same total cap as though both conditions allocated equal room to the final answer.

For such models, an evaluation record should separate reasoning tokens from answer tokens where the format permits. A total cap without that split can conceal why an answer was absent.

\section{1.4 Operational rule}

A generation-based benchmark should report four length fields with every accuracy result:

\begin{enumerate}[leftmargin=1.6em,itemsep=2pt,topsep=3pt]
\item the configured maximum number of new tokens;
\item the actual generation-length distribution, preferably median and high percentiles;
\item the share of outputs that reached the cap; and
\item the stopping conditions used by the engine.
\end{enumerate}

The cap should be selected on a held-out calibration slice. Increase it until truncation is near zero and the score is within a predefined tolerance of the plateau. The selected value is specific to the model, task, prompt format, and reasoning mode; it should not be inherited from a harness default without measurement.

\begin{table}[htbp]\centering\TableFontNine\TableRowAir
\TableTitle{Generation-cap sweep}
\begin{tabularx}{\linewidth}{@{}S[table-format=4.0] S[table-format=2.2] S[table-format=3.1] S[table-format=+2.2,print-implicit-plus] >{\RaggedRight\arraybackslash}X@{}}
\arrayrulecolor{accent}\toprule
{\thd{Cap}} & {\thd{Strict acc.}} & {\thd{Truncation}} & {\thd{$\Delta$ prior}} & \thd{Interpretation under this run}\\[-4pt]
{\thu{tokens}} & {\thu{percent}} & {\thu{percent}} & {\thu{pp}} & \\
\arrayrulecolor{rule}\midrule
8     & 0.08  & 100.0 & {n/a}  & Fully cap-limited\\
\zrow 16 & 1.36 & 100.0 & +1.28  & Fully cap-limited\\
64    & 2.27  & 100.0 & +0.91  & Fully cap-limited\\
\zrow 128 & 4.85 & 99.1 & +2.58 & Almost fully cap-limited\\
256   & 51.55 & 59.2  & +46.70 & Capability and truncation mixed\\
\zrow 512 & 92.04 & 3.0 & +40.49 & First near-plateau setting\\
1024  & 93.18 & 0.2   & +1.14  & Plateau reference\\
\arrayrulecolor{accent}\bottomrule
\end{tabularx}
\numcaptab{Generation-cap sweep. Changing the cap from 8 to 1,024 tokens moved strict accuracy from 0.08\% to 93.18\% as truncation fell from 100.0\% to 0.2\%. The score becomes interpretable only after truncation approaches zero.}
\TableNote{The first row has no prior cap, so its delta is not applicable. Qwen3-8B; GSM8K test, n = 1,319; vLLM 0.11.0; greedy decoding; temperature = 0. Evidence: E1.}
\end{table}

\begin{findingbox}In E1, changing the generation cap from 8 to 1,024 tokens moved strict GSM8K accuracy from 0.08\% to 93.18\% while truncation fell from 100.0\% to 0.2\%. A generation score is not interpretable as capability while a substantial share of responses end at the cap. Report the cap, actual length distribution, stopping reason, and truncation rate, and calibrate the cap for each model--task condition.
\end{findingbox}

Once outputs are allowed to finish, the next measurement choice is how the final answer is extracted from the completed text.

\chapter{Answer Extraction Is Part of the Evaluation}

\measuredq{How much can the reported score change when the generated text is held byte-for-byte constant and only the answer extractor changes?}

\begin{figure}
\centering
\includegraphics{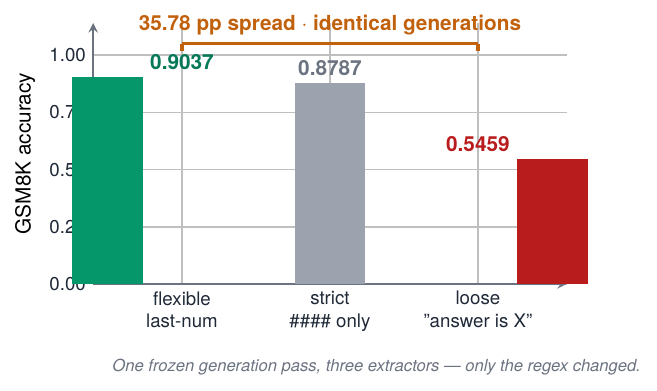}
\caption{\textbf{One generation set, three extraction rules.} A frozen set of Qwen3-8B GSM8K outputs was scored with three heuristic extractors. Accuracy ranged from 54.59\% to 90.37\%, a spread of 35.78 percentage points, although the generated bytes were identical in every condition. Conditions: GSM8K test, n = 1,319; generation cap = 512; vLLM 0.11.0; greedy decoding. Source: E2.}
\end{figure}

\medskip{}

A generation benchmark includes a second computation after the model finishes: the generated text must be converted into the answer used by the scorer. That conversion is not neutral. It encodes assumptions about markers, punctuation, number formats, and the location of the final answer.

E2 isolated this step by generating once and freezing the outputs. The same bytes were then scored with three heuristic extractors. The last-number rule produced 90.37\% accuracy, the strict number-after-\texttt{\#\#\#\#} rule produced 87.87\%, and the loose phrase/equals-pattern heuristic produced 54.59\%. The 35.78-point spread was created entirely downstream of generation.

\section{2.1 Why surface-form rules disagree}

A mathematically correct answer can appear in several textual forms: \texttt{18}, \texttt{\$18}, \texttt{18.0}, a sentence ending in 18, or a marked final-answer line. A surface-form extractor recognizes only the patterns encoded in its implementation. When a correct response uses a different form, extraction can fail even though the underlying answer is present.

In this output set, the strict \texttt{\#\#\#\#} rule returned 1,159 correct answers. The last-number rule returned 1,192, thirty-three more. Inspection recorded that some responses continued after a marker or contained a later final value, allowing the last-number rule to recover answers the strict rule missed. This is an observed behavior of the tested output set, not evidence that last-number extraction is universally preferable.

The loose heuristic failed in the opposite direction. It returned 720 correct answers because patterns intended to identify a final answer also matched intermediate arithmetic in some generations. A permissive rule can therefore introduce false extraction as well as recover format variation.

\section{2.2 The extractor is part of the measurement contract}

The benchmark result can be written schematically as a pipeline:

\begin{center}
{\small\sffamily\color{ink}\texttt{prompt} $\rightarrow$ generated text $\rightarrow$ \textbf{\color{accent}extractor} $\rightarrow$ normalized answer $\rightarrow$ score}
\end{center}

Holding the first two stages fixed did not hold the score fixed. The extraction and normalization stages must therefore be treated as part of the benchmark definition rather than as an unreported post-processing detail.

E2 does not establish that one of the three heuristics is correct for all mathematical outputs. It establishes that the choice is consequential and that plausible rules can fail in different ways. It also does not compare base and instruct models; claims about differential formatting effects between those model classes require a separate experiment.

\section{2.3 Operational rule}

Every generation-based accuracy result should report:

\begin{enumerate}[leftmargin=1.6em,itemsep=2pt,topsep=3pt]
\item the extractor name and version or the exact rule;
\item numerical normalization rules, including commas, signs, decimals, units, and percentages;
\item treatment of multiple candidate answers;
\item treatment of missing or malformed answers; and
\item the verifier or scorer used after extraction.
\end{enumerate}

For mathematics, equivalence-aware verification is preferable in principle to literal surface matching because equivalent forms should receive the same score. It should still be validated against a manually labeled sample, including outputs with intermediate numbers, multiple boxed expressions, units, and malformed formatting. Comparisons across models must use the same validated extraction pipeline unless the purpose of the study is to measure formatting compliance itself.

\begin{table}[htbp]\centering\TableFontNine\TableRowAir
\TableTitle{Scores from identical generated bytes}
\begin{tabularx}{\linewidth}{@{}Q{1.28in} S[table-format=2.2] S[table-format=4.0] >{\RaggedRight\arraybackslash}X@{}}
\arrayrulecolor{accent}\toprule
\thd{Extraction rule} & {\thd{Accuracy}} & {\thd{Correct}} & \thd{Observed behavior}\\[-4pt]
 & {\thu{percent}} & {\thu{items}} & \\
\arrayrulecolor{rule}\midrule
Last number in the response & 90.37 & 1192 & Recovered later final values in some responses\\
\zrow Number after the literal \texttt{\#\#\#\#} marker & 87.87 & 1159 & Missed responses whose final value did not sit at the marker\\
Loose phrase/equals-pattern heuristic & 54.59 & 720 & Selected intermediate arithmetic in some responses\\
\arrayrulecolor{accent}\bottomrule
\end{tabularx}
\numcaptab{Scores from identical generated bytes. Three heuristic extractors scored one frozen generation set from 54.59\% to 90.37\%; the extractor alone changed the reported score by 35.78 percentage points.}
\TableNote{Generated bytes identical across rows; GSM8K test, n = 1,319; cap = 512; Qwen3-8B; vLLM 0.11.0; greedy decoding. Evidence: E2. The strict rule accepts a number only after the literal \texttt{\#\#\#\#} marker; the last-number rule takes the final numeric expression anywhere; the loose rule checks ``answer is,'' ``final answer,'' or a terminal equals pattern before falling back to the last number. These are operational descriptions, not distributed reference code.}
\end{table}

\begin{findingbox}In E2, one frozen generation set scored between 54.59\% and 90.37\% under three heuristic extraction rules. The experiment establishes that answer extraction and normalization are part of the benchmark definition. E2 did not measure an equivalence verifier or compare base and instruct models. Report the exact extractor, validate it on labeled examples, and hold it fixed across model comparisons.
\end{findingbox}

After extraction, multiple-choice evaluation introduces another measurement choice: whether the model generates an answer or is scored through token probabilities under a particular prompt template.

\chapter{MMLU Scores Depend on the Scoring Path}

\measuredq{How much can a Massive Multitask Language Understanding (MMLU) \autocite{hendrycks2021mmlu} score change when model weights and questions are held fixed but the prompt template, few-shot context, and scoring path change?}

\begin{figure}
\centering
\includegraphics{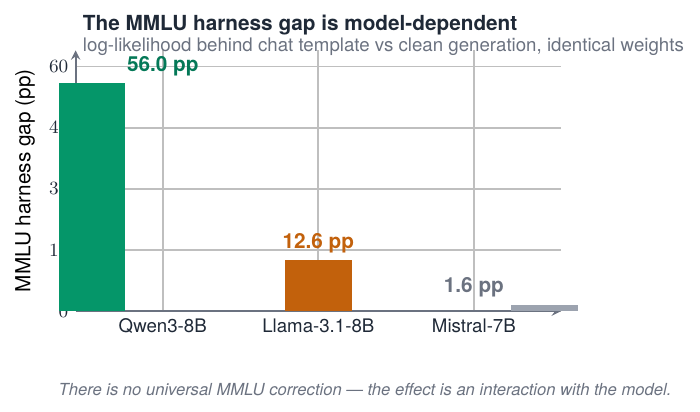}
\caption{\textbf{Four scoring configurations on identical weights.} In the original E3 run, three compatible configurations scored between 75.20\% and 77.20\%, while first-position answer-letter scoring behind a chat template scored 21.10\%. The full configuration spread was 56.10 percentage points; the compatible-configuration spread was 2.00 points. Conditions: MMLU, 1,000-item sample, vLLM 0.11.0, greedy where generation was used. Model recorded in the E3 journal as Qwen3-14B. Source: E3.}
\end{figure}

\medskip{}

Multiple-choice benchmarks can be evaluated through different computational paths. One path asks the model to generate an answer and then extracts a choice letter. Another scores the likelihood assigned to each answer option at a specified token position. Both can be valid, but only when the prompt format and scoring position match the assumptions of the method.

E3 held the weights and question sample fixed while varying the scoring configuration. Generate-and-parse with a chat template scored 77.20\% at zero shot and 75.20\% at five shot. A conventional raw-prompt, five-shot choice-log-likelihood path scored 75.80\%. These three results were within 2.00 percentage points.

A fourth condition applied first-position answer-letter scoring behind a chat template and scored 21.10\%. Under that path, the scorer assumed that the relevant answer-letter probability occurred at the first scored position. The applied prompt format did not reliably satisfy that assumption. The resulting near-chance score was therefore a scoring-path incompatibility rather than a clean estimate of subject knowledge.

\section{3.1 Two different lessons from one sweep}

The E3 sweep supports two distinct conclusions. First, an incompatible combination of template and scoring position can dominate the result, producing a 56.10-point full spread. Second, even the three compatible configurations differed by 2.00 points. That smaller difference can still affect comparisons among models separated by only one or two points.

The full evaluation condition must therefore include more than the dataset name. It should identify the prompt template, few-shot count, exact examples and order, choice formatting, scoring method, scored token position, normalization rule, and model tokenizer.

\section{3.2 The interaction is model-dependent}

The hardening study repeated the generate-versus-log-likelihood contrast on additional model families. The reported gap was 12.6 points on Llama-3.1-8B \autocite{grattafiori2024llama3} and 1.6 points on Mistral-7B \autocite{jiang2023mistral}, compared with 56 points in the Qwen condition. The qualitative result was therefore not universal: the scoring-path interaction was large on one family, material on another, and small on the third.

The study did not isolate why the magnitudes differed. Possible contributors include where each prompt template places the expected answer, how answer letters are tokenized, the calibration of choice probabilities, and post-training behavior. These are hypotheses rather than measured mechanisms. The measured conclusion is narrower: a scoring path validated on one family should not be assumed equivalent on another.

\begin{table}[htbp]\centering\TableFontNine\TableRowAir
\TableTitle{Cross-family scoring-path gap}
\begin{tabularx}{\linewidth}{@{}Q{1.32in} S[table-format=2.1] >{\RaggedRight\arraybackslash}X@{}}
\arrayrulecolor{accent}\toprule
\thd{Model / condition} & {\thd{Gap}} & \thd{Interpretation under this run}\\[-4pt]
 & {\thu{pp}} & \\
\arrayrulecolor{rule}\midrule
Qwen3-14B (E3) & 56.0 & Large path incompatibility here\\
\zrow Llama-3.1-8B (H2) & 12.6 & Material interaction on replication\\
Mistral-7B (H2) & 1.6 & Small interaction; family-specific\\
\arrayrulecolor{accent}\bottomrule
\end{tabularx}
\numcaptab{Cross-family scoring-path gap. The gap is an interaction between model and scoring path, not a constant to subtract; the magnitude varies across model families, so the exact condition is reported per row.}
\TableNote{Loglikelihood-behind-chat-template vs generate-and-parse, identical weights. The H2 replication plan specifies MMLU-500; the original E3 run used 1,000 items. Evidence: E3 and H2.}
\end{table}

\section{3.3 Few-shot example order is another measured condition}

E5 held the model, five examples, and 1,000 MMLU questions fixed while changing only the order of those examples. Accuracy ranged from 35.8\% to 69.8\%, a 34.0-point spread. The natural source order happened to produce the lowest score in this five-order sweep.

This result demonstrates that example order can be consequential under a particular prompt and model. It does not estimate the average order effect across all permutations or model families. A stronger general claim would require more permutations, repeated samples, and additional models. The chapter states the observed range and its scope without labeling individual orders safe or unsafe.

\begin{table}[htbp]\centering\TableFontNine\TableRowAir
\TableTitle{Five orders of the same five examples}
\begin{tabularx}{\linewidth}{@{}>{\RaggedRight\arraybackslash}X C C@{}}
\arrayrulecolor{accent}\toprule
\thd{Order ID} & \thd{Example permutation} & \thd{Accuracy (percent)}\\
\arrayrulecolor{rule}\midrule
0 & \texttt{[0, 1, 2, 3, 4]} & 35.8\\
\zrow 1 & \texttt{[3, 4, 0, 1, 2]} & 51.0\\
2 & \texttt{[3, 2, 1, 4, 0]} & 48.0\\
\zrow 3 & \texttt{[4, 1, 3, 0, 2]} & 69.8\\
4 & \texttt{[4, 2, 3, 0, 1]} & 68.9\\
\arrayrulecolor{accent}\bottomrule
\end{tabularx}
\numcaptab{Five orders of the same five examples, 34.0 points apart. Only the order changed; the model, examples, and questions were fixed. This is one five-order sweep, not an uncertainty estimate over the space of permutations.}
\TableNote{Qwen3-8B; MMLU, n = 1,000; five fixed exemplars; vLLM 0.11.0; greedy decoding. Evidence: E5.}
\end{table}

\section{3.4 Operational rule}

An MMLU result should be treated as a tuple rather than a single model--dataset pair: the model checkpoint, tokenizer, prompt template, few-shot examples and order, scoring path, scored position, normalization, harness and version, and sample and seed. Comparisons should hold that tuple fixed except for the model variable being studied.

When a new model family is introduced, verify that the scoring path is compatible with its prompt format and that the expected answer position is actually the position being scored. A small manually inspected sample can detect catastrophic path mismatches before a full run.

\begin{findingbox}The original E3 sweep produced a 56.10-point full spread, but three compatible configurations were within 2.00 points and the near-chance row came from a template--scoring-position incompatibility. Cross-family hardening showed that the gap was model-dependent: 56 pp on the Qwen run, 12.6 pp on Llama, 1.6 pp on Mistral. Report the complete scoring path, validate it for each family, and pin few-shot examples and order in every comparison.
\end{findingbox}

The next comparison holds the scoring path fixed and changes the model artifact through quantization.

\chapter{Quantization Effects Vary by Task and Toolchain}

\measuredq{Does a quantized model artifact that matches BF16 on one benchmark also match it across different task types?}

\begin{figure}
\centering
\includegraphics[width=\linewidth]{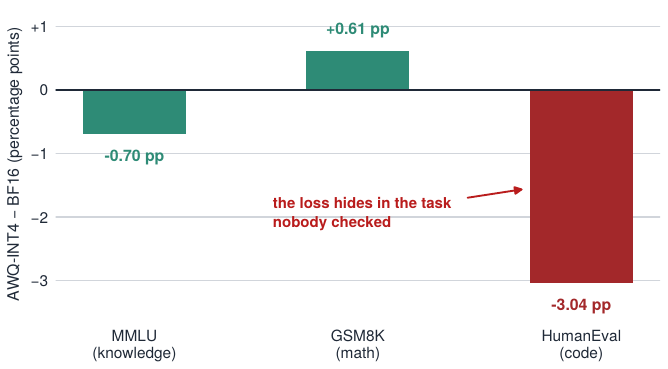}
\caption{\textbf{Task-specific deltas for one Qwen3-14B AWQ artifact.} Relative to BF16, AWQ-INT4 changed MMLU by $-0.70$ percentage points, GSM8K by $+0.61$ points, and HumanEval by $-3.04$ points. Conditions: Qwen3-14B; vLLM 0.11.0; greedy decoding; MMLU n = 1,000, GSM8K n = 1,319, HumanEval n = 164. Source: E8. The study was not seed-repeated, so small deltas should not be read as precise effects.}
\end{figure}

\medskip{}

A quantization decision applies to a model artifact, not to a benchmark in the abstract. The result can depend on the model family, quantization method, calibration procedure, runtime kernels, and task. A near-zero change on one benchmark therefore does not establish that the artifact is equivalent across a deployment workload.

E8 compared the Qwen3-14B bfloat16 (BF16) checkpoint with a checkpoint compressed using AWQ, an activation-aware weight-quantization method \autocite{lin2024awq}. The artifact was evaluated on MMLU, GSM8K, and HumanEval \autocite{chen2021humaneval} under one fixed inference setup. MMLU changed from 76.90\% to 76.20\%, a decrease of 0.70 percentage points. GSM8K changed from 94.31\% to 94.92\%, an increase of 0.61 points. HumanEval changed from 87.80\% to 84.76\%, a decrease of 3.04 points, corresponding to five fewer passing problems out of 164.

\section{4.1 The Qwen result is task-specific}

The three deltas do not support a single lossless label. They form a task vector: approximately unchanged on the tested knowledge and mathematics evaluations, and lower on the tested code evaluation. A sign-off based only on MMLU would not have revealed the HumanEval difference.

E8 did not isolate the mechanism. The HumanEval change may be related to exact-sequence sensitivity, the particular AWQ calibration, sampling noise, or other model--toolchain interactions. The experiment establishes where the observed delta appeared; it does not establish a universal hierarchy in which code is always more quantization-sensitive than knowledge or mathematics.

\begin{table}[htbp]\centering\TableFontNine\TableRowAir
\TableTitle{Qwen3-14B BF16 and AWQ-INT4}
\begin{tabularx}{\linewidth}{@{}Q{0.74in} S[table-format=2.2] S[table-format=2.2] S[table-format=+1.2,print-implicit-plus] >{\RaggedRight\arraybackslash}X@{}}
\arrayrulecolor{accent}\toprule
\thd{Task} & {\thd{BF16}} & {\thd{AWQ-INT4}} & {\thd{Diff.}} & \thd{Interpretation under this run}\\[-4pt]
 & {\thu{percent}} & {\thu{percent}} & {\thu{pp}} & \\
\arrayrulecolor{rule}\midrule
MMLU & 76.90 & 76.20 & -0.70 & Small negative difference\\
\zrow GSM8K & 94.31 & 94.92 & +0.61 & No evidence of a loss\\
HumanEval & 87.80 & 84.76 & -3.04 & Largest drop; 5 of 164 items\\
\arrayrulecolor{accent}\bottomrule
\end{tabularx}
\numcaptab{Qwen3-14B BF16 and AWQ-INT4. The three task deltas are uneven: the largest negative delta appeared on HumanEval, not MMLU or GSM8K.}
\TableNote{Qwen3-14B; vLLM 0.11.0; greedy decoding; BF16 versus the Qwen3-14B-AWQ W4A16 derivative identified as Q-Q3-14-AWQ in Appendix~\ref{app:roster}. The retained notes do not preserve its immutable repository revision or calibration set, so the result is artifact-scoped rather than an exact-byte reproduction claim.}
\end{table}

\section{4.2 The pattern did not reproduce on Mistral}

The hardening program repeated the task comparison with Mistral-7B BF16 and a Mistral AWQ artifact. In that run, the AWQ score was higher on MMLU, GSM8K, and HumanEval by 4.9, 2.6, and 3.0 percentage points, respectively. The journal treated these small positive differences as within noise and concluded that the Qwen code-specific pattern did not reproduce.

This replication changes the general claim. The defensible conclusion is not that AWQ preserves knowledge while reducing code quality. It is that task-specific validation is necessary because the direction and location of quantization deltas can change with the model and quantization toolchain.

\begin{table}[htbp]\centering\TableFontNine\TableRowAir
\TableTitle{Cross-family scope check}
\begin{tabularx}{\linewidth}{@{}Q{0.74in} S[table-format=2.1] S[table-format=2.1] S[table-format=+1.1,print-implicit-plus] >{\RaggedRight\arraybackslash}X@{}}
\arrayrulecolor{accent}\toprule
\thd{Task} & {\thd{BF16}} & {\thd{AWQ}} & {\thd{Diff.}} & \thd{Scope conclusion}\\[-4pt]
 & {\thu{percent}} & {\thu{percent}} & {\thu{pp}} & \\
\arrayrulecolor{rule}\midrule
MMLU & 51.0 & 55.9 & +4.9 & Qwen's MMLU delta did not reproduce\\
\zrow GSM8K & 50.3 & 52.8 & +2.6 & No measured degradation here\\
HumanEval & 34.2 & 37.2 & +3.0 & Qwen's code-specific loss absent\\
\arrayrulecolor{accent}\bottomrule
\end{tabularx}
\numcaptab{Cross-family scope check. On Mistral-7B AWQ, the Qwen code-specific loss did not reproduce; the hardening journal characterizes these small positive deltas as within noise.}
\TableNote{Mistral-7B-Instruct-v0.3 BF16 versus its AWQ derivative, vLLM 0.23. The retained H2 notes preserve HumanEval n = 164 but not the MMLU and GSM8K sample counts or per-cell uncertainty. These rows are therefore a cross-family scope check, not precision estimates. Evidence: H2-E8.}
\end{table}

\section{4.3 Validate the artifact, not only the bit width}

A label such as INT4 does not uniquely identify a quantized model. Reproducibility requires the checkpoint, quantization method, group size, calibration data, weight and activation precision, excluded layers, runtime kernels, engine version, and hardware. Two artifacts with the same nominal bit width can differ because those choices differ.

This chapter does not establish that method is more important than bit width, nor that smaller models are always more affected. Those questions require controlled sweeps. It does establish that a benchmark-level label such as ``lossless on MMLU'' is too broad to describe an artifact intended for several task families.

\section{4.4 Operational validation protocol}

A deployment validation should begin with the intended workload rather than a generic benchmark average:

\begin{enumerate}[leftmargin=1.6em,itemsep=2pt,topsep=3pt]
\item identify the task families and output constraints represented in production;
\item compare the quantized artifact with its exact baseline under identical inference settings;
\item report per-task deltas and item-level transitions, not only a blended average;
\item repeat small or noisy evaluations across seeds or bootstrap samples;
\item record throughput, memory, and quality separately; and
\item approve the artifact only for the model, toolchain, hardware, and task envelope that was measured.
\end{enumerate}

A weighted deployment score may be useful when its weights reflect real traffic and error costs. An unweighted average of unrelated benchmark percentages should not be used as evidence that a quantization is generally lossless.

\begin{findingbox}In the Qwen3-14B E8 comparison, AWQ-INT4 was close to BF16 on MMLU and GSM8K but 3.04 points lower on HumanEval. The pattern did not reproduce on Mistral-7B AWQ. Quantization quality is therefore an artifact-, task-, and toolchain-specific property. Validate the exact artifact across the tasks represented in deployment, and report the full inference and quantization metadata.
\end{findingbox}

Appendix~\ref{app:roster} connects the checkpoint and quantized-artifact keys used here; Appendix~\ref{app:workloads} connects the benchmark and scorer names to their reader-facing definitions.

The final chapter of Part I returns to generation length and identifies the plateau above the truncation transition.

\chapter{Finding the Generation-Budget Knee}

\measuredq{What changes when the generation cap is increased from a truncating regime to the accuracy plateau, and what changes when additional headroom is added beyond that plateau?}

\begin{figure}
\centering
\includegraphics{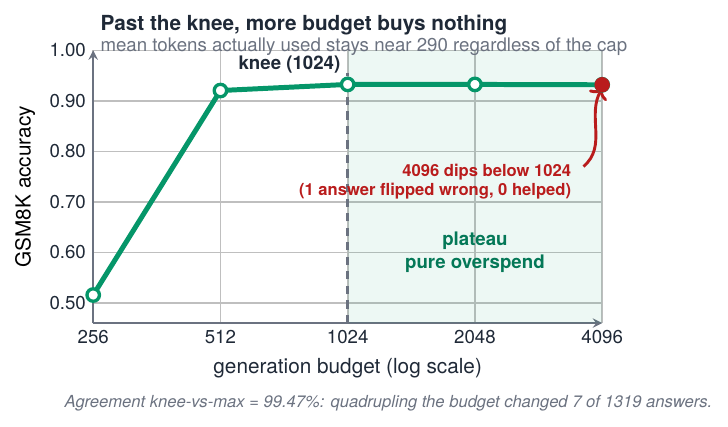}
\caption{\textbf{Accuracy plateau and actual generation length.} Qwen3-8B on GSM8K (n = 1,319), vLLM 0.11.0, greedy decoding. Accuracy increased sharply from a cap of 256 to 1,024 and then remained approximately flat through 4,096. Mean actual generation length increased from 290.4 tokens at a 1,024 cap to 293.3 at 4,096. Source: E6a.}
\end{figure}

\medskip{}

Chapter 1 established the lower boundary: a cap can truncate responses and suppress accuracy. The complementary question is what happens after truncation has largely disappeared. E6a swept the cap from 256 to 4,096 tokens while recording both accuracy and actual tokens emitted.

Accuracy rose from 51.55\% at 256 tokens to 91.96\% at 512 and 93.25\% at 1,024. It remained 93.25\% at 2,048 and 93.18\% at 4,096. Mean actual generation length, however, remained near 290 tokens once the cap reached 1,024 because most responses stopped before using the available headroom.

\section{5.1 Two regimes must be kept separate}

The evidence is clearest when the below-knee and above-plateau comparisons are presented separately.

In E6b, a 256-token cap was compared with 1,024 on a 512-item GSM8K slice. Agreement was 56.84\%. The larger cap improved 197 items, harmed none, and left 315 neutral. This comparison crossed the truncation transition, so additional room materially changed the measurement.

In E6a, a 1,024-token cap was compared with 4,096 on all 1,319 items. Agreement was 99.47\%; only seven final answers changed. No incorrect answer became correct, one correct answer became incorrect, and the remaining changed outputs did not alter correctness. This comparison remained on the accuracy plateau.

\begin{table}[htbp]\centering\TableFontNine\TableRowAir
\TableTitle{Below-knee and above-plateau comparisons}
\begin{tabularx}{\linewidth}{@{}Q{1.02in} c S[table-format=2.2] >{\RaggedRight\arraybackslash}X@{}}
\arrayrulecolor{accent}\toprule
\thd{Regime} & \thd{Cap pair} & {\thd{Agreement}} & \thd{Correctness transitions}\\[-4pt]
 & {\thu{tokens}} & {\thu{percent}} & \\
\arrayrulecolor{rule}\midrule
Below transition $\to$ plateau & 256 / 1,024 & 56.84 & 197 improved, 0 worsened, 315 neutral (n = 512); the smaller cap was still truncation-confounded\\
\zrow Plateau $\to$ more headroom & 1,024 / 4,096 & 99.47 & 0 improved, 1 worsened; 7 answers changed (n = 1,319); no measured accuracy gain\\
\arrayrulecolor{accent}\bottomrule
\end{tabularx}
\numcaptab{Below-knee and above-plateau comparisons. Below the transition, extra room recovers truncated answers; above the plateau, extra headroom changes little. These are two different experiments with different sample sizes; they must not be merged into one helps/hurts range.}
\TableNote{E6b for 256 $\to$ 1,024 (n = 512); E6a for 1,024 $\to$ 4,096 (n = 1,319). Qwen3-8B; GSM8K; vLLM 0.11.0; greedy decoding. Evidence: E6a, E6b.}
\end{table}

\section{5.2 A cap is not the same as consumed tokens}

The cap is a constraint, not a token invoice. Increasing the cap from 1,024 to 4,096 quadrupled the available maximum, but mean actual generation length rose only from 290.4 to 293.3 tokens. The measured mean-token increase was therefore approximately 1.0\%, not 4$\times$.

This distinction changes the cost interpretation. The plateau demonstrates that a larger cap did not improve average accuracy under the tested condition. It does not demonstrate a fourfold average cost increase. A complete cost statement would require actual input and output tokens, wall-clock latency, and high-percentile generation lengths. The current data suggest a small mean increase and a possible tail risk from rare long generations; the tail was not quantified in E6a.

\begin{table}[htbp]\centering\TableFontNine\TableRowAir
\TableTitle{E6a cap sweep}
\begin{tabularx}{\linewidth}{@{}S[table-format=4.0] S[table-format=2.2] S[table-format=3.1] S[table-format=+2.2,print-implicit-plus] >{\RaggedRight\arraybackslash}X@{}}
\arrayrulecolor{accent}\toprule
{\thd{Cap}} & {\thd{Strict acc.}} & {\thd{Mean used}} & {\thd{$\Delta$ vs 1,024}} & \thd{Interpretation}\\[-4pt]
{\thu{tokens}} & {\thu{percent}} & {\thu{tokens}} & {\thu{pp}} & \\
\arrayrulecolor{rule}\midrule
256   & 51.55 & 235.0 & -41.70 & Still below the usable range\\
\zrow 512 & 91.96 & 286.3 & -1.29 & Near the plateau\\
1024  & 93.25 & 290.4 & {0.00}  & Plateau reference\\
\zrow 2048 & 93.25 & 290.7 & {0.00} & No measured accuracy gain\\
4096  & 93.18 & 293.3 & -0.07  & No gain; 1 answer flipped wrong\\
\arrayrulecolor{accent}\bottomrule
\end{tabularx}
\numcaptab{E6a cap sweep. Accuracy plateaued while mean actual generation length stayed near 290 tokens: the cap rose 16-fold while mean actual token use rose only from 235 to 293. Report cap and actual use as separate quantities.}
\TableNote{Qwen3-8B; GSM8K test, n = 1,319; vLLM 0.11.0; greedy decoding; temperature = 0. Evidence: E6a.}
\end{table}

\section{5.3 Selecting an operating cap}

A practical cap can be selected on a held-out calibration slice: sweep a range of caps; record accuracy, truncation, actual length distribution, and stopping reason; identify the smallest cap whose accuracy lies within a declared tolerance of the plateau and whose truncation rate is acceptably low; confirm the setting on a separate slice; and repeat the calibration when the model, prompt template, task, or reasoning mode changes.

The choice of tolerance is operational. For one application, a one-point difference may be material; for another, the latency or memory cost may dominate. The chapter does not call one cap safe in the abstract.

\section{5.4 Adaptive escalation is a proposed extension}

A system may attempt to use a lower cap for routine items and escalate selected items to a larger cap. E6b motivates this idea by showing that a below-knee cap harms a subset of problems. The experiments reported here did not evaluate a difficulty classifier, uncertainty signal, or escalation policy, so adaptive routing should be presented as a proposed design rather than a measured result.

A valid follow-up would define the routing signal in advance, measure the fraction escalated, and compare accuracy, actual tokens, latency, and cost against a single calibrated cap. Without such a study, the operational recommendation is simply to calibrate the global cap rather than assume that the largest available value is necessary.

\section{5.5 Reasoning-output modes}

Reasoning-output modes require their own sweep because the final answer may appear after a long reasoning block. Record reasoning tokens and answer tokens separately when possible. Do not assume that a prompt-level request for a reasoning budget will be obeyed precisely; later experiments in the book test adherence directly. The same principle still applies: choose limits from measured completion behavior rather than from an inherited default.

\begin{findingbox}Below the transition, additional room recovered truncated answers: in E6b, moving from 256 to 1,024 tokens improved 197 of 512 items and harmed none. Above the plateau, additional headroom changed little: in E6a, 1,024 and 4,096 agreed on 99.47\% of 1,319 answers, with 0 improvements and 1 regression. Because mean actual use rose only 2.9 tokens, cost must be measured from consumed tokens and latency rather than inferred from the cap.
\end{findingbox}

The first five chapters establish the measurement conditions required before a benchmark number can be interpreted as model capability.

\chapter*{The Measurement Contract for Part I}
\addcontentsline{toc}{chapter}{The Measurement Contract for Part I}
\markboth{The Measurement Contract for Part I}{The Measurement Contract for Part I}

The first five chapters show that a benchmark result is produced by a pipeline, not by model weights alone. Generation limits determine whether an answer can finish; extraction rules determine which part of the text is scored; prompt and scoring paths define the multiple-choice measurement; quantization artifacts interact with tasks and toolchains; and cap selection determines whether an evaluation is still truncation-confounded or already on the plateau.

Those five findings are used in two different situations, so they are collected here in two forms.

The first situation is diagnostic. A score has already come back wrong, and the question is which stage of the pipeline produced it. Map~I.1 orders the checks by how cheap they are to run and by how large the observed effect was in Part~I, so the reader can work down the list by symptom rather than reading the chapters again.

The second situation is editorial. A number is about to be published, and the question is what has to be printed beside it for a reader to interpret it. Table~I.1 is that list. It has one row per chapter: the dependency that chapter measured, the condition under which the measurement holds, and the fields a report must carry. The two are meant to be used together, the map before the run and the table before the release.

\begin{figure}[htbp]
\centering
\includegraphics[width=\linewidth]{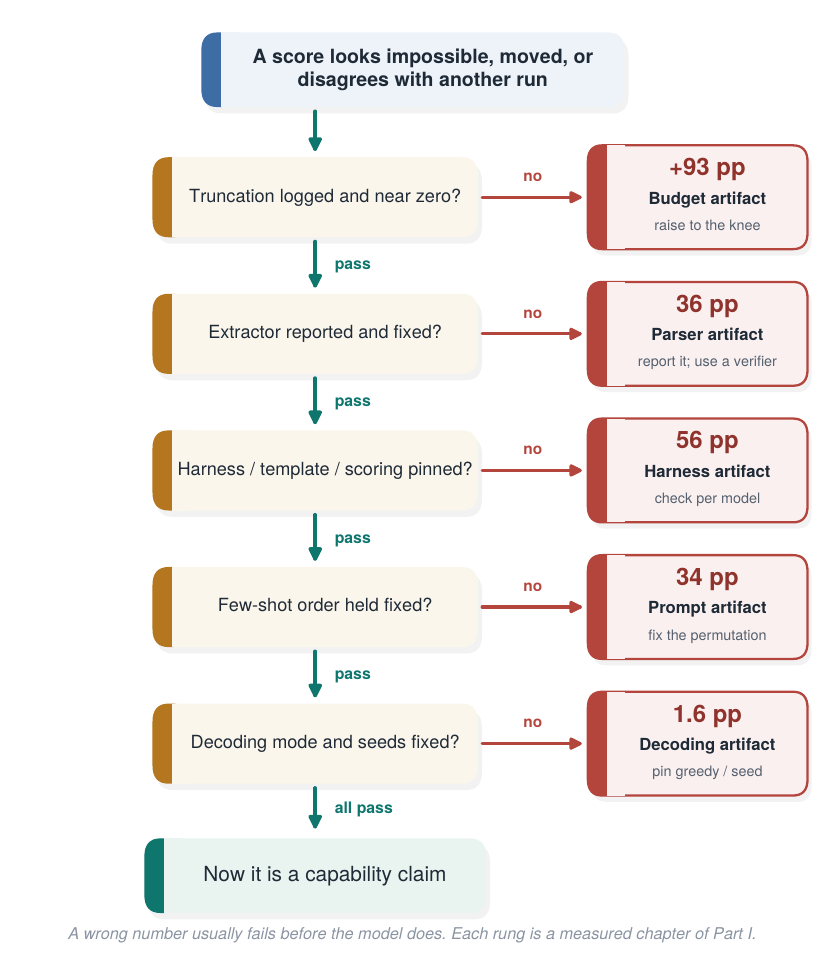}
\caption*{\textbf{Map~I.1.\ Evaluation triage for Part~I.} A score that looks impossible, has moved, or disagrees with another run is checked through decoding, generation budget, extractor, harness, and backend before it is read as a capability result. The percentage-point figures on the right are the effect sizes measured in Chapters~1--5, not general constants. Synthesis of Chapters~1--5; consult it by symptom.}
\end{figure}

\begin{statictable}\TableFontEightSix\TableRowAir
\TableTitle{Measurement contract for evaluation results}
\begin{tabularx}{\linewidth}{@{}Q{0.24in} >{\hsize=1.08\hsize\RaggedRight\arraybackslash}X >{\hsize=0.92\hsize\RaggedRight\arraybackslash}X >{\hsize=1.00\hsize\RaggedRight\arraybackslash}X@{}}
\arrayrulecolor{accent}\toprule
\thd{Ch.} & \thd{What moved} & \thd{Where it holds} & \thd{What to report}\\
\arrayrulecolor{rule}\midrule
1 & \textbf{Generation cap.} Strict accuracy moved from 0.08\% to 93.18\% as truncation fell from 100.0\% to 0.2\% & Qwen3-8B, GSM8K, one engine and decoding condition & Cap, actual length distribution, stopping reason, truncation rate\\
\zrow 2 & \textbf{Extractor.} Three rules scored identical bytes from 54.59\% to 90.37\% & One frozen Qwen3-8B GSM8K set; no verifier or base/instruct comparison & Exact extractor, normalization, missing-answer policy, verifier version\\
3 & \textbf{Scoring path.} Full E3 spread 56.10 pp; compatible paths within 2.00 pp; cross-family gap model-dependent & Qwen3-14B E3; sample size and condition listed per row & Template, few-shot examples and order, scoring path, scored position, harness version\\
\zrow 4 & \textbf{Quantization.} Qwen3-14B AWQ showed its largest negative delta on HumanEval; not reproduced on Mistral AWQ & Model-, artifact-, task-, and toolchain-dependent & Exact quant artifact, calibration, precision, engine, hardware, per-task deltas\\
5 & \textbf{Budget knee.} Below transition, 256 $\to$ 1,024 improved 197/512; above plateau, 1,024 $\to$ 4,096 improved 0/1,319, mean use $+2.9$ tokens & Two experiments and sample sizes; cap is not consumed tokens & Calibration sweep, truncation, actual tokens, latency distribution, chosen tolerance\\
\arrayrulecolor{accent}\bottomrule
\end{tabularx}
\TableCaption{\textbf{Table I.1.} Measurement contract for evaluation results. One row per chapter: the dependency that was measured, the condition under which it holds, and the fields that must be printed with the number. Read with Map~I.1, which orders the same five checks by symptom.}
\TableNote{Synthesis of Chapters 1--5. Evidence: E1, E2, E3/H2, E8/H2-E8, E6a/E6b.}
\end{statictable}

Part II can then ask a different question: when the complete measurement condition is fixed, which numerical execution choices can still change the generated text, and whether those text changes alter correctness.
\clearpage

%
%


\part{Output Stability and Reproducibility}

\begin{partopener}
How batch context, sparse execution, attention backends, and prefix caching affect generated text, and how to measure whether that affects the task outcome.
\end{partopener}

Part I established that a reported score is interpretable only when its measurement conditions are explicit. Part II applies the same discipline to generated text. In the experiments that follow, the model weights and prompt remain fixed while one execution condition changes: batch context, sparse-model path, attention backend, or prefix-cache state.

The part separates two questions. The first is whether the generated bytes differ. The second is whether the task outcome changes. Chapters 6 to 9 address the first question. Chapter 10 performs the second measurement for prefix caching on GSM8K. Exact replay and task-level correctness are therefore treated as related but distinct requirements.

\begin{figure}[htbp]
\centering
\includegraphics[width=\linewidth]{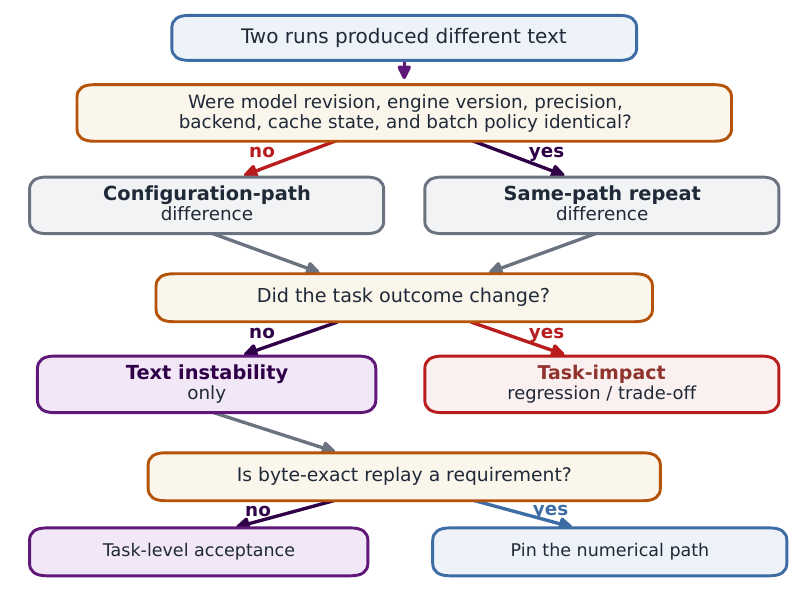}
\caption*{\textbf{Map~II.1.\ A reproducibility decision path.} When two runs of a fixed prompt produce different text, first ask whether the execution path was identical; then whether the task outcome changed; then whether byte-exact replay is actually a requirement. A decision workflow for Part II, not a measured result.}
\end{figure}

\chapter{Greedy Decoding Does Not Guarantee Bitwise Reproducibility}

\begin{figure}
\centering
\includegraphics{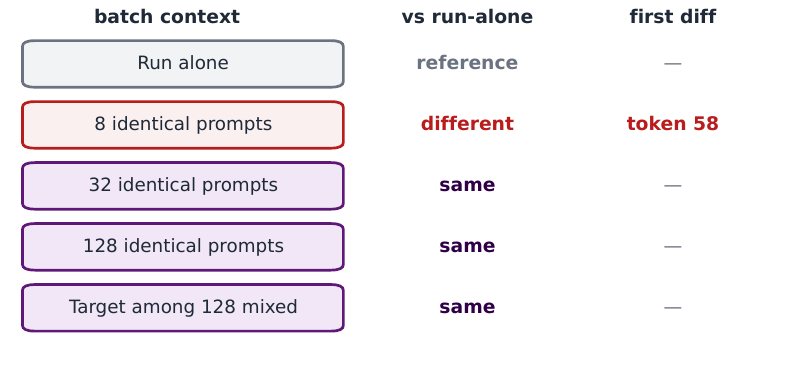}
\caption{\textbf{Illustrative batch-context sweep.} One fixed Qwen3-8B prompt was evaluated under five batch contexts with greedy decoding in vLLM 0.11.0. In this sweep, the batch-of-8 output first differed at token 58; the other tested contexts matched. Dashes in the figure mean that no differing token was observed. The figure demonstrates possibility, not a divergence rate. Source: E7.}
\end{figure}

\medskip{}

At temperature zero, an inference engine normally uses greedy token selection: at each decoding step it chooses the token with the highest computed logit. This removes sampling randomness. It does not guarantee that two executions will compute exactly the same logits or produce identical bytes.

\section{6.1 Controlled result}

Experiment E7 ran one fixed prompt through Qwen3-8B under five batch contexts: alone, in a batch of 8 identical prompts, in batches of 32 and 128 identical prompts, and as one target request among 128 mixed prompts. The model revision, prompt, decoding mode, and generation cap were fixed.

In the original sweep, the batch-of-8 output first differed from the run-alone output at token 58. The other tested contexts matched the run-alone completion in that sweep. This establishes that a batch-context change can alter a greedy completion. It does not estimate how often such a change occurs.

\begin{statictable}\TableFontNine\TableRowAir
\TableTitle{Illustrative E7 batch-context sweep}
\TableDeck{Read every row against the run-alone reference. Only the batch-of-8
completion differed; ``no difference observed'' describes this one sweep, not a
guarantee about later runs.}
\begin{tabularx}{\linewidth}{@{}>{\RaggedRight\arraybackslash}X >{\RaggedRight\arraybackslash}X C@{}}
\arrayrulecolor{accent}\toprule
\thd{Batch context} & \thd{Completion vs run-alone} & \thd{First difference}\\
\arrayrulecolor{rule}\midrule
Run alone & Reference completion & Not applicable\\
\zrow 8 identical prompts & Different completion & Token 58\\
32 identical prompts & Same completion & No difference observed\\
\zrow 128 identical prompts & Same completion & No difference observed\\
Target among 128 mixed & Same completion & No difference observed\\
\arrayrulecolor{accent}\bottomrule
\end{tabularx}
\numcaptab{Illustrative E7 batch-context sweep. A batch-context change altered one greedy completion; this is a possibility demonstration, not a divergence rate. One fixed prompt; the batch-of-8 context changed the completion, the others did not.}
\TableNote{Every row is a single sweep. Qwen3-8B, one fixed prompt, vLLM 0.11.0, greedy decoding, temperature zero, generation cap 256 tokens. Evidence: E7.}
\end{statictable}
\vspace{-0.45\baselineskip}

The later H1 hardening pass showed why the token-58 example must remain illustrative. Repeated batch-context sweeps did not preserve the original absolute completion counts: the hardened Qwen3-8B sweep produced one distinct completion, while another dense Qwen model produced two. The stable conclusion is therefore not that a particular batch size always causes a fork. It is that greedy decoding alone does not provide a cross-batch bitwise-equality guarantee.

\section{6.2 Interpretation}

A plausible explanation is that batch composition changes the numerical execution path. GPU kernels can use different shapes, tiling choices, and reduction orders as the batch changes. Small floating-point differences normally leave the highest-logit token unchanged. When two candidate tokens are close, however, a small difference can change the argmax; later tokens then condition on a different history.

This mechanism is consistent with the observed fork, but it was not directly instrumented in E7. The experiment did not record the top-two token margins at the divergence step or trace the internal reduction order.

\begin{figure}
\centering
\includegraphics{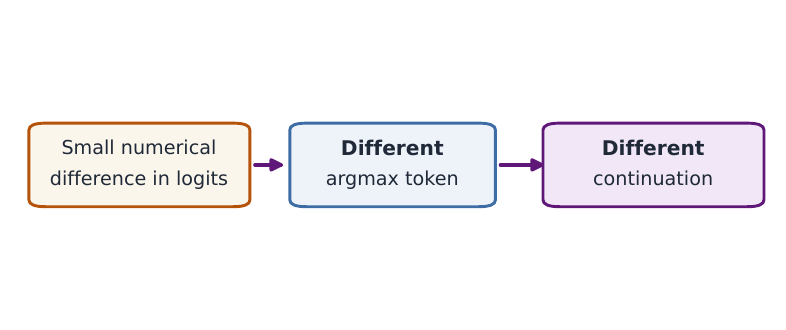}
\caption{\textbf{Schematic, not measured logits.} A small numerical difference between two near-tied candidate tokens can change the argmax, after which the continuation diverges. The near-tie values are illustrative; E7 did not record the top-two margins at the divergence token.}
\end{figure}

\section{6.3 Scope and boundary}

The experiment used one fixed prompt and a small set of batch contexts. It demonstrates possibility, not prevalence. The exact number of distinct completions and the first-divergence position are sensitive to the prompt, model, engine, generation length, and execution environment.

A fixed batch shape is also not a complete replay guarantee. Other operations, scheduling choices, precision settings, and engine versions can still alter the numerical path. Exact reproduction requires a broader execution manifest, developed in \hyperlink{chapter.11}{Chapter 11}.

\section{6.4 Operational rule}

For byte-exact replay, record the model and tokenizer revision, engine and version, attention backend, precision, cache state, and batch policy. Use deterministic operations where supported and benchmark their cost on the deployed workload; that cost was not measured in E7.

For ordinary evaluation and serving, decide whether the requirement is identical text or stable task performance. A raw-text equality test is appropriate only when byte identity is part of the product or audit requirement. Otherwise, compare the task outcome under paired conditions.

\begin{findingbox}Temperature zero removes sampling, not every source of execution variability. One E7 sweep produced a batch-dependent fork, while hardening showed that the exact fork count was not stable. Treat batch policy as part of a bitwise-reproducibility record, and do not use a single prompt as a general divergence rate.
\end{findingbox}

The next chapter asks whether sparse mixture-of-experts execution adds to this batch-context sensitivity.

\chapter{Batch Sensitivity in the Tested Mixture-of-Experts Models}

\begin{figure}
\centering
\includegraphics{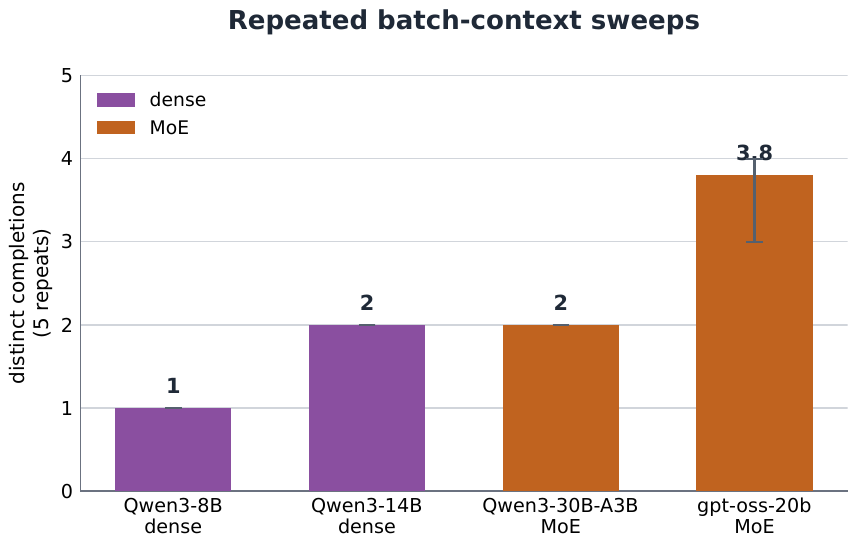}
\caption{\textbf{Batch-context variability, dense versus MoE.} Across repeated sweeps, the tested sparse models produced at least as many distinct completions as the dense controls. The original single-sweep 2-versus-4 contrast did not reproduce under hardening; treat exact counts as distributions, not headline magnitudes. Source: E7, H1, H2.}
\end{figure}

\medskip{}

Mixture-of-experts models \autocite{shazeer2017moe} add route selection and expert dispatch to the inference path. The experiments in this chapter ask a narrow question: under the tested batch-context sweeps, did sparse models produce at least as much output variability as dense controls?

\section{7.1 Controlled result}

The original E7 comparison used one fixed prompt with Qwen3-8B dense and Qwen3-30B-A3B MoE under five batch contexts. In that single sweep, the dense model produced two distinct completions and the MoE produced four; the earliest visible differences occurred at token 58 for the dense model and token 13 for the MoE.

Those exact magnitudes did not survive hardening. H1 repeated the sweep five times and reported one distinct completion for Qwen3-8B and two for Qwen3-30B-A3B. The original 2-versus-4 contrast was therefore a fragile single-sweep result. The qualitative direction, MoE variability at least as large as the dense control, survived.

A later H2 extension tested a non-Qwen MoE, gpt-oss-20b \autocite{openai2025gptoss}. It produced a mean of 3.8 distinct completions across five repeats, with a range of 3 to 4. That replication supports the broader observation that the tested MoE serving paths were not less sensitive to batch context than the dense controls.

\begin{table}[htbp]\centering\TableFontNine\TableRowAir
\TableTitle{Repeated batch-context sweeps}
\begin{tabularx}{\linewidth}{@{}Q{0.98in} l c S[table-format=1.0] >{\RaggedRight\arraybackslash}X@{}}
\arrayrulecolor{accent}\toprule
\thd{Model} & \thd{Arch.} & \thd{Distinct compl.} & {\thd{Repeats}} & \thd{Interpretation}\\
\arrayrulecolor{rule}\midrule
Qwen3-8B & dense & 1 & 5 & no fork in this sweep\\
\zrow Qwen3-14B & dense & 2 & 5 & dense models vary too\\
Qwen3-30B-A3B & MoE & 2 & 5 & at least the 8B control\\
\zrow gpt-oss-20b & MoE & 3.8 (3--4) & 5 & non-Qwen MoE replication\\
\arrayrulecolor{accent}\bottomrule
\end{tabularx}
\numcaptab{Repeated batch-context sweeps. The hardened distribution, not the original single-sweep 2-versus-4 count; the tested MoE models were at least as batch-sensitive as the dense controls.}
\TableNote{Distinct completions across each repeated sweep. These rows are \emph{not} a matched architecture ablation: prompt, size, family, and engine path differ. Evidence: H1, H2.}
\end{table}

\section{7.2 Interpretation}

An MoE layer routes each token to a subset of experts. A small perturbation near a top-$k$ route boundary can, in principle, select a different expert set and alter the downstream computation. This makes route selection a plausible additional sensitivity beyond the dense numerical path described in \hyperlink{chapter.6}{Chapter 6}.

The output experiments did not directly record expert IDs, router margins, or per-expert load. They therefore do not prove that the batch caused a routing change. Separate quantization forensics showed that route selection can flip under weight perturbation, which supports the general sensitivity of the router, but batching-specific route flips remain an interpretation rather than a direct E7 measurement.

\section{7.3 Scope and boundary}

The Qwen comparison is not a matched dense-versus-MoE ablation. The models differ in size, total parameters, active parameters, architecture, and serving code path. The gpt-oss replication adds a second MoE family, but it does not remove those confounds.

The robust claim is therefore limited: in the tested batch-context sweeps, the MoE models produced at least as many distinct completions as the dense controls, and one non-Qwen MoE produced more. Exact completion counts and fork positions are not stable headline quantities.

\section{7.4 Operational rule}

For reproducibility-sensitive MoE serving, record the expert-parallel configuration, routing implementation, batch policy, engine version, and placement topology. Test replay on the actual deployment rather than assuming dense-model behavior transfers.

Deterministic or replayable routing may be useful where the engine supports it, but this program did not benchmark such controls or their throughput cost. Present them as engineering options, not measured fixes.

\begin{findingbox}Repeated Qwen sweeps and a non-Qwen MoE replication show that the tested sparse models were at least as batch-sensitive as dense controls. The exact 2-versus-4 headline did not reproduce, and batching-specific expert-route changes were not directly logged. Report the repeated distribution and the architecture boundary, not one dramatic fork count.
\end{findingbox}

The next chapter changes a different execution component, the attention backend, and measures its effect on generated text.

\chapter{Attention Backends Affect Reproducibility}

\begin{figure}
\centering
\includegraphics[width=\linewidth]{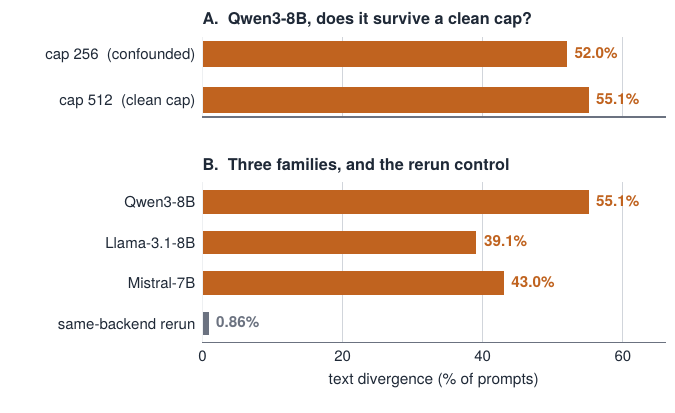}
\caption{\textbf{Backend-switch divergence and controls.} Switching FlashAttention to FlashInfer changed generated text on 52 to 55\% of Qwen3-8B GSM8K prompts, persisting when the cap was raised above the truncation knee, while same-backend reruns differed by under 1\%. Cross-family divergence was 39.1\% (Llama) and 43.0\% (Mistral). Source: E11, H1, H2.}
\end{figure}

\medskip{}

Attention backends implement the same mathematical attention operation through different kernels and scheduling paths. The experiment in this chapter asks whether switching that implementation changes generated text when the model, prompt set, decoding mode, and other configuration fields are held fixed.

\section{8.1 Controlled result}

FlashAttention is an exact, IO-aware attention algorithm; Flash\-Attention-2 changes parallelism and work partitioning; and FlashInfer is a serving-oriented attention-kernel library \autocite{dao2022flash,dao2023flash2,ye2025flashinfer}. These sources identify the compared implementation families. The 39--55\% output-divergence result is measured evidence from E11/H1/H2, not a result reported by those papers.

E11 evaluated Qwen3-8B on 256 GSM8K prompts under FlashAttention and FlashInfer in vLLM 0.11.0 with greedy decoding. At a generation cap of 256 tokens, the two backends produced different text on 52.0\% of prompts. The corresponding accuracies, 53.1\% and 54.7\%, were not interpretable as normal model quality because the cap was below the task's truncation knee.

The clean-budget rerun raised the cap to 512 tokens. Text divergence remained 55.1\%, while accuracy rose to 90.2\% under FlashAttention and 92.6\% under FlashInfer. Raising the cap above the knee therefore removed the truncation artifact without removing the backend-associated text divergence.

H1 added the most important attribution control: repeated runs under the same backend differed by only 0.86\% $\pm$ 0.16\%. The large FlashAttention-versus-FlashInfer difference is therefore associated with the backend switch rather than ordinary same-path rerun variation in that setup.

Cross-family replication produced the same qualitative pattern: 39.1\% divergence on Llama-3.1-8B and 43.0\% on Mistral-7B. The magnitude varied by model, but the effect appeared in all three tested families.

\begin{table}[htbp]\centering\TableFontNine\TableRowAir
\begin{tabularx}{\linewidth}{@{}Q{1.16in} c c >{\RaggedRight\arraybackslash}X@{}}
\arrayrulecolor{accent}\toprule
\thd{Comparison} & \thd{Text div.} & \thd{Score, two paths} & \thd{Interpretation}\\[-4pt]
 & \thu{percent} & \thu{percent} & \\
\arrayrulecolor{rule}\midrule
Qwen3-8B, cap 256 & 52.0 & 53.1 / 54.7 & cap-confounded accuracy\\
\zrow Qwen3-8B, cap 512 & 55.1 & 90.2 / 92.6 & persisted above the knee\\
Qwen3-8B reruns & \pmv{0.86}{0.16} & not reported & rerun noise is far smaller\\
\zrow Llama-3.1-8B & 39.1 & not reported & second-family replication\\
Mistral-7B & 43.0 & not reported & third-family replication\\
\arrayrulecolor{accent}\bottomrule
\end{tabularx}
\numcaptab{Backend-switch evidence and controls, FlashAttention versus FlashInfer. The same-backend rerun control is the cleanest evidence that the switch, not rerun noise, drove the divergence.}
\TableNote{GSM8K, greedy decoding, temperature zero, vLLM 0.11.0. The Qwen and H2 cross-family comparisons used 256 prompts. Qwen was measured at caps 256 and 512 as shown; the Llama and Mistral replications used the original cap-256 protocol. The same-backend control used five repeats. Only the two Qwen backend-switch rows have reported score pairs. Evidence: E11, H1, H2.}
\end{table}

\section{8.2 Interpretation}

Different attention backends can choose different tiling strategies, reduction orders, and accumulation paths. These differences provide a plausible explanation for the observed output divergence. The experiment did not inspect internal tensors or identify the exact operation that first differed.

The median first visible divergence between the two Qwen output strings occurred around character 224. That measurement locates where the generated text first separated. It does not establish that numerical error grew gradually to that point.

\begin{figure}
\centering
\includegraphics{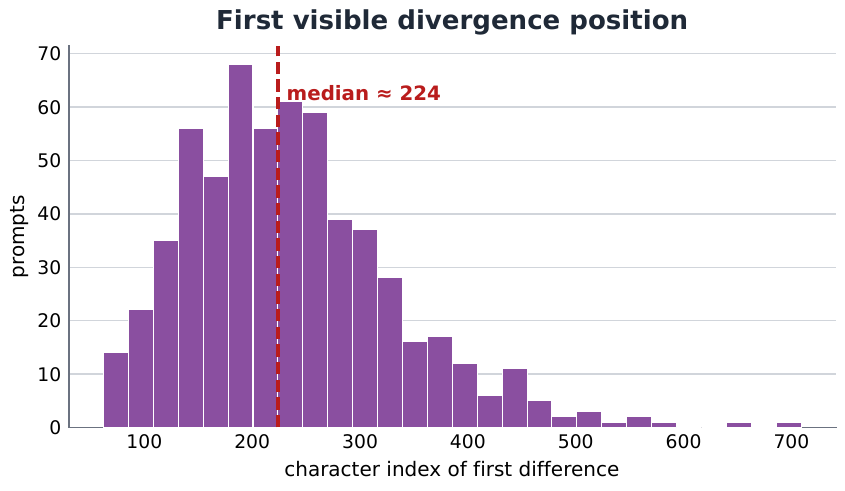}
\caption{\textbf{First visible divergence position.} For the Qwen backend comparison, the median first differing character occurred near 224. This locates the visible fork in the generated strings; it does not measure the internal growth of numerical error. Source: E11.}
\end{figure}

\section{8.3 Reproducibility and quality are separate questions}

The backend switch clearly affected byte-level reproducibility. The clean-budget run also produced a 2.3 percentage-point accuracy difference on one 256-item slice. That difference should be reported, but it does not by itself establish a stable backend quality ranking. A ranking claim would require repeated task evaluation with uncertainty and preferably more than one task.

The relevant conclusion is narrower: a result cannot be reproduced byte for byte unless the backend and engine version are part of the recorded configuration. A backend change should also trigger a paired task-impact test, because text divergence alone does not say whether the task outcome improved, worsened, or remained stable.

\section{8.4 Scope and boundary}

The result covers two attention backends, three model families, one engine version, and a math-reasoning dataset. It should not be presented as a universal 39 to 55\% law. Backend implementations and numerical behavior change across engine versions, precisions, hardware, and sequence shapes. The strongest general statement supported here is that attention-backend choice affected generated text in every tested family, with model-specific magnitude.

\section{8.5 Operational rule}

Record the exact attention backend and engine version with every reproducibility-sensitive result. Hold them fixed for byte-level comparisons and include a same-backend rerun control to distinguish backend effects from ordinary rerun variation.

When a backend changes, evaluate both text divergence and task impact. Do not use a truncated accuracy result to characterize backend quality, and do not infer harmlessness from text divergence alone.

\begin{findingbox}FlashAttention and FlashInfer produced different text on 39 to 55\% of prompts across the three tested families, while the Qwen same-backend control was below 1\%. The effect persisted when the Qwen generation cap was raised above the truncation knee. Backend and engine version belong in the execution manifest; quality impact requires a separate repeated evaluation.
\end{findingbox}

The next chapter changes the cache path rather than the kernel, and finds a similar text effect alongside a workload-dependent speedup.

\chapter{Prefix Caching Can Change Generated Text}

\begin{figure}
\centering
\includegraphics[width=\linewidth]{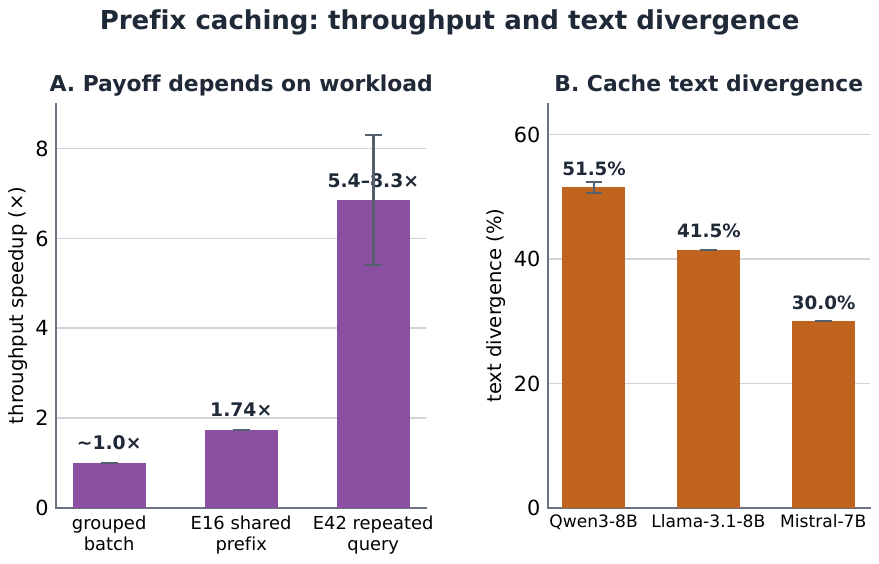}
\caption{\textbf{Two separate measurements for prefix caching.} The throughput benefit ranged from about 1.0$\times$ to more than 5$\times$ depending on the reuse pattern, and enabling the cache changed generated text on 30 to 52\% of prompts across three families. Performance and reproducibility are distinct properties. Source: E16, E42, H1, H2.}
\end{figure}

\medskip{}

Prefix caching avoids recomputing attention state for a prefix that appears again. It is useful when requests share a system prompt, few-shot examples, retrieved context, or a growing conversation history. In vLLM 0.11, automatic prefix caching (APC) reuses key--value (KV) state when new requests share a prefix with cached requests \autocite{vllm2025apc011}; the cache-associated output divergence and E41 task impact reported here are internal evidence, not claims made by that documentation. Two measurements are required: how much computation the cache saves, and whether the cache-on path produces the same generated text as the cache-off path.

\section{9.1 Performance depends on the reuse pattern}

The measured payoff varied substantially with workload shape.

In the earlier reinforcement-learning (RL) rollout study, enabling automatic prefix caching added almost no benefit when identical prompts were already processed together in one grouped batch; the scheduler had already amortized much of the shared prefill. The same study measured a 1.40$\times$ throughput gain when a roughly 1.5k-token prefix was reused across sequential batches.

E16 used a general shared-prefix workload with Qwen3-8B in vLLM 0.11.0. Throughput rose from approximately 6{,}337 to 11{,}032 tokens per second, a 1.74$\times$ gain.

E42 used a different repeated-query workload: a roughly 3k-token shared prefix reused across 40 queries in vLLM 0.23. Prefix caching reduced average time per query by 5.4 to 8.3 times.

The sweep also varied how much of the GPU's memory the engine was allowed to hold for weights and cached attention state, from 90\% down to 25\%. That matters here because a smaller allowance leaves less room for cached prefixes, so it is the setting most likely to make caching stop paying. vLLM exposes the allowance as one fraction between 0 and 1 \autocite{vllm2025engineargs}. No cache-negative crossover appeared at any of the four tested values in that single-node GPU-resident setup. CPU or NVMe offload behavior was not tested because the intended offload stack was unavailable.

\begin{table}[htbp]\centering\TableFontNine\TableRowAir
\TableTitle{Prefix-cache benefit varies by workload}
\begin{tabularx}{\linewidth}{@{}Q{1.15in} >{\RaggedRight\arraybackslash}X c Q{0.85in}@{}}
\arrayrulecolor{accent}\toprule
\thd{Workload} & \thd{Prefix condition} & \thd{Speedup} & \thd{Engine}\\[-4pt]
 & & \thu{ratio} & \\
\arrayrulecolor{rule}\midrule
Identical prompts & shared inside one batch & 1.00 & grouped batch\\
\zrow Across batches & \textasciitilde1.5k-token prefix & 1.40 & RL rollout study\\
E16 shared prefix & long prefix, 200 prompts & 1.74 & vLLM 0.11.0\\
\zrow E42 repeated query & \textasciitilde3k prefix, 40 queries & 5.4--8.3 & vLLM 0.23\\
\arrayrulecolor{accent}\bottomrule
\end{tabularx}
\numcaptab{Prefix-cache benefit varies by workload. Identical prompts already sharing one batch gained nothing, while a 3k-token prefix reused across 40 queries gained 5.4 to 8.3 times. These rows are different workloads, not a controlled speedup curve; they should not be averaged.}
\TableNote{Speedup is the measured throughput or per-query ratio from automatic prefix caching; all rows are Qwen3-8B. Different workloads and engine versions. Evidence: E16, E42, RL rollout study.}
\end{table}

\section{9.2 Text divergence under cache reuse}

The E16 cache-on and cache-off paths also produced different generated text. The hardened Qwen3-8B estimate was 51.5\% $\pm$ 0.9\% across five repeats. Cross-family replication found 41.5\% divergence on Llama-3.1-8B and 30.0\% on Mistral-7B.

\begin{table}[htbp]\centering\TableFontNine\TableRowAir
\TableTitle{Cache-on versus cache-off text divergence}
\begin{tabularx}{\linewidth}{@{}Q{1.02in} c Q{0.92in} >{\RaggedRight\arraybackslash}X@{}}
\arrayrulecolor{accent}\toprule
\thd{Model} & \thd{Text div.} & \thd{Repetition} & \thd{Interpretation}\\[-4pt]
 & \thu{percent} & & \\
\arrayrulecolor{rule}\midrule
Qwen3-8B & \pmv{51.5}{0.9} & five repeats & hardened estimate\\
\zrow Llama-3.1-8B & 41.5 & single run & present in second family\\
Mistral-7B & 30.0 & single run & present in third family\\
\arrayrulecolor{accent}\bottomrule
\end{tabularx}
\numcaptab{Cache-on versus cache-off text divergence. Enabling the cache changed generated text in all three tested families; only the Qwen row carries repeated-run uncertainty.}
\TableNote{Shared-prefix GSM8K workload, n = 200 per comparison, generation cap = 128, greedy decoding, vLLM 0.11.0. Only the Qwen row was repeated five times; the Llama and Mistral rows are single-run replications. Evidence: E16, H1, H2.}
\end{table}

Enabling APC changes the execution path. The observed text divergence shows that the cache-on and cache-off paths were not byte-equivalent. The experiment did not compare internal KV tensors or isolate whether the first mismatch came from cached values, block scheduling, attention-kernel choice, or another implementation detail. The internal explanation should therefore remain bounded.

\begin{table}[htbp]\centering\TableFontNine\TableRowAir
\TableTitle{No cache-negative crossover in the tested sweep}
\begin{tabularx}{\linewidth}{@{}C C C C@{}}
\arrayrulecolor{accent}\toprule
\thd{GPU memory} & \thd{Cache on} & \thd{Cache off} & \thd{Speedup}\\[-4pt]
\thu{fraction} & \thu{ms / query} & \thu{ms / query} & \thu{ratio}\\
\arrayrulecolor{rule}\midrule
0.90 & 14.7 & \phantom{0}84.1 & 5.70\\
\zrow 0.55 & 15.6 & \phantom{0}84.1 & 5.40\\
0.35 & 15.8 & \phantom{0}89.2 & 5.65\\
\zrow 0.25 & 15.5 & 129.1 & 8.34\\
\arrayrulecolor{accent}\bottomrule
\end{tabularx}
\numcaptab{No cache-negative crossover in the tested GPU-resident sweep. On a roughly 3k-token prefix reused across 40 queries, the cache was faster at every tested memory setting, and the observed speedup was largest at the lowest setting.}
\TableNote{GPU memory is the share of device memory the engine may hold, set through vLLM's \code{gpu\_memory\_utilization} engine argument \autocite{vllm2025engineargs}. Qwen3-8B, roughly 3k-token shared prefix, 40 queries, five repeats, vLLM 0.23, single-node GPU-resident APC. Offload-tier behavior untested. Evidence: E42.}
\end{table}

\section{9.3 Scope and boundary}

Text divergence does not establish that task quality decreased. It establishes a reproducibility difference. \hyperlink{chapter.10}{Chapter 10} measures the task impact for one setting: cache on versus off on GSM8K.

The speedup is also not universal. Prefix length, number of reuses, whether reuse occurs within or across batches, cache residency, engine version, and offload tier all affect the result. The strongest operational conclusion is conditional: APC is valuable when the deployed workload contains material repeated-prefix reuse, and its correctness impact should be tested on the served task.

\section{9.4 Operational rule}

Measure actual prefix reuse before estimating benefit. Report the engine version, prefix length distribution, cache residency policy, and whether reuse occurs within a batch or across requests.

For exact replay, record and hold fixed the cache state and policy. For task-level deployment, perform a paired cache-on/cache-off evaluation using the same items and report both text divergence and task-impact transitions.

\begin{findingbox}Prefix caching ranged from nearly neutral to more than 5$\times$ beneficial across the measured workloads, depending on how and when prefixes were reused. It also changed generated text on 30 to 52\% of prompts across three tested families. Treat APC as both a workload-specific performance feature and a reproducibility setting, and assess task impact separately.
\end{findingbox}

Text divergence on its own does not answer the question that matters for deployment: did the task outcome change? The next chapter measures that directly.

\chapter{Measure Task Impact Separately from Text Divergence}

\begin{figure}
\centering
\includegraphics{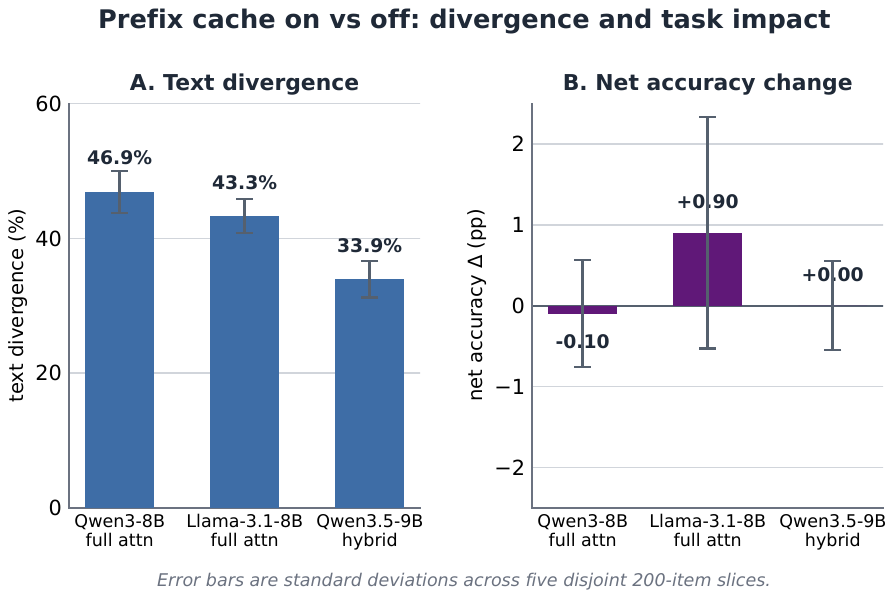}
\caption{\textbf{Text divergence and task-score impact.} For the cache-on/cache-off comparison, text differed on 34 to 47\% of prompts (mean $\pm$ SD across five slices), while the mean accuracy change was small relative to slice-to-slice variation. The two panels measure different quantities. Source: E41.}
\end{figure}

\medskip{}

A changed output string and a changed task outcome are different events. Text divergence asks whether two generated sequences are identical. Task impact asks whether the changed sequence alters correctness, utility, safety, or another predeclared outcome metric. Both measurements can matter, but one cannot substitute for the other.

This chapter applies that distinction to prefix caching. It does not establish the impact of batch-shape or attention-backend changes; those require separate paired studies.

\section{10.1 Controlled result}

E41 compared prefix caching off and on for the same GSM8K items under greedy decoding. Cache-off was the \emph{reference execution condition}, not the ground truth; correctness was determined by the GSM8K reference answer.

The hardened experiment used five disjoint 200-question slices and three model setups: Qwen3-8B, Llama-3.1-8B, and Qwen3.5-9B. For each slice it measured generated-text divergence, net accuracy change (cache on minus cache off), and the correct-to-wrong transition rate. The reported values are means $\pm$ standard deviations across the five slices.

\begin{table}[htbp]\centering\TableFontNine\TableRowAir
\TableTitle{Cache-on versus cache-off impact metrics}
\begin{tabularx}{\linewidth}{@{}>{\hsize=1.24\hsize\RaggedRight\arraybackslash}X >{\hsize=0.92\hsize\centering\arraybackslash}X >{\hsize=0.92\hsize\centering\arraybackslash}X >{\hsize=0.92\hsize\centering\arraybackslash}X@{}}
\arrayrulecolor{accent}\toprule
\thd{Model setup} & \thd{Text div.} & \thd{Accuracy $\Delta$} & \thd{\mbox{C\,$\to$\,W}}\\[-4pt]
 & \thu{percent} & \thu{pp} & \thu{percent}\\
\arrayrulecolor{rule}\midrule
Qwen3-8B, full & \pmv{46.9}{3.1} & \pmv{−0.10}{0.66} & \pmv{1.5}{0.7}\\
\zrow Llama-3.1-8B, full & \pmv{43.3}{2.5} & \pmv{+0.90}{1.43} & \pmv{2.0}{1.1}\\
Qwen3.5-9B, hybrid & \pmv{33.9}{2.7} & \pmv{+0.00}{0.55} & \pmv{1.1}{0.5}\\
\arrayrulecolor{accent}\bottomrule
\end{tabularx}
\numcaptab{Cache-on versus cache-off impact metrics, mean and standard deviation across five slices. Text divergence was substantial while the mean accuracy change stayed small relative to slice-to-slice variation. \mbox{C\,$\to$\,W} is the \mbox{correct-to-wrong} transition rate.}
\TableNote{Five disjoint GSM8K-200 slices, APC off versus on, greedy decoding, vLLM 0.23. Transition rates use all 200 paired items in each slice; cache-off is the reference execution path, while GSM8K answers determine correctness. Evidence: E41.}
\end{table}

The mean accuracy changes were small relative to slice-to-slice variation, and no model showed a consistent negative shift. At the same time, correct-to-wrong transitions were not zero: they occurred in approximately 1.1 to 2.0\% of items as currently reported. A near-zero net delta therefore reflects a balance of harmful and beneficial transitions, not the absence of per-item changes.

Because the transition denominator is all paired items, the mean wrong-to-correct rate can be recovered from the printed means as correct-to-wrong plus the net accuracy change: 1.4\% for Qwen3-8B, 2.9\% for Llama-3.1-8B, and 1.1\% for Qwen3.5-9B. Those three values are derived means; the retained summary does not contain their \mbox{per-slice} standard deviations.

\section{10.2 Use the paired transition matrix}

A complete impact analysis should report four paired outcomes rather than a single ``answers wrong'' figure.

\begin{table}[htbp]\centering\TableFontNine\TableRowAir
\TableTitle{The paired transition matrix}
\begin{tabularx}{\linewidth}{@{}Q{1.10in} Q{1.10in} >{\RaggedRight\arraybackslash}X@{}}
\arrayrulecolor{accent}\toprule
\thd{Reference path} & \thd{Changed path} & \thd{Interpretation}\\
\arrayrulecolor{rule}\midrule
correct & correct & stable correct result\\
\zrow correct & wrong & harmful transition\\
wrong & correct & beneficial transition\\
\zrow wrong & wrong & stable error\\
\arrayrulecolor{accent}\bottomrule
\end{tabularx}
\numcaptab{The paired transition matrix. Reporting all four cells is more informative than a single wrong-answer bar, and text divergence belongs in a separate column: a correct-to-correct item may still be worded differently.}
\TableNote{Identical text can be wrong under both paths, and divergent text can be correct under both.}
\end{table}

The E41 results support a bounded conclusion: for the tested cache-on/cache-off GSM8K setup, substantial text divergence did not produce a consistent net accuracy loss. They do not establish that every divergent output was semantically equivalent, and they do not imply that a 1 to 2\% harmful transition rate is acceptable for every application.

\section{10.3 Architecture scope}

The tested Qwen3.5-9B row had the lowest raw divergence and raw correct-to-wrong rate. That observation is useful, but it is not a controlled proof that hybrid attention caches more cleanly than full attention. The rows differ in model family, baseline accuracy, prompt fit, and architecture. The initial E41 note also reported poor Qwen3.5 baseline accuracy under the shared-prefix prompt.

The defensible statement is narrower: the tested Qwen3.5-9B setup did not show a large cache-induced accuracy regression on GSM8K. A general architecture claim would require matched models or a broader controlled study, plus baseline and conditional transition rates.

\section{10.4 Operational rule}

When an optimization changes generated text, run a paired impact study on the served task. Predeclare the acceptance criteria and report the text-divergence rate; baseline and changed-condition task scores; net score delta with uncertainty; correct-to-wrong and wrong-to-correct transitions; and whether byte-exact replay is a separate requirement.

For prefix caching, the available results support enabling APC on similar repeated-prefix workloads when the paired task-impact metrics remain within the application's declared limits. This conclusion is cache- and task-specific. Batch-context and backend changes should be evaluated with the same method before being called benign.

\begin{figure}[htbp]
\centering
\includegraphics[width=\linewidth]{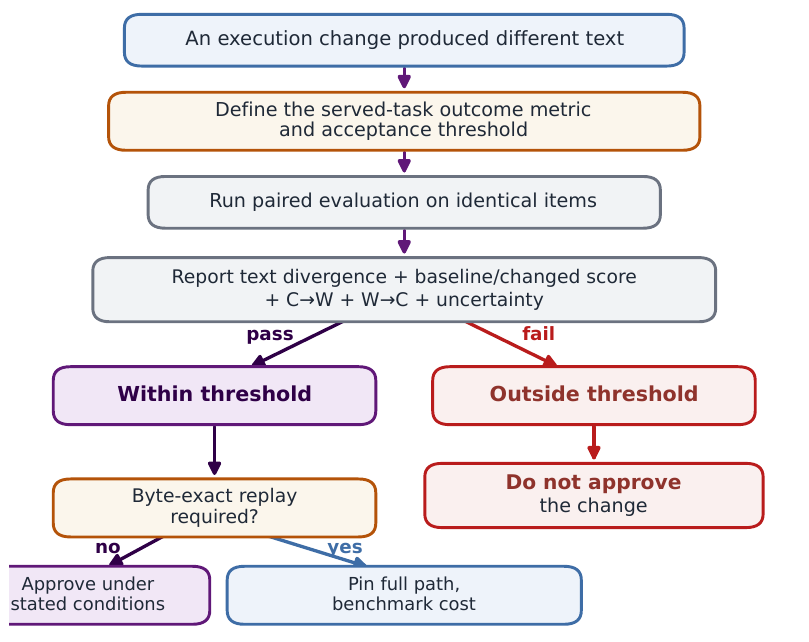}
\caption{\textbf{Paired-impact acceptance workflow.} When an execution change alters generated text, define the served-task metric and a predeclared threshold, run a paired evaluation on identical items, and report divergence, scores, and transitions before deciding. A decision workflow, not a measured result.}
\end{figure}

\begin{findingbox}E41 found 34 to 47\% cache-path text divergence but only small mean accuracy changes relative to slice variation on GSM8K. Harmful per-item transitions still occurred, so the correct conclusion is not ``nothing changed.'' The reusable method is to separate text divergence, net task impact, and paired correct/wrong transitions, then apply a task-specific acceptance threshold.
\end{findingbox}

\chapter{Define the Reproducibility Contract}
\label{chapter:repro-contract}

\begin{figure}
\centering
\includegraphics[width=\linewidth]{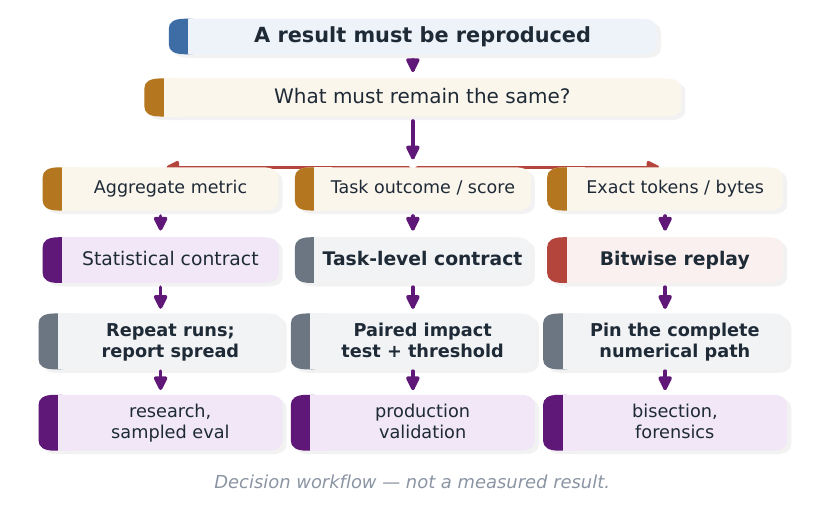}
\caption{\textbf{Three reproducibility contracts.} Statistical, task-level, and bitwise reproducibility require different equalities and different acceptance tests. Naming the required level, and its test, comes before recording the execution manifest. A decision workflow, not a measured result.}
\end{figure}

\medskip{}

Reproducibility is not a single requirement. A benchmark paper may need stable aggregate scores. A production change may need stable task outcomes. A rollout replay or bug bisection may require identical token IDs. These are different contracts, and they should not be evaluated with the same acceptance test. Broader machine-learning reproducibility work motivates complete disclosure of code, data, hyperparameters, and evaluation conditions \autocite{pineau2021repro}; the three inference-specific levels used here---statistical reproducibility, task-level reproducibility, and bitwise replay---are this book's synthesis and should not be attributed wholesale to that paper.

Part II identified several execution variables that can change generated text under greedy decoding: batch context, the tested sparse-model execution paths, attention backend, and prefix-cache state. The evidence does not support one universal conclusion about their task impact. Prefix caching received a paired GSM8K impact study; batch-context and backend changes did not receive the same repeated task-level analysis.

\section{11.1 Three levels of reproducibility}

\emph{Statistical reproducibility} requires aggregate metrics to agree within a declared interval across repeated runs, seeds, slices, or traces. It is the appropriate contract for sampled evaluation, latency distributions, and most systems measurements.

\emph{Task-level reproducibility} requires a changed execution path to remain within a task-specific acceptance threshold. The generated wording may differ while correctness, safety, utility, or another declared outcome remains stable. This is the contract evaluated by a paired impact study such as E41.

\emph{Bitwise replay} requires identical token IDs or output bytes. It is appropriate for bug bisection, forensic audits, and rollout replay when downstream state depends on the exact sequence.

\begin{table}[htbp]\centering\TableFontEightSix\TableRowAir
\TableTitle{Three reproducibility contracts}
\begin{tabularx}{\linewidth}{@{}Q{0.68in} >{\RaggedRight\arraybackslash}X >{\RaggedRight\arraybackslash}X >{\RaggedRight\arraybackslash}X@{}}
\arrayrulecolor{accent}\toprule
\thd{Contract} & \thd{Required equality} & \thd{Acceptance test} & \thd{Typical use}\\
\arrayrulecolor{rule}\midrule
Statistical & Metric distribution & Repeat in interval & research claims\\
\zrow Task-level & Score in tolerance & Paired impact test & production validation\\
Bitwise replay & Identical tokens & Exact equality & bisection, audit\\
\arrayrulecolor{accent}\bottomrule
\end{tabularx}
\numcaptab{Three reproducibility contracts. Each level has its own required equality and acceptance test, and the contract should be named before the run: ``reproducible'' is incomplete without the level and its test.}
\TableNote{Framework for Part II rather than a measured result.}
\end{table}

\section{11.2 The exact-replay manifest}

Bitwise replay requires more than a sampling seed. At minimum, the manifest should record the model checkpoint and revision or hash; the tokenizer revision and exact chat-template bytes; the raw prompt bytes and dataset item IDs; the engine, framework, CUDA, driver, and kernel-library versions; the hardware model, GPU count, tensor/pipeline/expert-parallel topology, and placement; the weight dtype, quantization method, KV-cache dtype, and relevant calibration artifacts; the attention backend and deterministic or eager-operation settings; prefix-cache enablement, cache state, reuse history, and request order; the batch policy, scheduler limits, concurrency, and co-batched request context; the decoding parameters, stop conditions, and random-number state; and the extractor or verifier when the reproduced object is a score rather than raw text.

These fields are \emph{necessary controls}, not an unconditional guarantee. Exact equality can still fail if an operation is nondeterministic or if the hardware and software environments are not equivalent. The correct claim is therefore that the manifest makes known differences visible and replay attempts diagnosable.

\section{11.3 Price only the controls that were measured}

The cost of reproducibility is not one number. Recording versions and pinning a backend may add little or no runtime cost. Disabling prefix caching removes a measured performance benefit on repeated-prefix workloads. E32 measured a 38\% throughput reduction for eager execution relative to the default CUDA-graph path in one Qwen3-8B benchmark. The cost of deterministic expert routing and batch-invariant operations was not measured in this program and should not be assigned a number here.

The operational choice should follow the required contract: use repeated runs and uncertainty for statistical reproducibility; use paired outcome metrics for task-level reproducibility; and pay the exact-replay controls only when identical tokens are required.

\section{11.4 What Part II established}

The Part II studies support four intentionally asymmetric conclusions, because the experiments were asymmetric: batch context can alter a greedy generation under the tested stack; the tested MoE models showed additional batch-context output variability, although routing changes were not directly instrumented; switching attention backends produced substantial text divergence across three tested model families; and prefix caching produced substantial text divergence, while a paired GSM8K study found no consistent net accuracy loss across the tested slices.

\begin{table}[htbp]\centering\TableFontEightSix\TableRowAir
\TableTitle{Part II in one table: text stability versus task impact}
\begin{tabularx}{\linewidth}{@{}Q{0.78in} >{\hsize=1.03\hsize\RaggedRight\arraybackslash}X >{\hsize=0.92\hsize\RaggedRight\arraybackslash}X >{\hsize=1.05\hsize\RaggedRight\arraybackslash}X@{}}
\arrayrulecolor{accent}\toprule
\thd{Execution change} & \thd{Text-stability evidence} & \thd{Task-impact evidence} & \thd{Defensible conclusion}\\
\arrayrulecolor{rule}\midrule
Batch context & Greedy-path changes possible; exact counts prompt-sensitive & \textit{Not measured in a paired study} & Belongs in exact-replay metadata; task impact open\\
\zrow Tested MoE paths & Greater batch-context variability; routes not logged & \textit{Not measured} & Replay needs added controls; routing claim bounded\\
Attention backend & 39--55\% divergence, three families; same-backend under 1\% & Both paths above 90\% on one clean slice; no repeated paired study & Report the backend; quality equivalence not established\\
\zrow Prefix cache & 34--47\% divergence, five slices, three models & Net GSM8K change near zero; \mbox{C\,$\to$\,W} 1.1--2.0\% & No consistent net loss on this task; local acceptance test still required\\
\arrayrulecolor{accent}\bottomrule
\end{tabularx}
\TableCaption{\textbf{Table II.1.} Part II in one table. Text instability was measured broadly; task impact was measured narrowly, for prefix caching only. A synthesis table must preserve the scope of the underlying experiments rather than filling missing measurements with reassuring words.}
\TableNote{Cells marked \textit{not measured} received no paired task-impact study. Evidence: E7, E11, E16, E41, H1, H2.}
\end{table}

\begin{findingbox}State the required reproducibility level before the run. Record the complete execution manifest. Use repeated uncertainty for statistical claims, paired task metrics for deployment decisions, and exact token comparison only when bitwise replay is the actual requirement.
\end{findingbox}

For the compact reproduction sequence and acceptance criteria, continue to Appendix~\ref{app:reproduction}; for the chapter-by-chapter evidence and scope map, use Appendix~\ref{app:evidence}.

%

\part{Serving Efficiency and Capacity}

\begin{partopener}
Throughput, latency, cost, and memory are different measurements.
\end{partopener}

Serving efficiency is determined by several layers of the execution system. The software path sets the attainable throughput baseline. Batch occupancy and arrival load define the cost--latency operating point. The KV-cache pool limits active-token capacity. Scheduler token budgets shape the balance between prefill and decode. These controls interact, but they should be measured separately: output throughput is not request throughput, median latency is not p99, and a derived GPU-rate estimate is not a complete infrastructure bill.

Throughout this part, TTFT means time to first token, ITL means inter-token latency, TPOT means time per output token, SLO means service-level objective, and p99 is the 99th percentile.

\begin{navbox}
{\sffamily\bfseries\footnotesize\color{ink}When a serving deployment is too slow, too costly, or unstable, where does the evidence point?}\par\vspace{4pt}
{\sffamily\footnotesize\color{ink}%
\begin{itemize}[leftmargin=1.2em,itemsep=3pt,topsep=2pt]
\item Is the benchmark using the intended production engine path? If not $\rightarrow$ establish a production-relevant baseline (\textbf{Ch.\ 12}).
\item Does achieved request rate meet offered load within the p99 SLO? If not $\rightarrow$ tune batch/concurrency and measure the cost--tail point (\textbf{Ch.\ 13}).
\item Does active-token demand fit the reported KV pool with margin? If not $\rightarrow$ reduce concurrency or context, or change memory strategy (\textbf{Ch.\ 14}).
\item Has the scheduler token budget been tuned for this model and traffic shape? If not $\rightarrow$ run a per-model TTFT/ITL/throughput sweep (\textbf{Ch.\ 15}).
\end{itemize}}
\end{navbox}

\chapter{Serving Software Determines Achievable Throughput}

\begin{figure}
\centering
\includegraphics[width=\linewidth]{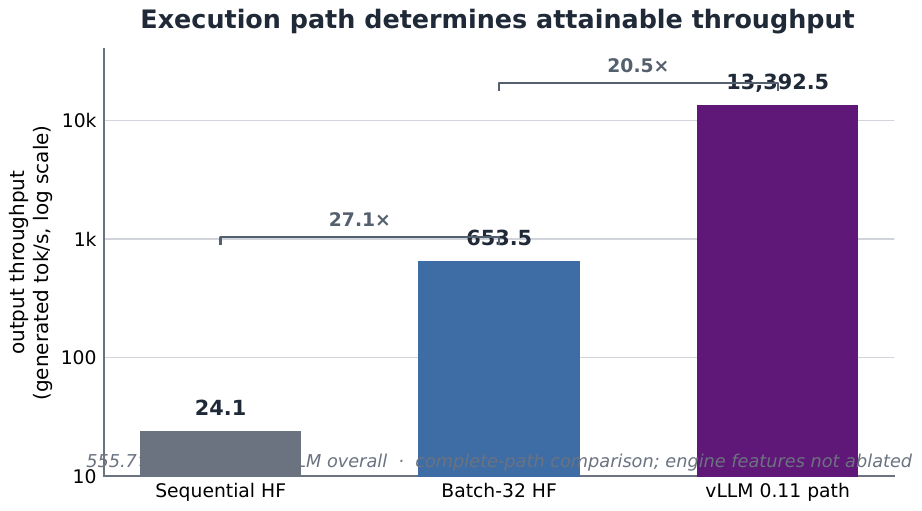}
\caption{\textbf{Output throughput across three execution paths.} The same Qwen3-8B checkpoint and 256-prompt GSM8K workload produced 24.1 tok/s (sequential HF), 653.5 tok/s (batch-32 HF), and 13{,}392.5 tok/s (vLLM 0.11). This is a complete-path comparison; individual engine features were not ablated. Source: measured.}
\end{figure}

\medskip{}

A throughput result describes a model \emph{and} an execution path. The same checkpoint and prompt set can produce very different throughput when requests are processed sequentially, in a fixed batch, or by a serving engine with continuous scheduling and specialized memory management.

The experiment in this chapter compares three complete execution paths. It does not isolate any one engine feature.

\section{12.1 Controlled comparison}

Qwen3-8B processed 256 GSM8K prompts on one H100 under greedy decoding with a 256-token generation cap. This table answers how much the measured throughput changed when the complete execution path changed while the model, hardware, prompt set, and decoding mode were held fixed.

\begin{table}[htbp]\centering\TableFontNine\TableRowAir
\TableTitle{Throughput across three execution paths}
\begin{tabularx}{\linewidth}{@{}Q{1.02in} >{\RaggedRight\arraybackslash}X S[table-format=5.1] S[table-format=3.1]@{}}
\arrayrulecolor{accent}\toprule
\thd{Execution path} & \thd{Batching policy} & {\thd{Output}} & {\thd{vs sequential}}\\[-4pt]
 & & {\thu{tok/s}} & {\thu{ratio}}\\
\arrayrulecolor{rule}\midrule
Hugging Face & one request at a time & 24.1 & 1.0\\
\zrow Hugging Face & fixed batch of 32 & 653.5 & 27.1\\
vLLM 0.11 & engine-managed batch & 13392.5 & 555.7\\
\arrayrulecolor{accent}\bottomrule
\end{tabularx}
\numcaptab{Throughput across three execution paths. One model, one workload; only the execution path and batching policy changed. The vLLM path was also 20.5 times the batched Hugging Face path. This is a serving-stack measurement, not a model-speed measurement in isolation.}
\TableNote{Qwen3-8B, one H100-80GB, 256 GSM8K prompts, greedy decoding, generation cap 256 tokens. The retained study notes identify vLLM 0.11.0 and the Hugging Face \code{generate} path but do not preserve the exact Transformers revision or utilization-sampling method; the comparison is therefore scoped to the reported complete paths. Evidence: E13.}
\end{table}

\section{12.2 Why concurrency changes device use}

Sequential generation exposes one active sequence at a time. That provides limited opportunity to amortize model-state movement and occupy parallel compute resources. Fixed batching increases the amount of work performed together and produced a 27.1$\times$ gain in this experiment.

The vLLM path improved throughput by a further 20.5$\times$ relative to the batch-32 Hugging Face path. Several components changed together, including scheduling, request packing, KV-cache management, kernel execution, and engine overhead. PagedAttention manages KV memory in non-contiguous blocks, while iteration-level scheduling allows requests at different generation stages to share execution \autocite{kwon2023paged,yu2022orca}. E13 compares complete execution stacks; it does not isolate either mechanism's individual contribution to the measured throughput gap. Paged KV management and continuous scheduling are plausible contributors, but this experiment did not measure their effects separately. The defensible statement is that the complete serving-engine path substantially outperformed the tested framework baselines under this workload; a feature-level claim would require separate ablations with the same engine and all other settings fixed.

\begin{figure}
\centering
\includegraphics[width=\linewidth]{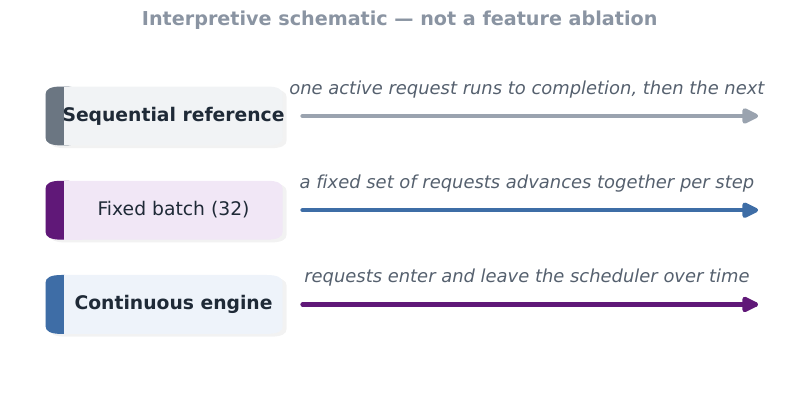}
\caption{\textbf{Interpretive schematic.} Sequential processing exposes one request at a time; fixed batching runs a group together; a continuous engine admits and retires requests over time. A schematic of why occupancy differs, not a measured ablation of engine features.}
\end{figure}

\section{12.3 Use each baseline for the question it answers}

The sequential path is useful as a reference and as a diagnostic of underutilization. The batch-32 path is a stronger production-oriented baseline because it removes the most obvious one-request bottleneck. The engine comparison then shows how much additional improvement remained after manual batching.

Reporting all three values is more informative than declaring one baseline legitimate and another illegitimate. The sequential ratio describes the full distance from a reference implementation; the batched ratio describes the distance from a basic optimized baseline. A precise throughput claim should also identify whether a tokens-per-second figure counts generated tokens only or total processed tokens.

\section{12.4 Operational rule}

For a production-throughput claim: benchmark the intended production engine; use a prompt and output-length distribution representative of the service; report batch or concurrency policy, request count, and arrival pattern; separate output throughput from request throughput and latency; and include a framework baseline only when its batching and measurement protocol are explicit.

Framework generation remains appropriate for reference outputs, debugging, small offline jobs, and correctness checks. It should not be used as the sole basis for production capacity planning.

\begin{findingbox}On the tested Qwen3-8B workload, manual batch-32 Hugging Face generation was 27.1$\times$ faster than the sequential reference, and the vLLM path was a further 20.5$\times$ faster. The experiment compares complete execution stacks, not isolated engine features. Report production throughput with the engine, batching policy, workload distribution, and measurement protocol.
\end{findingbox}

The next chapter holds the engine fixed and varies batch occupancy and arrival load to locate the cost--latency operating point.

\chapter{Batch Occupancy Sets the Cost--Latency Operating Point}

\begin{figure}
\centering
\includegraphics{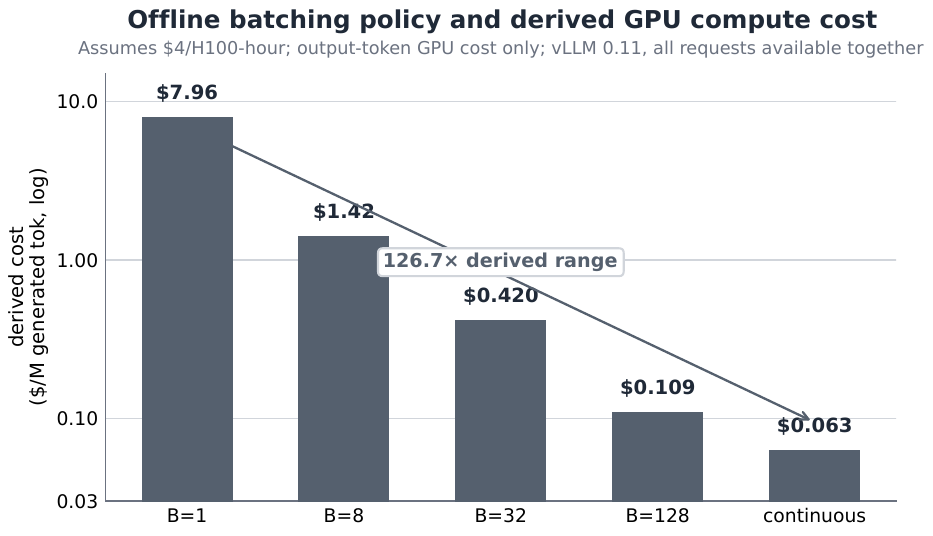}
\caption{\textbf{Offline batching policy and derived GPU compute cost.} On one fixed Qwen3-8B/H100 workload, derived GPU cost per million output tokens fell 126.7$\times$ from static B=1 to continuous batching. Assumes \$4/H100-hour, output-token GPU cost only; all requests were available together. Source: E14.}
\end{figure}

\medskip{}

Serving efficiency and online latency are related but distinct measurements. An offline throughput sweep can estimate GPU compute cost. An online arrival trace can characterize achieved rate and the latency tail. They should be reported together only when their engine, workload, and arrival conditions are identical. This chapter presents three separate results: an offline batching sweep, an online TTFT study, and a model-selection comparison based on cost per correct answer.

\section{13.1 Offline throughput and derived GPU compute cost}

E14 ran Qwen3-8B on one H100 under several batching policies, using 512 prompts, greedy decoding, and a 256-token cap in vLLM 0.11. The dollar values assume \$4 per H100-hour and are derived from measured output throughput: \$/M output tokens $=$ (\$4 / 3600\,s) / (output tok/s) $\times$ 1{,}000{,}000.

\begin{table}[htbp]\centering\TableFontNine\TableRowAir
\TableTitle{Offline batching sweep}
\begin{tabularx}{\linewidth}{@{}>{\RaggedRight\arraybackslash}X C C C C@{}}
\arrayrulecolor{accent}\toprule
\thd{Serving policy} & \thd{Output} & \thd{Cost} & \thd{Relative} & \thd{Latency*}\\[-4pt]
 & \thu{tok/s} & \thu{\$/M out tok} & \thu{ratio} & \thu{s}\\
\arrayrulecolor{rule}\midrule
Static B=1   & 139.6    & 7.96  & 126.7 & 1.83\\
\zrow Static B=8   & 947      & 1.17  & 18.6  & 1.92\\
Static B=32  & 3{,}377  & 0.329 & 5.2   & 2.13\\
\zrow Static B=128 & 10{,}230 & 0.109 & 1.7   & 2.82\\
Continuous   & 17{,}686 & 0.063 & 1.0   & 6.54\\
\arrayrulecolor{accent}\bottomrule
\end{tabularx}
\numcaptab{Offline batching sweep. Within one fixed model and hardware, batching policy changed derived GPU compute cost by 126.7 times. This is derived GPU compute cost, not a complete infrastructure bill.}
\TableNote{All requests arrived together, so the offline latency column is not an online p99 measure; the retained E14 summary does not specify a more granular latency definition. Static B=1 and B=8 used rate-based subsamples of 32 and 64 prompts. Qwen3-8B, one H100, vLLM 0.11, greedy, 256-token cap, \$4/H100-hour. Evidence: E14.}
\end{table}

Within this fixed-model experiment, batching policy changed derived GPU compute cost by 126.7$\times$. The result does not imply that batching dominates every serving bill across model choices, providers, or workloads; it demonstrates that low occupancy can dominate cost even when the model and hardware remain unchanged. The derived values exclude non-GPU resources, idle capacity, failures, networking, storage, and orchestration.

\section{13.2 Online TTFT under Poisson arrivals}

A separate H5 experiment used vLLM 0.23 and random requests with 256 input tokens and 128 output tokens. Two hundred requests were generated at each Poisson offered rate.

\begin{table}[htbp]\centering\TableFontEightSix\TableRowAir
\TableTitle{Online arrival-load study}
\begin{tabularx}{\linewidth}{@{}C C C C C C@{}}
\arrayrulecolor{accent}\toprule
\thd{Offered} & \thd{Achieved} & \thd{Served} & \thd{Median} & \thd{p99} & \thd{p99 : median}\\[-4pt]
\thu{req/s} & \thu{req/s} & \thu{percent} & \thu{TTFT, ms} & \thu{TTFT, ms} & \thu{ratio}\\
\arrayrulecolor{rule}\midrule
4  & 3.77  & 94 & 67.2 & 79.7  & 1.19\\
\zrow 16 & 12.79 & 80 & 73.8 & 88.2  & 1.20\\
64 & 31.46 & 49 & 84.0 & 123.8 & 1.47\\
\arrayrulecolor{accent}\bottomrule
\end{tabularx}
\numcaptab{Online arrival-load study. As offered load rose, achieved rate fell below offered and the p99 TTFT tail widened. A single completed trace per rate; no uncertainty interval was recorded.}
\TableNote{Qwen3-8B, vLLM 0.23, random 256-input/128-output workload, 200 requests per rate, Poisson arrivals. These online load--tail values are a different stack and workload from Table 13.1. Evidence: H5.}
\end{table}

Under this workload, achieved rate fell increasingly below offered rate, and the p99 TTFT separated further from the median at the highest tested load. These online load--tail measurements should not be presented as the directly measured latency price of the E14 cost cells, because the two experiments used different stacks and workloads. A unified operating frontier would rerun batching or concurrency settings under one engine, one arrival trace, and one prompt/output distribution while measuring throughput, achieved rate, TTFT, ITL, latency, and derived cost together.

\begin{figure}
\centering
\includegraphics[width=\linewidth]{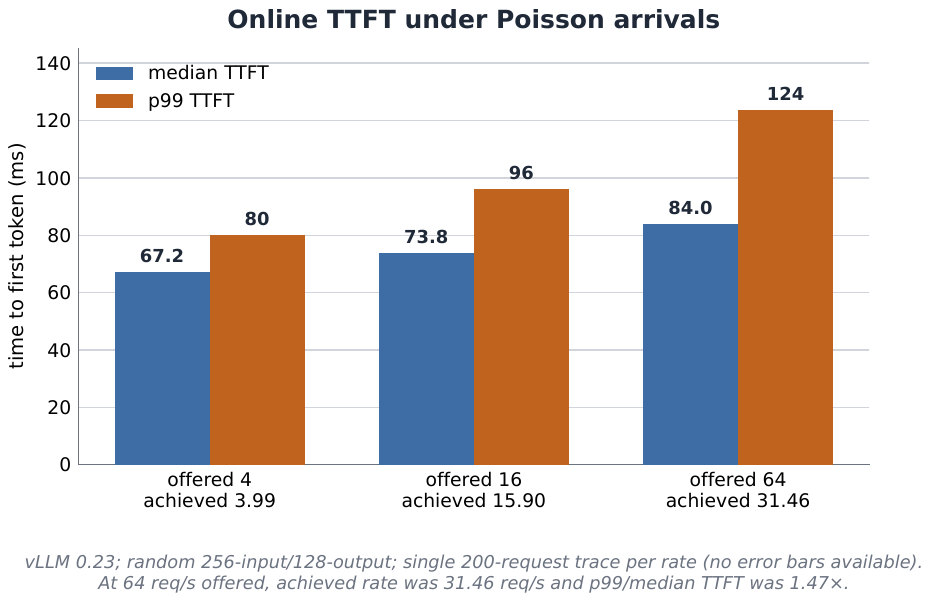}
\caption{\textbf{Online TTFT under Poisson arrivals.} At 64 req/s offered, achieved rate was 31.46 req/s and p99/median TTFT was 1.47$\times$. A single trace per rate; no error bars available. Conditions: Qwen3-8B, vLLM 0.23, random 256/128 workload. Source: H5.}
\end{figure}

\section{13.3 Select an operating point against the SLO}

The appropriate concurrency and scheduling policy depends on both offered load and the service-level objective. A configuration is not efficient merely because output throughput is high; it must also sustain the offered request rate and remain within the declared tail-latency limit. A production study should report offered and achieved request rate; output throughput; median and p99 TTFT; median and p99 end-to-end latency; prompt and output-length distributions; queueing or rejection behavior; and the cost equation with its included resources. The operating point is the highest-load configuration that satisfies the workload's acceptance criteria, not necessarily the cell with the highest tok/s.

\section{13.4 Model choice on cost per correct answer}

A separate E6c comparison shows why throughput and accuracy should be combined. On GSM8K, the tested Qwen3-8B path had the lowest derived cost per correct answer, while Qwen3-14B had the highest measured accuracy.

\begin{table}[htbp]\centering\TableFontNine\TableRowAir
\TableTitle{Accuracy alone does not determine cost per correct answer}
\begin{tabularx}{\linewidth}{@{}>{\hsize=0.94\hsize\RaggedRight\arraybackslash}X C C C >{\hsize=1.06\hsize\centering\arraybackslash}X@{}}
\arrayrulecolor{accent}\toprule
\thd{Model} & \thd{Accuracy} & \thd{Tokens} & \thd{Output} & \thd{Cost}\\[-4pt]
 & \thu{percent} & \thu{per correct} & \thu{tok/s} & \thu{\$ per 1,000 correct}\\
\arrayrulecolor{rule}\midrule
Qwen3-8B  & 91.99 & 259.3 & 13{,}106 & 0.0248\\
\zrow Qwen3-14B & 94.73 & 235.0 & 8{,}750  & 0.0323\\
Qwen3-32B & 94.53 & 239.9 & 2{,}236  & 0.1293\\
\arrayrulecolor{accent}\bottomrule
\end{tabularx}
\numcaptab{Accuracy alone does not determine cost per correct answer. The most accurate model was not the cheapest per correct answer on this task: the 32B path cost 5.2 times as much per correct answer as the 8B path and did not exceed 14B accuracy.}
\TableNote{GSM8K-512, one H100 per model, vLLM 0.11, greedy, 512-token cap, \$4/H100-hour. Task- and stack-specific. Evidence: E6c.}
\end{table}

\begin{findingbox}E14 measured a 126.7$\times$ range in derived GPU compute cost across offline batching policies for one fixed model. H5 separately showed a wider TTFT tail and lower served/offered ratio at the highest Poisson load. Measure cost and tail latency in one unified online sweep, and select models on cost per accepted outcome, not accuracy or tok/s alone.
\end{findingbox}

The next chapter turns to the memory pool that ultimately bounds how many requests can run at once.

\chapter{KV-Cache Capacity Limits Concurrency}

\begin{figure}
\centering
\includegraphics[width=\linewidth]{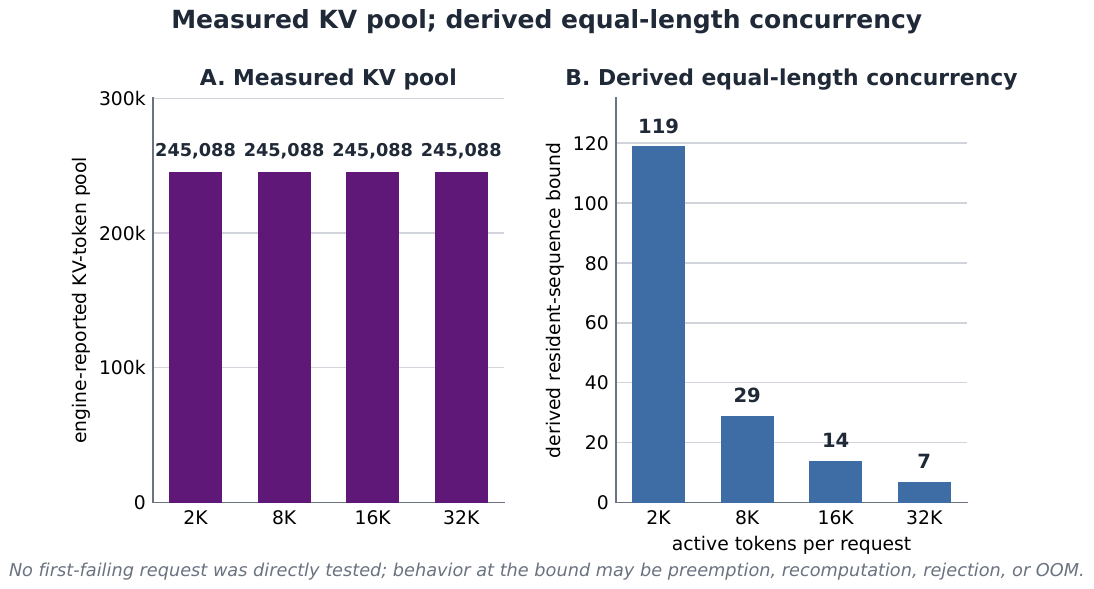}
\caption{\textbf{Measured KV pool; derived equal-length concurrency.} The engine reported a 245{,}088-token KV pool for Qwen3-14B; dividing it by equal active-token lengths yields a full-context bound of 119 at 2{,}048 tokens down to 7 at 32{,}768. The bound is derived, not a directly tested first-failing request. Source: E15 (measured pool), derived concurrency.}
\end{figure}

\medskip{}

The KV cache occupies a finite pool of GPU memory. As active context grows, fewer sequences can remain resident in that pool. This relationship is useful because the engine can often report the pool directly and the resulting capacity bound can be calculated before a load test. The experiment measured the engine-reported KV pool; it did not directly execute the first request above every derived boundary.

\section{14.1 Measured pool and derived equal-length bound}

E15 loaded Qwen3-14B in vLLM 0.11 on one H100-80GB with eager execution, allowing the engine to hold 85\% of device memory for weights and cached attention state. The context-length limit was varied across its tested settings, and at every one of them the engine reported the same KV-cache pool of 245{,}088 token slots. This table converts that measured token pool into an equal-length planning bound.

\begin{table}[htbp]\centering\TableFontNine\TableRowAir
\TableTitle{Engine-reported pool and derived full-context bound}
\begin{tabularx}{\linewidth}{@{}C C C@{}}
\arrayrulecolor{accent}\toprule
\thd{Active context} & \thd{Engine KV pool} & \thd{Derived bound}\\[-4pt]
\thu{tokens per request} & \thu{token slots} & \thu{concurrent sequences}\\
\arrayrulecolor{rule}\midrule
2{,}048  & 245{,}088 & 119\\
\zrow 8{,}192  & 245{,}088 & 29\\
16{,}384 & 245{,}088 & 14\\
\zrow 32{,}768 & 245{,}088 & 7\\
\arrayrulecolor{accent}\bottomrule
\end{tabularx}
\numcaptab{Engine-reported pool and derived full-context bound: how many equal-length active sequences fit if the entire reported pool were available to them. The bound is a conservative equal-length floor of pool divided by context, not a tested failure point.}
\TableNote{Every row is a measured pool with a derived bound. Qwen3-14B, one H100-80GB, vLLM 0.11, eager execution, engine memory share 0.85. Assumes equal-length sequences; no first-failing request was executed. Evidence: E15.}
\end{table}

\section{14.2 What happens near capacity}

When active demand approaches the pool, the engine may reduce effective concurrency through preemption and later recomputation. Admission limits may queue or reject work. Some configurations may fail with an out-of-memory error. Which behavior occurs depends on the engine version, scheduler, allocation strategy, and configured limits. E15 did not directly test the 120th 2{,}048-token request or the eighth 32{,}768-token request, so no single failure mode should be assigned to those exact positions. Preemption count, recomputation, request rejection, queue length, and OOM events are all relevant monitoring signals, and capacity pressure can appear as latency degradation before a hard failure.

\section{14.3 Estimating capacity analytically}

For a conventional full-attention cache, an approximate per-token memory cost is $2 \times \text{layers} \times \text{KV heads} \times \text{head dim} \times \text{bytes/element}$. Dividing available KV memory by that value gives an approximate token capacity. Treat it as a planning aid, not a universal formula: cache layout and memory use differ under tensor parallelism, cache quantization, hybrid attention, sliding windows, compressed KV designs, speculative decoding, and engine-specific block allocation. When available, use the engine-reported block count multiplied by block size for the actual launch configuration, and retain the analytic estimate as an independent sanity check.

\section{14.4 Plan with an active-token distribution}

Production traffic rarely consists of equal-length full-window requests. Capacity planning should use the distribution of prompt lengths, expected output growth, and admitted concurrency. A practical check is that the sum of active prompt and generated tokens for admitted requests, plus reserved headroom, stays within the reported KV-token pool. Prefix sharing may reduce repeated blocks; output growth increases demand after admission; block rounding and fragmentation reduce effective capacity. A representative load test is therefore required after the arithmetic estimate.

The share of device memory the engine is allowed to hold trades KV-pool size against headroom for activations: a higher share can increase the pool while reducing protection against transient demand, and a lower share reserves headroom while reducing resident-token capacity. Neither direction is universally correct. In vLLM this share is the \code{gpu\_memory\_utilization} engine argument \autocite{vllm2025engineargs}.

\section{14.5 Operational procedure}

Record the engine-reported KV block count and block size; estimate active-token demand from the production length distribution, including output growth; reserve an explicit memory and token headroom policy; set admission or maximum-sequence limits from the estimate; run a representative arrival trace; monitor preemption, recomputation, rejection, queueing, and OOM; and revise the limit until the capacity and latency criteria are both satisfied.

\begin{figure}
\centering
\includegraphics[width=\linewidth]{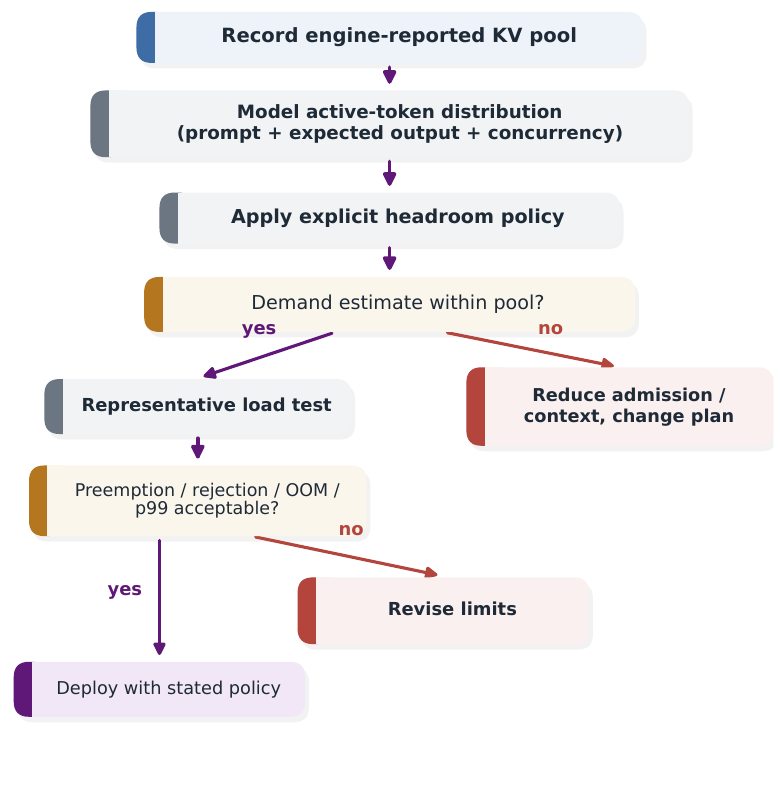}
\caption{\textbf{Capacity validation workflow.} Record the engine-reported KV pool, model the active-token distribution, apply a headroom policy, set admission limits, and confirm with a representative load test. A decision workflow, not a measured result.}
\end{figure}

\begin{findingbox}E15 measured a 245{,}088-token KV pool for Qwen3-14B under one H100/vLLM configuration. Dividing that pool by equal active-token lengths yields illustrative bounds of 119 at 2{,}048 tokens and 7 at 32{,}768. These are derived planning values, not directly observed failure points. Deployment limits should include output growth, length distribution, headroom, and a representative load test.
\end{findingbox}

The final chapter of this part tunes the scheduler token budget and finds that the best setting depends on the model.

\chapter{Chunked-Prefill Tuning Is Model- and Workload-Specific}

\begin{figure}
\centering
\includegraphics[width=\linewidth]{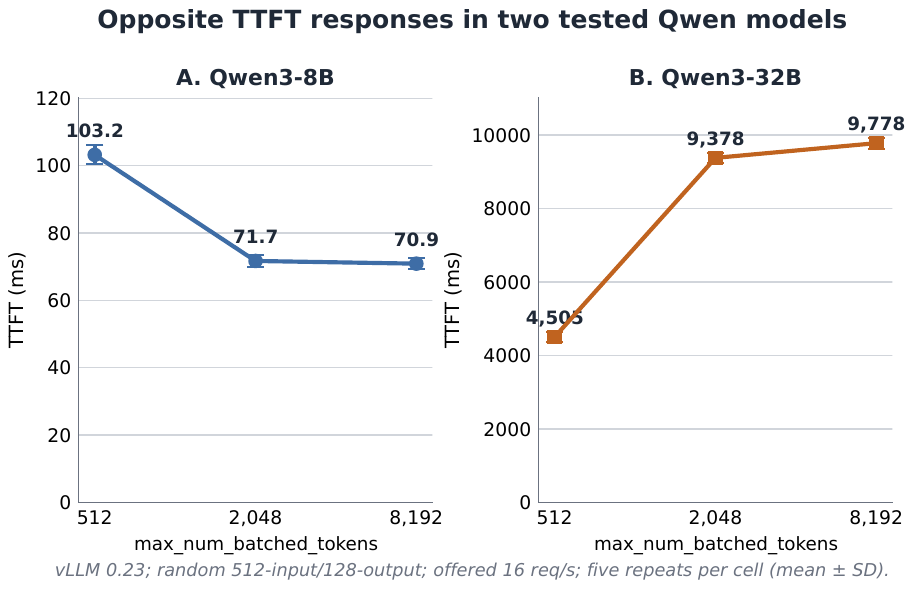}
\caption{\textbf{Opposite TTFT responses in two tested Qwen models.} Raising the scheduler token budget from 512 to 2{,}048 lowered Qwen3-8B mean TTFT (103.2$\pm$2.8 to 71.7$\pm$1.7 ms) but raised Qwen3-32B mean TTFT (4505$\pm$144 to 9378$\pm$141 ms). ITL stayed near 23--24 ms (8B) and 41 ms (32B). Conditions: vLLM 0.23, random 512/128, 16 req/s, five repeats. Source: E36.}
\end{figure}

\medskip{}

A serving engine decides how many tokens to process in one scheduler iteration. That ceiling is the \emph{scheduler token budget}. With chunked prefill enabled, it also governs how much prompt processing is interleaved with ongoing decode work, so it is a deployment parameter rather than a model-independent default. vLLM documents chunked prefill as a mechanism that mixes prefill chunks with decode work and exposes the budget as a latency and throughput control, under the name \code{max\_num\_batched\_tokens} \autocite{vllm2025chunked011}; that documentation describes the control, and E36 supplies the model-specific measurements below.

E36 tested three budgets, 512, 2{,}048, and 8{,}192, on Qwen3-8B and Qwen3-32B under one vLLM 0.23 arrival workload. The two models showed opposite TTFT responses.

\section{15.1 Hardened result}

The hardening run used five benchmark repeats per cell with random 512-input/128-output requests at an offered rate of 16 requests per second.

\begin{table}[htbp]\centering\TableFontEightSix\TableRowAir
\TableTitle{Scheduler token budget by model}
\begin{tabularx}{\linewidth}{@{}>{\RaggedRight\arraybackslash}X C C C@{}}
\arrayrulecolor{accent}\toprule
\thd{Model} & \thd{Token budget} & \thd{TTFT} & \thd{ITL}\\[-4pt]
 & \thu{tokens} & \thu{ms, mean \textpm\ SD} & \thu{ms, mean \textpm\ SD}\\
\arrayrulecolor{rule}\midrule
Qwen3-8B  & 512      & \pmv{103.2}{2.8} & \pmv{24.4}{0.4}\\
\zrow Qwen3-8B  & 2{,}048  & \pmv{71.7}{1.7}  & \pmv{23.5}{0.1}\\
Qwen3-8B  & 8{,}192  & \pmv{70.9}{1.9}  & \pmv{23.3}{0.2}\\
\zrow Qwen3-32B & 512      & \pmv{4505}{144}  & \pmv{41.3}{0.2}\\
Qwen3-32B & 2{,}048  & \pmv{9378}{141}  & \pmv{40.9}{0.2}\\
\zrow Qwen3-32B & 8{,}192  & \pmv{9778}{301}  & \pmv{41.0}{0.1}\\
\arrayrulecolor{accent}\bottomrule
\end{tabularx}
\numcaptab{Scheduler token budget by model. The same setting did not produce the same latency response after the model size changed: the two Qwen models preferred opposite settings under this workload.}
\TableNote{Token budget is vLLM's \code{max\_num\_batched\_tokens}. vLLM 0.23, \code{vllm bench serve}, random 512-input/128-output, offered rate 16 req/s, five repeats per cell. Output throughput is reported in the E36 discussion; the hardened five-seed throughput for the Qwen3-32B/512 setting was not internally consistent, so only the verified TTFT and ITL are tabled here. Evidence: E36.}
\end{table}

For Qwen3-8B, increasing the budget from 512 to 2{,}048 reduced mean TTFT from 103.2 to 71.7 ms, approximately 30.5\%; the 2{,}048 and 8{,}192 cells were close. For Qwen3-32B, the same increase raised mean TTFT from 4{,}505 to 9{,}378 ms, approximately 2.08$\times$, and the 8{,}192 cell was similar to 2{,}048. Across all six cells, ITL changed little relative to the TTFT differences: under this workload, the scheduler token budget affected first-token delay more strongly than steady decode spacing.

\section{15.2 Interpretation}

The result is consistent with a model-dependent balance between prompt-processing service time and decode scheduling. A larger token budget may process more prefill work before the scheduler returns to other requests. That can be beneficial when the prefill work is short enough that batching efficiency dominates, and harmful when a longer prefill interval delays first-token service for other requests. E36 did not record kernel timelines, scheduler queue residence, or per-iteration prefill/decode composition, so the mechanism is presented as an interpretation. A follow-up causal study would collect those traces while varying one scheduling parameter at a time.

\section{15.3 Scope}

The experiment covers two Qwen models, one GPU class, one engine version, one 512-input/128-output distribution, one offered rate, and three scheduler settings. It does not establish a universal rule that large models always require smaller values or that small models always require larger values. The robust conclusion is operational: a scheduler token budget tuned for one deployment should be remeasured after changing the model, engine, hardware, parallel topology, or traffic distribution.

\section{15.4 Tuning workflow}

Select a representative arrival trace and prompt/output distribution; sweep the scheduler token budget across a supported range; measure achieved request rate, TTFT median/p99, ITL or TPOT, end-to-end latency, and output throughput; repeat each cell and report mean/median plus spread; choose the region that satisfies the declared SLO rather than picking from TTFT alone; and repeat the sweep after a model-size, engine-version, GPU, or traffic change.

\begin{findingbox}In E36, Qwen3-8B and Qwen3-32B showed opposite TTFT responses to the tested scheduler token budgets: the 8B TTFT fell from 103.2$\pm$2.8 to 71.7$\pm$1.7 ms at 2{,}048, while the 32B TTFT rose from 4505$\pm$144 to 9378$\pm$141 ms. ITL remained nearly flat. This is a deployment-specific tuning result, not a universal model-size law.
\end{findingbox}

\chapter{Boundary Conditions for Three Common Serving Interventions}

\begin{figure}
\centering
\includegraphics{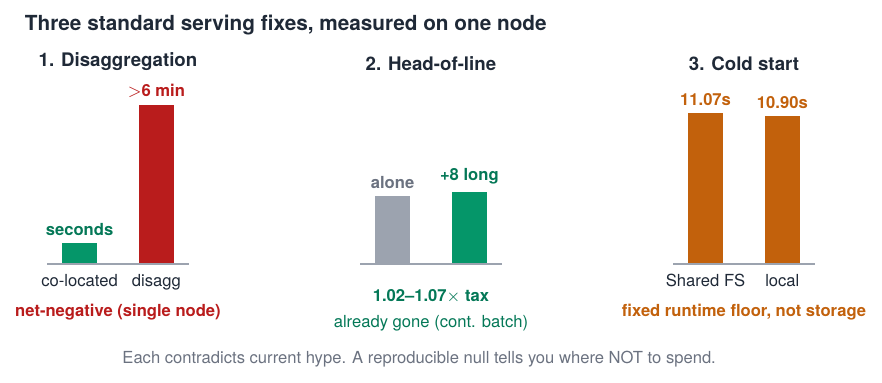}
\caption{\textbf{Three serving interventions, measured under one node.} Single-node prefill/decode disaggregation was net-negative because key-value transfer dominated without a fast fabric; under continuous batching, adding long requests inflated short-request latency by only 1.02--1.07$\times$; and warm-restart cold start was a runtime-bound floor near 11 seconds that storage tier did not change, while a later cache-drop study measured a separate storage-bound first-touch regime. Each result defines a boundary rather than a universal rejection.}
\end{figure}

\medskip{}

Serving interventions are useful only when they address the active bottleneck under the deployed topology and workload. This chapter examines three commonly proposed interventions: prefill/decode disaggregation, mitigation of head-of-line blocking, and faster model storage for startup.

The three studies do not test one shared mechanism. They share a decision rule: before adopting an intervention, identify the resource or scheduling condition it is intended to improve, then measure whether that condition exists in the target deployment.

Under the tested single-node connector, prefill/decode disaggregation was slower than the co-located baseline because key-value transfer and host-memory spill dominated. Under the tested vLLM 0.23 continuous-batching setup, adding long requests produced little short-request latency inflation. Cold-start measurements separated a storage-bound first-touch regime from a runtime-bound warm-restart floor. Each result therefore defines a boundary, not a universal rejection of the underlying technique.

\section{16.1 Single-node disaggregation was net-negative}

Prefill/decode disaggregation has become a common serving design. It runs the prompt-processing phase on one set of accelerators and the token-generation phase on another, so that a long prefill cannot stall ongoing decodes. LMCache provides a KV-cache layer for reuse and offload across inference components \autocite{cheng2025lmcache}, and NIXL provides a versioned transfer library used by the tested connector path \autocite{nvidia2025nixl}. The registered prediction was that it would cut the decode tail on long-prompt traffic and be net-negative on short-bursty traffic, with the key-value-transfer overhead dominating in the short case.

Under the tested connector, topology, and traffic shapes, the disaggregated configuration was slower than the co-located baseline on both shapes. The same load that a co-located server completed in seconds did not complete under disaggregation on this node: even 30 short requests ran for more than six minutes without finishing, accompanied by thousands of host-memory-spill events. On a single node there is no fast inter-accelerator path for the key-value cache, so every prefilled cache round-trips through host pinned memory and the system becomes transfer-bound. Disaggregation did not reduce the decode tail; it increased end-to-end latency.

The result is scoped precisely. The co-located tail numbers stand on their own: a co-located server does have a real decode-tail effect on long prompts, roughly a 4.9$\times$ p99 inflation, that short traffic barely triggers. The single-node disaggregation intervention paid a transfer cost that erased that benefit. A win from disaggregation requires multiple nodes with a dedicated fast key-value-transfer fabric. The tested single-node setup does not support a recommendation to disaggregate prefill and decode.

\section{16.2 Head-of-line blocking under continuous batching}

A first-come-first-served queue can allow a few long requests to block the short ones behind them and inflate their tail latency. The standard mental model estimates that this inflates short-request p99 by two to ten times.

The measured effect was much smaller on the tested engine, and it reproduced across model families. E39 fired 40 short requests alone, then the same 40 interleaved with 8 concurrent long (800-token) requests, and measured the short-request slowdown. The inflation was 2 to 7\% in the median and 0 to 4\% at p99, even with the long requests sharing the batch. Continuous batching interleaves at the token level: a long request advances one token at a time alongside other active requests rather than holding the engine until completion. The pattern held on both Llama-3.1-8B (1.07$\times$) and Mistral-7B (1.02$\times$), which points to the scheduler rather than a checkpoint-specific effect. In this setup, continuous batching already limited the familiar head-of-line penalty, so an additional mitigation addressed little remaining delay. This is a measured boundary for the tested scheduler and traffic shape, not a universal claim about all continuous-batching engines.

\section{16.3 Cold-start regimes: storage-bound first touch and runtime-bound warm restart}

Cold-start behavior has storage-bound and runtime-bound regimes. E40 decomposed a Qwen3-8B (16.38 GB) cold start, comparing a shared network file system against local block-backed storage, five fresh-process repeats per tier.

The first study measured only the warm-cache regime and could not isolate first touch. In that regime, storage tier was statistically indistinguishable. Engine initialization took 11.07 seconds from the shared network file system and 10.90 seconds from local block-backed storage, within noise, and raw reads reached the node's memory bandwidth (roughly 10 gigabytes per second) from both, because the node's page cache in host memory retains the weights after first touch. First-token latency after initialization was 0.137 seconds. So the roughly 11-second warm restart is not weight-fetch; it is a runtime-initialization and warmup floor that storage choice does not reduce.

A later host-level cache-drop experiment measured the storage-sensitive first-touch regime directly. With the host page cache flushed before each read and reads strictly serialized, cold raw reads ran at 34.7 MB/s from the shared network file system versus 130 MB/s from local block-backed storage, a 3.75$\times$ raw-read gap. Cold engine initialization was 201.2 seconds versus 138.5 seconds, a 1.45$\times$ gap. The engine-level gap is smaller than the raw-read gap because cold initialization is weight-read plus the fixed 11-second runtime floor plus deserialization and accelerator transfer, and only the weight-read portion is storage-bound. Cold initialization is therefore roughly 18$\times$ the warm floor: weight-load dominates cold start, but the fixed floor dilutes the tier-to-tier speedup.

The scope is specific. Here, ``local'' was block-backed overlay storage on this node, not local instance-attached NVMe, so the 3.75$\times$ and 1.45$\times$ tier ratios are lower bounds on what dedicated NVMe would give. The measurement used one model size; the storage-bound share of cold start grows with model size, so the tier ratio would widen toward the raw-read gap on a larger checkpoint.

\section{16.4 A common operational procedure}

The three studies support a common operational procedure. First, state the mechanism an intervention is expected to improve. Second, measure the relevant resource or latency component under the target topology. Third, retain the intervention only if the measured bottleneck is present and the end-to-end result improves. A null result can be operationally valuable because it removes an intervention from the critical path, but the conclusion must remain tied to the tested connector, scheduler, storage state, and node topology.

\begin{table}[htbp]\centering\TableFontEightSix\TableRowAir
\TableTitle{Three interventions, three boundaries}
\begin{tabularx}{\linewidth}{@{}Q{1.12in} >{\RaggedRight\arraybackslash}X >{\RaggedRight\arraybackslash}X@{}}
\arrayrulecolor{accent}\toprule
\thd{Intervention} & \thd{Measured under the tested node} & \thd{Condition where it may help}\\
\arrayrulecolor{rule}\midrule
Disaggregation & Net-negative; transfer-bound & Multiple nodes, fast KV fabric\\
\zrow Head-of-line fix & Only 1.02--1.07$\times$ inflation & Schedulers without batching\\
Faster storage & Warm 11.07 vs 10.90 s; cold 34.7 vs 130 MB/s & Cold, storage-bound first touch\\
\arrayrulecolor{accent}\bottomrule
\end{tabularx}
\numcaptab{Each intervention defines a boundary. Under the tested single node, two addressed conditions that were not active, and cold start separated into a storage-bound and a runtime-bound regime.}
\TableNote{Single node, vLLM 0.23. Evidence: E37, E39, E40.}
\end{table}

\begin{table}[htbp]\centering\TableFontNine\TableRowAir
\TableTitle{Cold start has two regimes}
\begin{tabularx}{\linewidth}{@{}Q{0.98in} >{\hsize=0.75\hsize\centering\arraybackslash}X >{\hsize=0.75\hsize\centering\arraybackslash}X >{\hsize=0.75\hsize\centering\arraybackslash}X >{\hsize=1.75\hsize\RaggedRight\arraybackslash}X@{}}
\arrayrulecolor{accent}\toprule
\thd{Regime} & \thd{Shared FS} & \thd{Local} & \thd{Ratio} & \thd{What dominates}\\
\arrayrulecolor{rule}\midrule
cold raw read    & 34.7 MB/s & 130 MB/s & 3.75$\times$ & storage throughput\\
\zrow cold engine init & 201.2 s   & 138.5 s  & 1.45$\times$ & weight fetch\\
warm engine init & 11.07 s   & 10.90 s  & 1.02$\times$ & runtime warmup\\
\arrayrulecolor{accent}\bottomrule
\end{tabularx}
\numcaptab{Cold start has two regimes: first touch is storage-bound, warm restart is runtime-bound. Cold init is roughly 18 times the warm floor, and the engine-level ratio is smaller than the raw-read ratio because only the weight-read portion is storage-bound.}
\TableNote{Shared FS is the shared network file system. Qwen3-8B (16.38 GB), host cache drop before each cold read, serialized reads. ``Local'' means local block-backed overlay storage, not instance-attached NVMe. Evidence: E40.}
\end{table}

\begin{findingbox}Three commonly proposed serving interventions were measured under one node. Single-node prefill/decode disaggregation was net-negative on both traffic shapes because key-value transfer round-tripped through host memory; the win requires multiple nodes with a fast fabric. Under continuous batching, adding long requests inflated short-request latency by 1.02--1.07$\times$ across Llama and Mistral. Cold start split into a storage-bound first touch (34.7 vs 130 MB/s) and a runtime-bound warm floor near 11 seconds. State the mechanism, measure the active bottleneck, and retain the intervention only if that bottleneck is present.
\end{findingbox}

\begin{table}[htbp]\centering\TableFontEightSix\TableRowAir
\TableTitle{Serving cost depends on the control plane}
\begin{tabularx}{\linewidth}{@{}Q{1.16in} C >{\RaggedRight\arraybackslash}X >{\RaggedRight\arraybackslash}X@{}}
\arrayrulecolor{accent}\toprule
\thd{Lever} & \thd{Best gain} & \thd{Bounded cost} & \thd{First production check}\\
\arrayrulecolor{rule}\midrule
Continuous batching & 126.7$\times$ & p99 tail & offered vs achieved\\
\zrow Engine vs plain HF & 20.5$\times$ & not comparable & use batched HF\\
KV-budget planning & prevents OOM & concurrency loss & capacity / context\\
\zrow Chunked prefill & $-27$\% TTFT & 2.08$\times$ on 32B & tune per model\\
P/D disaggregation & none here & KV transfer cost & needs a fast fabric\\
\arrayrulecolor{accent}\bottomrule
\end{tabularx}
\TableCaption{\textbf{Table III.1.} Serving cost depends strongly on scheduling, occupancy, and workload. The largest measured wins and losses came from batching, KV budgeting, and scheduler settings rather than the model weights.}
\TableNote{Best gain is the largest improvement measured in this part; the chunked-prefill figures are for Qwen3-8B and Qwen3-32B respectively. Synthesis of Chapters 12--16. Evidence: E13, E14, E15, E36, E37.}
\end{table}

Supplementary operating measurements continue in Appendix~\ref{app:energy} for energy, Appendix~\ref{app:speculative} for speculative decoding, Appendix~\ref{app:modalities} for multimodal and embedding workload extensions, and Appendix~\ref{app:operations} for runtime failure diagnosis.
\clearpage

%


\addtocontents{toc}{\protect\newpage}
\part{Interface Controls and Their Limits}

\begin{partopener}
An interface control changes one property; it does not change the others by implication.
\end{partopener}

The preceding parts examined evaluation validity, numerical execution, and serving behavior. This part turns to interface-level controls: requesting a structured format, constraining decoding to a grammar, validating tool calls, and prompting a target reasoning length.

These controls act on different objects. A format request changes the prompt. A grammar constraint changes the decoder's admissible token set. A schema can guarantee parseability without establishing semantic correctness. A prompted thinking target can influence wording without imposing a hard token limit. Treating these controls as interchangeable produces misleading expectations.

Chapters 17--20 measure each control separately and state the contract it can support. The purpose is not to argue that the controls are ineffective. It is to identify which property each control changes, which property remains unmeasured, and what must be validated before deployment.

\chapter{The Format Tax Is the Decoder, Not the Request}

\begin{figure}
\centering
\includegraphics{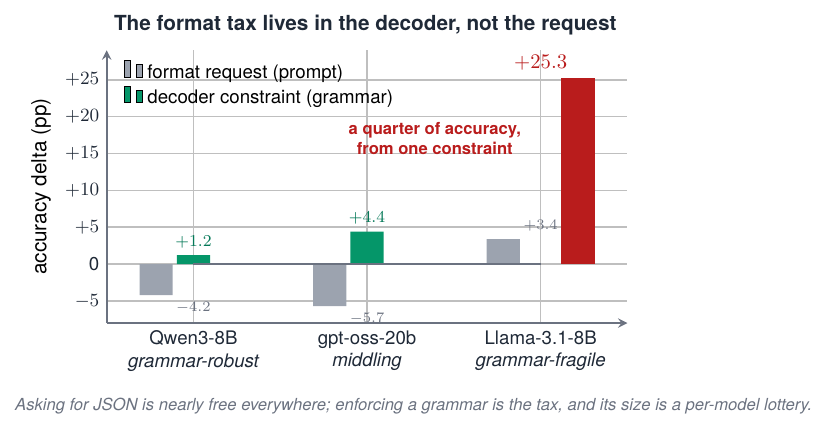}
\caption{\textbf{The format tax decomposed.} Across the three tested checkpoints, requesting JSON in the prompt had a small or beneficial measured effect. The larger cost was concentrated in the grammar constraint, and it varied by checkpoint: 1.2 points on Qwen3-8B, 4.4 points on gpt-oss-20b (noisier cell), and 25.3 points on Llama-3.1-8B. Five seeds, GSM8K-200, thinking disabled, vLLM 0.23. Evidence: E33.}
\end{figure}

\medskip{}

Requesting JSON from a model is widely believed to cost accuracy. The measured effect was real on one checkpoint and small on the others, but the cost came from constraining the decoder to a grammar rather than from asking for the format, and the magnitude was checkpoint-specific. Establishing this required catching a confound in the first measurement.

The tax decomposes into two parts, measured across three checkpoint families. The format request asks for JSON in the prompt with no enforcement; the decoder constraint forces the model's tokens to match a JSON grammar during generation, using a structured-output backend such as XGrammar, a context-free-grammar-based structured-generation engine \autocite{dong2025xgrammar}. E33 measured freeform accuracy, JSON-requested accuracy, and JSON-constrained accuracy on GSM8K, five seeds each, and subtracted to isolate each component. The measured +1.2 and +25.3 point effects are E33 results, not properties reported by that engine's paper.

\section{17.1 The confound that flagged itself}

The first run produced an impossible number. Qwen3-8B's freeform accuracy returned 0.272, although the model scores about 0.92 on GSM8K. That gap indicated a broken measurement rather than a format effect. The freeform condition had left thinking mode enabled, so the model emitted long reasoning chains that were truncated at the token limit before reaching the answer, the truncation floor from \hyperlink{chapter.1}{Chapter 1}. The apparent 27-point format effect was a thinking-mode truncation artifact, not a format effect.

This is the discipline the book applies throughout, seen in miniature. An impossible number is a signal to investigate rather than a data point to report. Re-running with thinking mode disabled on every family removed the confound, and only then did the reconciled result appear. When a measurement is impossible, resolve the confound before reporting the number, because the impossibility is the measurement indicating that it is broken.

\begin{figure}
\centering
\includegraphics{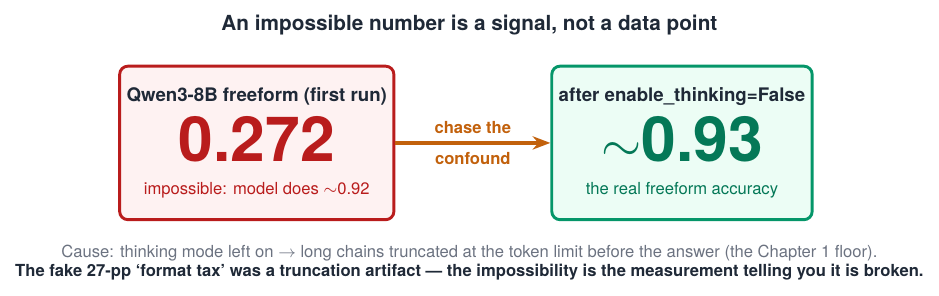}
\caption{\textbf{The confound that flagged itself.} Qwen3-8B's freeform accuracy first returned 0.272, inconsistent with the model's roughly 0.92 GSM8K accuracy. Thinking mode had been left enabled, so long reasoning chains were truncated before the answer, the Chapter 1 floor. The apparent 27-point format effect was a measurement artifact, and the impossibility identified it.}
\end{figure}

\section{17.2 The request is small; the constraint carries the cost}

Define the sign convention directly. Request effect equals freeform accuracy minus JSON-requested accuracy. Decoder effect equals JSON-requested accuracy minus grammar-constrained accuracy. Positive values indicate a decrease under the changed condition.

The clean result inverts the common expectation that requesting JSON in the prompt is the expensive step. Across the three tested checkpoints, the request effect was small, from $-5.7$ to $+3.4$ percentage points, and on two checkpoints requesting JSON slightly improved accuracy. The prompt request itself was not the larger cost in this study.

The measured cost was concentrated in the decoder constraint, and there it varied substantially across the tested checkpoints. Constraining tokens to a JSON grammar cost Qwen3-8B 1.2 points, cost gpt-oss-20b 4.4 points on a noisier cell, and cost Llama-3.1-8B 25.3 points, a 25.3-percentage-point decrease from a single change. The task, seeds, and schema were held fixed; only the constrained checkpoint differed, and the enforcement cost ranged from near zero to roughly a quarter of the checkpoint's accuracy.

Structured output can therefore reduce accuracy, but the mechanism is grammar enforcement rather than the prompt, and the magnitude is a checkpoint-specific response that requires validation. Some checkpoints are grammar-robust and pay almost nothing; some are grammar-fragile and pay substantially. Determining which requires measuring the specific checkpoint.

\begin{table}[htbp]\centering\TableFontEightSix\TableRowAir
\TableTitle{The request is small; the decoder constraint carries the cost}
\begin{tabularx}{\linewidth}{@{}Q{0.92in} C C C C C@{}}
\arrayrulecolor{accent}\toprule
\thd{Model} & \thd{Freeform} & \thd{Asked} & \thd{Forced} & \thd{Request $\Delta$} & \thd{Decoder $\Delta$}\\[-4pt]
 & \thu{percent} & \thu{percent} & \thu{percent} & \thu{pp} & \thu{pp}\\
\arrayrulecolor{rule}\midrule
Qwen3-8B     & 90.9 & 95.1 & 93.9 & −4.2 & +1.2\\
\zrow gpt-oss-20b  & 73.8 & 79.5 & 75.1 & −5.7 & +4.4\\
Llama-3.1-8B & 83.9 & 80.5 & 55.2 & +3.4 & +25.3\\
\arrayrulecolor{accent}\bottomrule
\end{tabularx}
\numcaptab{The request effect is small or beneficial; the decoder-constraint effect ranged from $+1.2$ pp on Qwen3-8B to $+25.3$ pp on Llama-3.1-8B. Asked means JSON requested in the prompt; forced means the decoder constrained to a JSON grammar.}
\TableNote{Request $\Delta$ = freeform $-$ asked; decoder $\Delta$ = asked $-$ forced; a positive value is a decrease. GSM8K-200, five seeds, thinking disabled, vLLM 0.23. Cells reconcile under that convention: the Qwen3-8B freeform value (90.9) follows from the asked cell (95.1) and the measured request effect ($-4.2$ pp), and the decoder effect ($+1.2$ pp) follows from the asked and forced cells. The gpt-oss forced cell is noisier and secondary. Evidence: E33.}
\end{table}

\section{17.3 What to validate before enforcing a schema}

Measure constrained and unconstrained accuracy on the intended checkpoint under the intended grammar before deploying anything that enforces a schema, because the cost is checkpoint-specific and the deployed checkpoint may be the one that pays 25 points. Where a checkpoint is grammar-fragile, consider a prompt-requested format instead of hard enforcement. If the model already produces valid JSON reliably, the accuracy cost of the constraint may be unnecessary. Establish that price by running both conditions on the checkpoint to be deployed.

The clean comparison attributes the larger effect to grammar-constrained decoding rather than to the request for JSON. The measured decoder-constraint cost was 1.2 percentage points for Qwen3-8B and 25.3 points for Llama-3.1-8B; the gpt-oss constrained cell was noisier and should remain secondary. The result is therefore checkpoint-specific. A structured-output deployment should compare unconstrained and constrained task performance on the intended checkpoint, schema, engine version, and prompt format.

\begin{findingbox}The structured-output effect measured here came mainly from grammar enforcement, not the format request. Across the three tested checkpoints, requesting JSON changed accuracy by $-5.7$ to $+3.4$ pp; constraining the decoder cost $+1.2$ pp on Qwen3-8B, $+4.4$ pp on gpt-oss-20b (noisy), and $+25.3$ pp on Llama-3.1-8B. An impossible early number (Qwen at 0.272) was a thinking-truncation confound. Compare constrained and unconstrained accuracy on the intended checkpoint before enforcing a schema, and consider a prompt-requested format on grammar-fragile checkpoints.
\end{findingbox}

\begin{figure}[htbp]
\centering
\includegraphics[width=\linewidth]{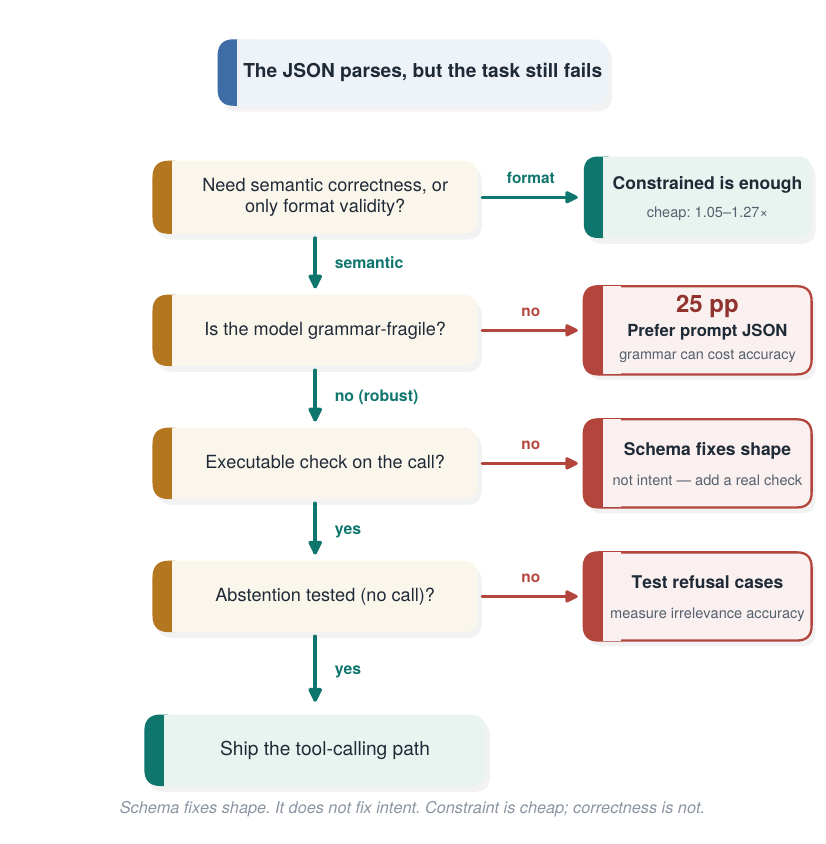}
\caption{\textbf{What to check first: structured output.} Decide whether the requirement is format or intent: a grammar constraint can reduce accuracy and does not change which function the model selected. A decision map for this chapter; consult it by symptom.}
\end{figure}
\nextlesson{\hyperlink{chapter.18}{Chapter 18}, the throughput component. If enforcement can cost accuracy, the next question is whether it also costs throughput on a current engine.}

\chapter{Guided-Decoding Overhead on a Modern Engine}

\begin{figure}
\centering
\includegraphics{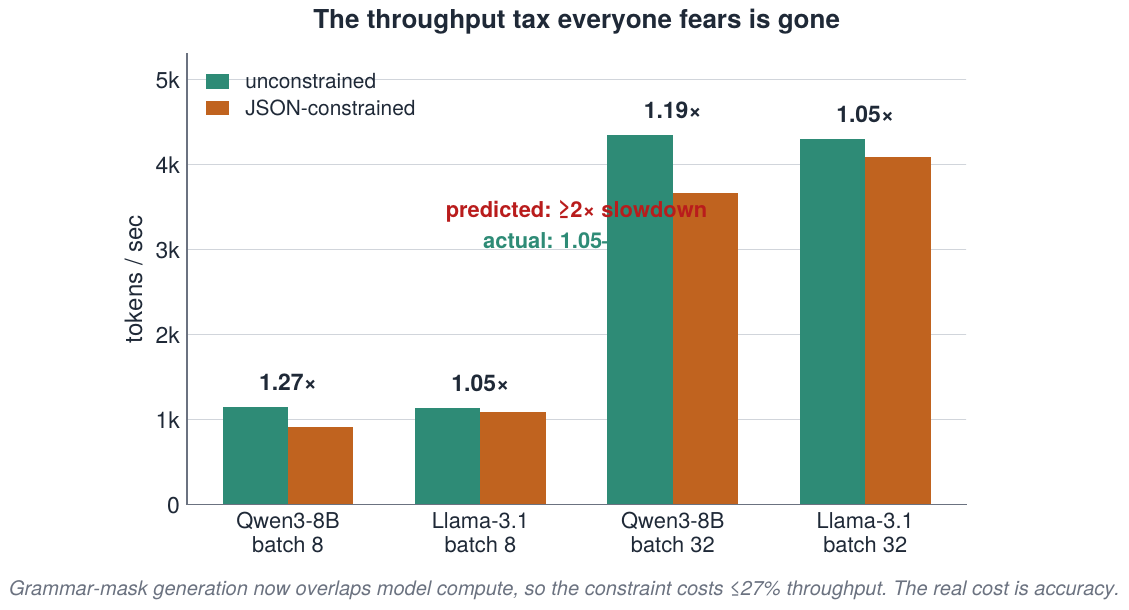}
\caption{\textbf{The measured guided-decoding overhead.} Constrained JSON decoding was predicted to at least halve throughput; on vLLM 0.23 it slowed throughput by a factor of 1.05 to 1.27 across both checkpoints and the reported batch sizes, corresponding to retaining roughly 79--95\% of unconstrained throughput. The overhead is smaller than the preregistered 2$\times$ threshold but nonzero. In the panel, ``gone'' refers to the predicted twofold slowdown, not to zero overhead. Five repeats. Evidence: E34.}
\end{figure}

\medskip{}

A common warning about structured output is that it reduces throughput: constraining the decoder means generating a grammar mask at every token, which was expected to stall the batch. That overhead was larger on older implementations. On the tested current engine the measured overhead was much smaller, and reporting it is useful because it moves attention from a reduced cost to the one that remains.

E34 measured throughput with and without a JSON constraint on Qwen3-8B and Llama-3.1-8B across batch sizes, with five repeats per cell. The prediction was that constrained throughput would degrade at least twofold at batch 8 and above. The measured slowdown was a factor of 1.05 to 1.27 across both checkpoints and the reported batches. At batch 8, Qwen3-8B went from 1155 to 908 tokens per second (1.27$\times$) and Llama-3.1-8B from 1121 to 1063 (1.05$\times$). At batch 32, Qwen3-8B went from 4387 to 3685 (1.19$\times$) and Llama-3.1-8B from 4460 to 4156 (1.07$\times$). The overhead was below the preregistered 2$\times$ threshold on every reported cell.

\section{18.1 The measured overhead and its scope}

A 1.27$\times$ slowdown corresponds to retaining about 78.7\% of unconstrained throughput; a 1.05$\times$ slowdown corresponds to retaining about 95.2\%. The overhead is materially smaller than the 2$\times$ threshold the older estimates predicted, but it is not zero. The reduction of roughly 5 to 21\% may be operationally acceptable, and it should be stated as a bounded cost rather than as no cost.

Older throughput estimates came from implementations that serialized mask generation with model compute: at each token the engine computed the grammar-admissible token set before sampling, and that pause stalled the batch. On the tested engine the observed overhead is consistent with grammar-mask generation being largely hidden behind model compute, but E34 measured throughput only; it did not instrument mask-generation timing, kernel overlap, or scheduler traces. The overlap explanation is therefore a hypothesis consistent with the observed overhead, not a directly measured mechanism.

The scope is specific. The result covers vLLM 0.23, the tested grammar backend, the tested JSON schema, two checkpoints, and the reported batch sizes. Cost estimates from older implementations should be remeasured on the current engine version, because an estimate that was correct on an earlier version can be misleading now. Re-evaluate the overhead after an engine, grammar-backend, schema, model, or workload change.

\begin{table}[htbp]\centering\TableFontNine\TableRowAir
\TableTitle{Guided-decoding throughput overhead on vLLM 0.23}
\begin{tabularx}{\linewidth}{@{}Q{0.36in} Q{0.86in} C C C@{}}
\arrayrulecolor{accent}\toprule
\thd{Batch} & \thd{Model} & \thd{Unconstrained} & \thd{Constrained} & \thd{Factor}\\[-4pt]
 & & \thu{tok/s} & \thu{tok/s} & \thu{ratio}\\
\arrayrulecolor{rule}\midrule
8  & Qwen3-8B     & 1{,}155 & 908     & 1.27\\
\zrow 8  & Llama-3.1-8B & 1{,}121 & 1{,}063 & 1.05\\
32 & Qwen3-8B     & 4{,}387 & 3{,}685 & 1.19\\
\zrow 32 & Llama-3.1-8B & 4{,}460 & 4{,}156 & 1.07\\
\arrayrulecolor{accent}\bottomrule
\end{tabularx}
\numcaptab{On vLLM 0.23, JSON constraints slowed throughput by a factor of 1.05 to 1.27, below the preregistered 2$\times$ threshold, retaining about 79 to 95\% of unconstrained throughput. The throughput cost was bounded; the larger and more variable effect appeared in accuracy.}
\TableNote{Qwen3-8B and Llama-3.1-8B, GSM8K, five repeats, vLLM 0.23, tested grammar backend and schema. Evidence: E34.}
\end{table}

\section{18.2 Where the structured-output cost concentrates}

Reading Chapters 17 and 18 together locates two costs. The throughput slowdown was 1.05 to 1.27$\times$ on the tested engine: nonzero, but below the preregistered 2$\times$ threshold. Requesting JSON in the prompt had a small measured effect across the three tested checkpoints. The larger and more variable effect came from grammar enforcement on accuracy, ranging from 1.2 points on Qwen3-8B to 25.3 points on Llama-3.1-8B. Both effects should be measured; the accuracy effect was the more checkpoint-sensitive one in these studies.

That changes the pre-deployment check. Do not reject guided decoding solely on the older twofold-throughput estimate; measure its current overhead on the intended stack. Also measure constrained accuracy for each checkpoint, because that is where the larger and more variable effect appeared here.

\begin{findingbox}On vLLM 0.23, the tested JSON constraint slowed throughput by a factor of 1.05--1.27 across the reported checkpoint and batch conditions, retaining roughly 79--95\% of unconstrained throughput. The overhead is smaller than the preregistered 2$\times$ threshold but nonzero. The experiment measured throughput only, so the mask-overlap explanation is a hypothesis, not a measured mechanism. Combined with the previous chapter, the larger measured structured-output cost was accuracy on grammar-fragile checkpoints. Re-evaluate the overhead after an engine, grammar-backend, schema, model, or workload change.
\end{findingbox}

\begin{figure}[htbp]
\centering
\includegraphics[width=\linewidth]{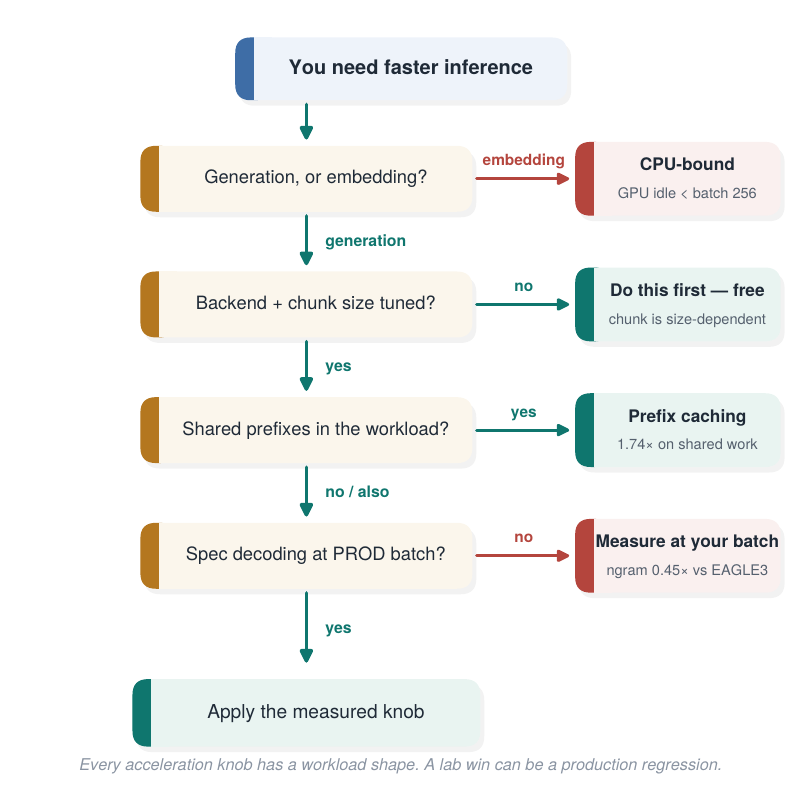}
\caption{\textbf{The acceleration control.} Every acceleration control has a workload shape; a value tuned on one model size or batch regime can degrade another, so measure the control on the intended shape. A decision map for this chapter; consult it by symptom.}
\end{figure}
\nextlesson{\hyperlink{chapter.19}{Chapter 19}, the semantic layer. Where enforcement is bounded in throughput and accuracy, it still guarantees only that the output parses, not that the call is correct.}

\chapter{Valid JSON Is Not a Correct Tool Call}

\begin{figure}
\centering
\includegraphics{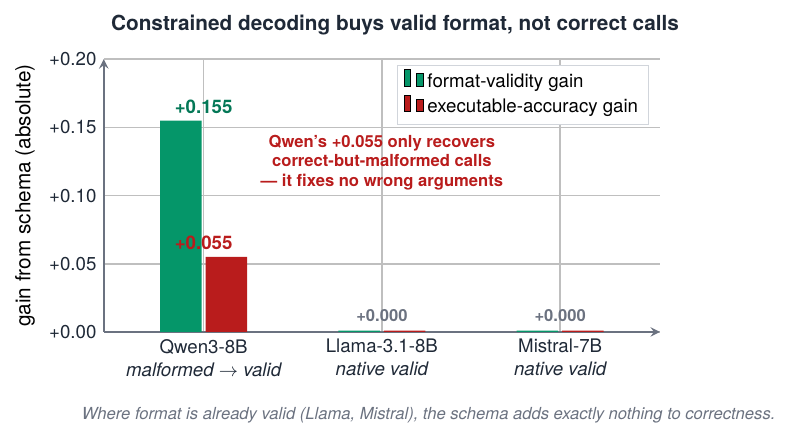}
\caption{\textbf{Constrained decoding raised format validity, not semantic accuracy.} Where a checkpoint already emitted valid tool calls (Llama, Mistral at 1.0 format validity), the schema changed BFCL AST accuracy by 0.0 percentage points. Where it raised format validity (Qwen3-8B, $+0.155$), the AST-accuracy gain was about a third the size ($+0.055$). Abstention accuracy varied widely across checkpoints. BFCL v4, four families, five seeds. Evidence: E35.}
\end{figure}

\medskip{}

This distinction underlies the previous two chapters and matters most because tool-calling is where structured output is most used. Constraining a decoder to a schema guarantees that the output parses. It does not guarantee that the call is correct. A well-formed call to the wrong function, or the right function with the wrong arguments, is still incorrect, and the schema does not prevent it. Format validity and semantic correctness are different quantities, and the constraint changes only the first.

The Berkeley Function Calling Leaderboard provides AST-based evaluation for function-call structure and correctness \autocite{patil2025bfcl}. The chapter uses three quantities consistently. Format validity is whether the output parses under the required schema. BFCL AST accuracy is whether the tool call matches the benchmark's expected function and arguments; because the BFCL procedure was an abstract-syntax-tree and semantic checker rather than execution of a real tool, the metric is named AST accuracy, not executable accuracy. Abstention accuracy is whether the model refrains from calling a tool when no call is appropriate. E35 measured these on BFCL v4 across four checkpoint families, five seeds, under the native output and under schema-constrained decoding. The retained notes identify the bundled \code{bfcl\_eval} package, official \code{ast\_checker}, \code{simple\_python} and \code{irrelevance} categories, and 40 items per category per seed, but they do not preserve the package version or item IDs. The numerical results are therefore scorer- and snapshot-scoped rather than an exact rerun specification.

\section{19.1 The schema changed format validity, not semantic accuracy}

Where a checkpoint already produced valid tool calls, the schema changed AST accuracy by 0.0 percentage points. Llama-3.1-8B and Mistral-7B both emitted valid JSON natively (format validity 1.000), and constraining them to the schema changed AST accuracy by 0.000 under the tested scorer. The constraint had nothing to change on the format side and changed nothing on the semantic side, because these are separate quantities.

Where the schema did change format validity, the semantic change was smaller and did not consist of corrected function choices. On Qwen3-8B the schema raised format validity by 0.155, from 0.845 to 1.000, and raised AST accuracy by 0.055, about a third as much, from 0.520 to 0.575. That small AST-accuracy gain is consistent with recovering calls that were already correct in substance but written in a form the parser rejected, rather than with converting incorrect calls into correct ones. The item-level transitions needed to assert that no wrong call became right were not printed here, so the conclusion is stated as consistent with format recovery rather than as an established never-fixed claim.

Constraining the output is therefore a validity change, not a reliability change. It guarantees that the call parses, which is useful, but it does not change whether the call is the correct one to make, which is the operational requirement. A system that constrains its tool calls to a schema and concludes that the calls are reliable has combined two quantities: a control that improves the measured quantity (format validity) while leaving the quantity of interest (semantic correctness) unchanged.

\begin{table}[htbp]\centering\TableFontEightSix\TableRowAir
\TableTitle{Format validity, AST accuracy, and abstention by checkpoint}
\begin{tabularx}{\linewidth}{@{}>{\RaggedRight\arraybackslash}X C C C C@{}}
\arrayrulecolor{accent}\toprule
\thd{Model} & \thd{Format} & \thd{$\Delta$} & \thd{AST acc.} & \thd{$\Delta$}\\[-4pt]
 & \thu{native} & \thu{with schema} & \thu{native} & \thu{with schema}\\
\arrayrulecolor{rule}\midrule
Qwen3-8B     & 0.845 & +0.155 & 0.520 & +0.055\\
\zrow Llama-3.1-8B & 1.000 & +0.000 & 0.555 & +0.000\\
Mistral-7B   & 1.000 & +0.000 & 0.545 & +0.000\\
\arrayrulecolor{accent}\bottomrule
\end{tabularx}
\numcaptab{Format validity and AST accuracy are separate quantities. Schema enforcement raised format validity where the native output was malformed; it did not change AST accuracy where format validity was already 1.000.}
\TableNote{Values are proportions, not percentages. BFCL v4, five seeds, vLLM 0.23. Abstention accuracy (irrelevance): Qwen3-8B 0.915, Llama-3.1-8B 0.790, Mistral-7B 0.460. A reasoning-format family is excluded as a scorer limitation (a flat-JSON parser cannot read its calls). Evidence: E35.}
\end{table}

\section{19.2 Abstention is the strongest operational axis}

Abstention accuracy separated the checkpoints more than AST accuracy did. On the irrelevance category, where the correct behavior is to refrain from calling a non-applicable function, Qwen3-8B abstained on 0.915 of cases and Llama-3.1-8B on 0.790, but Mistral-7B abstained on only 0.460, issuing a function call on more than half of the requests where no call was appropriate. Two checkpoints with similar AST accuracy (Mistral 0.545, Llama 0.555) differed by roughly a factor of two in abstention accuracy, so single-turn AST accuracy alone does not indicate deployment safety.

\section{19.3 A scorer limitation}

One family is excluded from the cross-checkpoint claim, and the reason is a scorer limitation. A reasoning model that emits its tool calls inside a reasoning-specific format could not be scored by the standard flat-JSON parser, which read its calls as invalid and its non-calls as correct abstentions (format validity 0.205, AST accuracy 0.105, abstention 0.975). Those are parser artifacts under the tested scorer, not model outcomes; scoring the checkpoint fairly requires a parser for its format. The cross-checkpoint claim therefore rests on the three standard-format families, and the fourth is recorded as a scorer limitation rather than folded into the result.

\section{19.4 What to validate for tool-calling}

Measure tool-call semantic correctness separately from format validity, because the constraint changes only the second and deployment requires the first. Use an AST or benchmark semantic check of whether the call matches the expected function and arguments, not only whether it parses, and measure abstention accuracy directly, because it varied more across checkpoints than AST accuracy did. A high format-validity number does not establish reliability. Constrained decoding is useful because valid output is a prerequisite, but it is a floor rather than a ceiling: it produces calls that parse, and producing calls that are correct is a separate requirement that a schema does not satisfy.

\begin{findingbox}Constrained decoding changed format validity, not semantic accuracy. Where a checkpoint already emitted valid tool calls (Llama, Mistral at 1.000 format validity), the schema changed BFCL AST accuracy by 0.0 percentage points; where it raised format validity (Qwen3-8B, $+0.155$), the AST-accuracy gain was about a third the size ($+0.055$) and is consistent with recovering correct-but-malformed calls. Abstention accuracy separated checkpoints more (0.460 to 0.915). Measure tool-call semantic correctness and abstention separately from format validity, and do not treat a high format-validity number as reliability.
\end{findingbox}

\nextlesson{\hyperlink{chapter.20}{Chapter 20}, the last interface control. A prompted thinking budget requests a reasoning length; the measurement checks whether that request controlled the reasoning length actually used.}

\chapter{Prompted Thinking-Budget Adherence}

\begin{figure}
\centering
\includegraphics{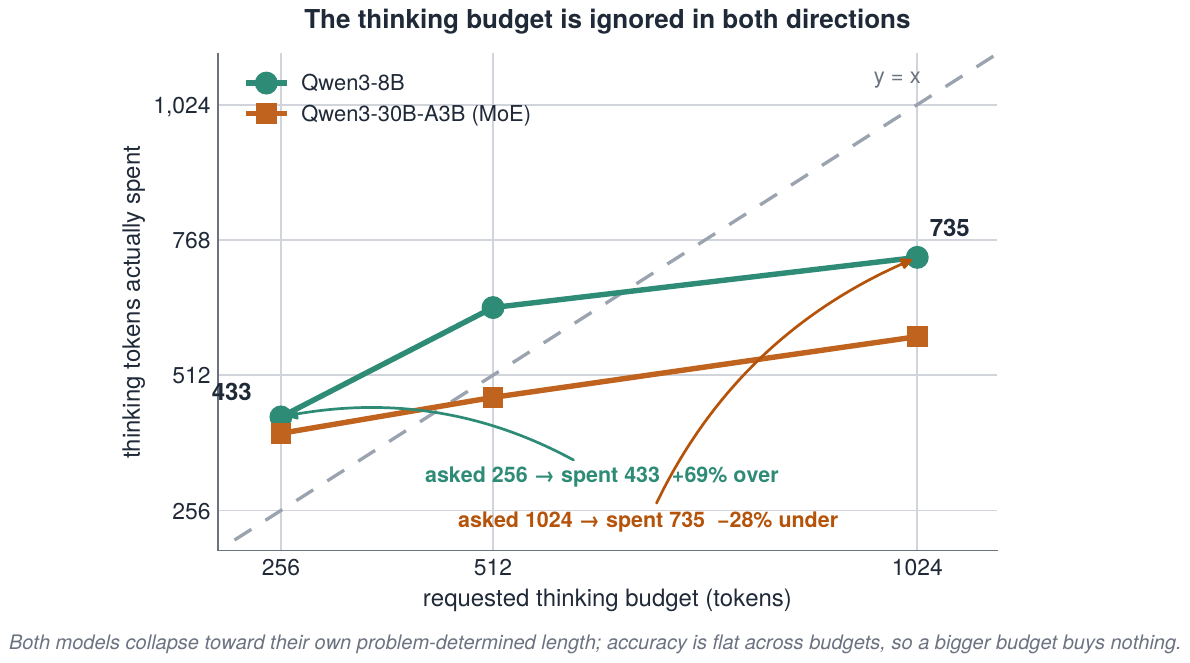}
\caption{\textbf{Requested thinking budget against tokens actually used.} The prompted target did not reliably control the measured reasoning length: asked for 256 tokens, Qwen3-8B used 433 (a 69\% overshoot); asked for 1024, it used 735 (a 28\% undershoot), converging toward a problem-determined length. Accuracy was flat across budgets. Qwen3-8B and Qwen3-30B-A3B, GSM8K, five seeds. Evidence: E46.}
\end{figure}

\medskip{}

The last interface control examined here is the thinking budget: a prompt asks a reasoning model with a thinking/non-thinking interface \autocite{qwen2025qwen3} to keep its reasoning under a target number of tokens. E46 tests whether the request controlled the reasoning length actually used. It did not: both tested checkpoints moved toward a problem-dependent length, overshooting tight targets and undershooting loose ones.

E46 measured adherence directly on two reasoning checkpoints, hardened to five seeds. For each requested budget it counted the tokens the model spent thinking and computed the adherence error as (actual $-$ requested) divided by requested; a positive value indicates overshoot. The requested budgets were 256, 512, and 1024 tokens, on GSM8K.

\section{20.1 Tight requests were overshot; loose requests were undershot}

The pattern was consistent and two-directional. Asked for 256 thinking tokens, Qwen3-8B used a mean of 433, a 69\% overshoot, with 82.5\% of problems over budget. Asked for 512, it used 667, a 30\% overshoot. Asked for 1024, it used 735, a 28\% undershoot, with only 15.7\% of problems over budget: the model did not pad its reasoning to reach a high budget. The adherence error moved from $+0.69$ at the tight budget to $-0.28$ at the loose one, and the reasoning length converged toward a problem-determined value near 430 to 735 tokens.

The same overshoot-at-tight, undershoot-at-loose pattern appeared on Qwen3-30B-A3B, a mixture-of-experts checkpoint of very different size, which converged toward its own reasoning length near 400 to 565 tokens: adherence error $+0.55$ at 256, $-0.06$ at 512, and $-0.45$ at 1024. This extends the observation beyond one checkpoint but does not establish architecture independence or a general law for prompted reasoning budgets. It holds in both tested Qwen checkpoints, one dense and one mixture-of-experts.

\begin{table}[htbp]\centering\TableFontNine\TableRowAir
\TableTitle{Prompted thinking budgets did not control reasoning length}
\begin{tabularx}{\linewidth}{@{}Q{0.92in} C C C C@{}}
\arrayrulecolor{accent}\toprule
\thd{Model} & \thd{Requested} & \thd{Actual} & \thd{Adherence} & \thd{Accuracy}\\[-4pt]
 & \thu{tokens} & \thu{tokens, mean} & \thu{error} & \thu{percent}\\
\arrayrulecolor{rule}\midrule
Qwen3-8B      & 256  & 433 & +0.69 & 92.0\\
\zrow Qwen3-8B      & 512  & 667 & +0.30 & 90.0\\
Qwen3-8B      & 1024 & 735 & −0.28 & 89.3\\
\zrow Qwen3-30B-A3B & 256  & 397 & +0.55 & 97.3\\
Qwen3-30B-A3B & 512  & 479 & −0.06 & 94.0\\
\zrow Qwen3-30B-A3B & 1024 & 565 & −0.45 & 96.7\\
\arrayrulecolor{accent}\bottomrule
\end{tabularx}
\numcaptab{The reasoning length converged toward a problem-determined value. Tight prompt budgets were overshot and loose ones undershot, with accuracy flat across budgets.}
\TableNote{Adherence error = (actual $-$ requested) / requested, mean over seeds. GSM8K, five seeds, vLLM 0.23. A reasoning-effort family (gpt-oss-20b) is not counted here because its reasoning tokens were not directly countable under the harness. Evidence: E46.}
\end{table}

\section{20.2 The budget did not change accuracy either}

Accuracy was flat across every budget. Qwen3-8B moved from 0.920 to 0.893 across the range, and Qwen3-30B-A3B stayed between 0.940 and 0.973, all within seed noise. Raising the requested budget did not change accuracy, consistent with the ceiling from \hyperlink{chapter.5}{Chapter 5}: on these problems the model already spent what it needed, so requesting more reasoning changed neither the reasoning length nor the answer.

Both intended uses of the control therefore did not hold under the tested setup. Setting a larger budget to raise accuracy had no detectable effect, because the model already reasoned to its problem-determined length. Setting a smaller budget to reduce token spend was not reliably honored, because the model overshot it: a request for 256 tokens produced 433. The control did not raise accuracy when increased and did not reduce cost when decreased. It is advisory in both directions, and a system that relies on it for either purpose relies on a control that did not govern the measured reasoning length.

\section{20.3 What to use for a hard bound}

Do not rely on a prompt-level thinking budget as a cost control or an accuracy lever, because neither held here. To bound the reasoning spend, use a hard mechanism, a token cap or a forced end to the reasoning block, rather than a prompt request the model overshot. The shape matches the rest of this part: a control that reads as governing (prompting a reasoning length) was advisory under measurement, and the reliable bound is the hard mechanism, the decoder cap rather than the prompt request. Cap the reasoning at the decoder to bound it, and do not expect the prompt to enforce the bound.

This closes Part IV. The four controls---requesting a format, constraining the decoder, validating a tool call, and prompting a reasoning length---act on different properties. In the tested checkpoints, the prompt request had the smaller accuracy effect; the decoder constraint carried a bounded throughput cost and a larger, checkpoint-specific accuracy cost. The schema made output parseable, not semantically correct. The prompted budget did not control measured reasoning length in either direction. For each control, measure the property it actually changes on the intended checkpoint and workload.

\begin{findingbox}A prompted thinking budget was advisory, not a control. The prompted target did not reliably control the measured reasoning length: asked for 256 tokens Qwen3-8B used 433 ($+69$\%, 82.5\% of problems over), asked for 1024 it used 735 ($-28$\%), converging toward a problem-determined length. The same overshoot-tight, undershoot-loose pattern appeared in both tested Qwen checkpoints, one dense and one mixture-of-experts, which extends the observation but does not establish architecture independence. Accuracy was flat across budgets. To bound reasoning spend, use a hard token cap or forced end, not a prompt request.
\end{findingbox}

\begin{statictable}\TableFontEightSix\TableRowAir
\TableTitle{Each interface control changed one property}
\begin{tabularx}{\linewidth}{@{}Q{1.16in} >{\RaggedRight\arraybackslash}X >{\RaggedRight\arraybackslash}X@{}}
\arrayrulecolor{accent}\toprule
\thd{Control} & \thd{Measured under the tested setup} & \thd{What to validate}\\
\arrayrulecolor{rule}\midrule
Requesting a format & Request effect $-5.7$ to $+3.4$ pp & Request freely; it is not the cost\\
\zrow Constraining the decoder & Throughput 1.05--1.27$\times$; accuracy 1.2--25.3 pp & Constrained accuracy per checkpoint\\
Validating a tool call & Format validity $\neq$ AST accuracy & AST accuracy and abstention\\
\zrow Prompting a reasoning length & Overshoot when tight, undershoot when loose & Bound at the decoder, not the prompt\\
\arrayrulecolor{accent}\bottomrule
\end{tabularx}
\TableCaption{\textbf{Table IV.1.} Each interface control changed one property. Structured output was measured here as an accuracy and semantic-correctness axis that varies by checkpoint, with a bounded throughput overhead.}
\TableNote{Synthesis of Chapters 17--20. Evidence: E33, E34, E35, E46.}
\end{statictable}

\nextlesson{\hyperlink{part.5}{Part V}, long-context capability and architecture. Leaving interface controls for the model's structure: how retrieval, aggregation, reasoning, memory capacity, and throughput behave across context length, and what a model comparison must control before it can support an architecture claim.}

\part{Long-Context Capability and Architecture}

\begin{partopener}
A context length is an input limit; retrieval, reasoning, memory, and throughput are separate capabilities, each of which can fail at a different length.
\end{partopener}

A context-window specification states how many tokens an implementation can accept. It does not state what the model can do across that entire span. Retrieval, aggregation, chained reasoning, memory capacity, and serving throughput are separate properties, and each can fail at a different length.

This part evaluates those properties separately. Chapters 21 and 22 compare retrieval, aggregation, and six-step variable tracking across context length. Chapter 23 reports the operational profile of a hybrid-attention checkpoint: reasoning performance, key-value-cache capacity, aggregate prefill throughput, and prefix-cache behavior. Chapter 24 examines the distinct memory, quantization, and reproducibility considerations of mixture-of-experts models. Chapter 25 closes the part by defining the controls required before a model comparison can support an architecture claim.

The governing rule is simple: every long-context or architecture statement must name the task, metric, threshold, hardware, engine, and comparison design. A context length without those conditions is an input limit, not a deployment guarantee.

\begin{figure}[htbp]
\centering
\includegraphics[width=\linewidth]{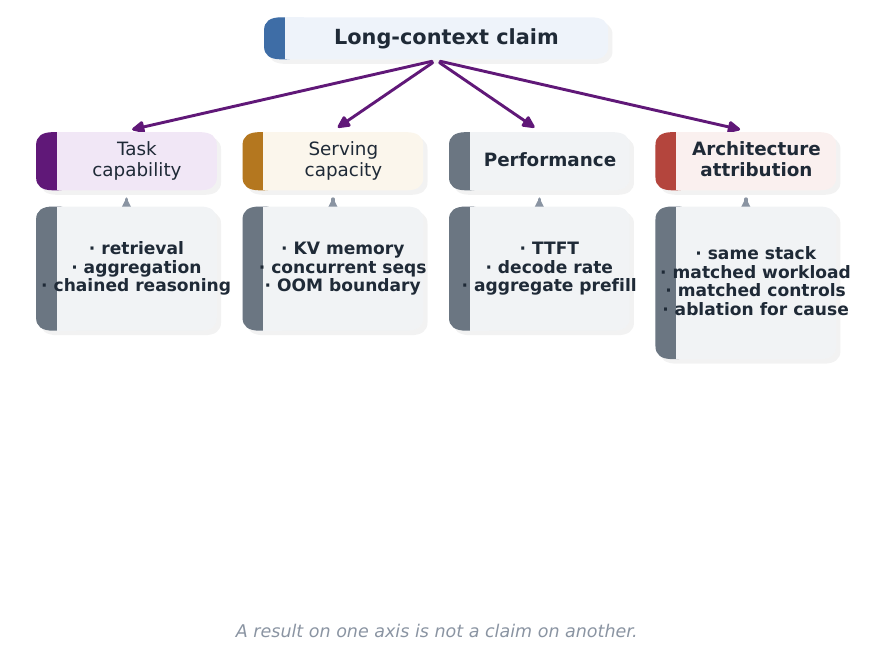}
\caption*{\textbf{Map~V.1.\ The four questions a single context number cannot answer.} Retrieval, integration and reasoning, serving capacity, and comparison design are distinct axes, each measured by a different test. A conceptual map for Part V, not a measured result.}
\end{figure}

\begin{statictable}\TableFontNine\TableRowAir
\TableTitle{Part V reader map}
\begin{tabularx}{\linewidth}{@{}>{\RaggedRight\arraybackslash}X >{\RaggedRight\arraybackslash}X Q{0.42in}@{}}
\arrayrulecolor{accent}\toprule
\thd{Deployment question} & \thd{Measurement required} & \thd{Ch.}\\
\arrayrulecolor{rule}\midrule
Can the checkpoint retrieve at the target length? & Fixed-position retrieval with exact scoring & 21\\
\zrow Can it reason across that length? & Aggregation and workload-matched multi-step tasks with a stated SLO & 21--22\\
Can the serving system hold the contexts and concurrency? & KV capacity, context distribution, memory margin, OOM boundary & 23, 14\\
\zrow Does a comparison isolate the claimed cause? & Same software path plus matched checkpoints, or an explicit confound statement & 23--25\\
\arrayrulecolor{accent}\bottomrule
\end{tabularx}
\TableCaption{\textbf{Table V.0.} Part V reader map: four deployment questions that one advertised context length cannot answer.}
\TableNote{Read with Map V.1, which shows the same four axes as a diagram.}
\end{statictable}

\chapter{Recall Does Not Establish Long-Context Reasoning}

\begin{figure}
\centering
\includegraphics[width=\linewidth]{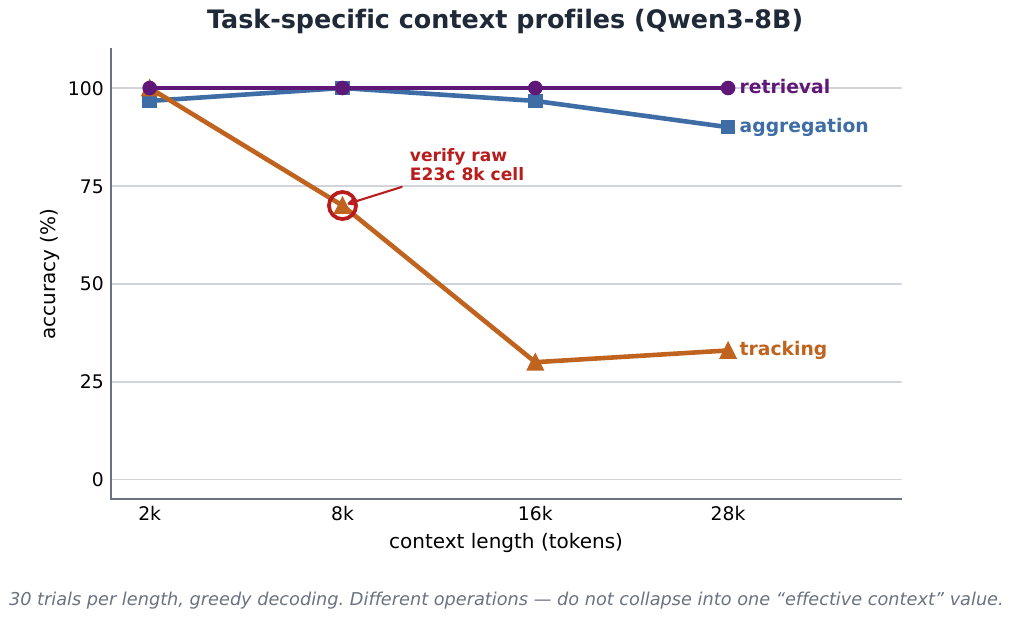}
\caption{\textbf{Task-specific context profiles for Qwen3-8B.} Fixed-position retrieval remained perfect through 28k, common-words aggregation declined to 90\% at 28k, and six-step variable tracking fell to approximately 30\% by 16k. Each task used 30 trials per length and greedy decoding. The curves describe different operations and should not be collapsed into one ``effective context'' value. The 8k tracking point is omitted. Source: E23c and H5.}
\end{figure}

\medskip{}

A context window can support one operation and fail another. To separate those cases, Qwen3-8B was evaluated at approximately 2k, 8k, 16k, and 28k tokens on three tasks: fixed-position retrieval, common-words aggregation, and six-step variable tracking. The context lengths and decoding mode were held fixed; the operation required by the task changed.

The results did not form one context curve. Retrieval remained correct across the tested range. Aggregation declined only modestly at 28k. Six-step variable tracking degraded sharply by 16k. The result is therefore not that long context ``works'' or ``does not work.'' It is that the usable length depends on what the model must do with the context.

\section{21.1 Retrieval, aggregation, and chained reasoning are different operations}

The fixed-position retrieval task asks the model to locate and reproduce a small number of high-salience strings. In the corrected four-needle protocol, all 30 trials were correct at 2k, 8k, 16k, and 28k. This establishes that the checkpoint could retrieve the planted strings under that protocol through the tested range.

The common-words task requires a broader operation: counts must be combined across the context to identify the most frequent terms. Performance remained high (29/30 at 2k, 30/30 at 8k, 29/30 at 16k, and 27/30 at 28k), but the final point shows a modest decline that the retrieval task did not reveal.

Six-step variable tracking requires a sequence of dependent updates distributed through the context. Under the strict-matcher E23c protocol, Qwen3-8B scored 30/30 at 2k, 9/30 at 16k, and 10/30 at 28k (with the 8k point omitted). This operation degraded much earlier than retrieval or aggregation.

The contrast is operational. Retrieval depends heavily on one or a few identifiable spans. Aggregation combines information from many spans but does not require a long dependency chain. Variable tracking requires the output of one step to remain available for the next. These differences provide a plausible explanation for the observed ordering, but the experiment measures task performance rather than internal causal mechanisms.

\begin{table}[htbp]\centering\TableFontNine\TableRowAir
\TableTitle{Three operations over the same context}
\begin{tabularx}{\linewidth}{@{}C C C C@{}}
\arrayrulecolor{accent}\toprule
\thd{Context} & \thd{Four-needle retrieval} & \thd{Common-words aggregation} & \thd{Six-step tracking}\\[-4pt]
\thu{tokens} & \thu{trials correct} & \thu{trials correct} & \thu{trials correct}\\
\arrayrulecolor{rule}\midrule
2k  & 30/30 & 29/30 & 30/30\\
\zrow 16k & 30/30 & 29/30 & 9/30\\
28k & 30/30 & 27/30 & 10/30\\
\arrayrulecolor{accent}\bottomrule
\end{tabularx}
\numcaptab{Three operations over the same context, on Qwen3-8B. Retrieval held through 28k while six-step tracking fell to roughly 30\% by 16k; the three tasks used different generators and scoring rules.}
\TableNote{Qwen3-8B; greedy decoding; 30 trials per length. The 8k six-step-tracking point is omitted because the raw cell was inconsistent between source summaries; the decline is bracketed between the 2k pass and the 16k result. Evidence: E23c, H5.}
\end{table}

\section{21.2 A saturated retrieval test can miss other degradation}

The same sensitivity issue appeared in the KV-cache study. In the hardened short-context serving experiment, the FP8-KV condition was approximately 1.2 percentage points below the automatic-precision condition (\pmv{0.9195}{0.0069} vs \pmv{0.9078}{0.0081}), with overlapping seed ranges. The single-needle 8k retrieval probe nevertheless remained perfect under both conditions. That result does not show that FP8 KV is harmless in every setting; it shows that the retrieval probe was not sensitive to the small difference measured on the short-context task.

A separate RL-rollout experiment applied a stricter distribution-preservation criterion and found a much larger discrepancy. The two studies answer different questions and should remain separate: one concerns task-level serving accuracy, while the other concerns whether an accelerated rollout matches a BF16 policy distribution closely enough for training.

\section{21.3 The first multi-needle probe was rejected}

The initial custom multi-needle experiment produced 25\% all-correct recall with four needles and 85\% with eight. Because the intended task should not become easier when more targets are added, the result indicated a defect in the test design.

Inspection showed that the placement algorithm changed the recency distribution as the needle count increased. The eight-needle condition placed more targets near the end of the context, while the four-needle condition left more targets in less favorable positions. Needle count and placement difficulty were therefore confounded. The result was excluded. The corrected protocol used fixed positions and exact scoring, and the clean retrieval result remained 30/30 through 28k.

\begin{codebox}
\noindent{\sffamily\bfseries\footnotesize\color{accent}Rejected probe.}\enspace{\sffamily\footnotesize The initial four- versus eight-needle comparison changed target placement with needle count and produced a non-monotonic difficulty pattern. The result was excluded. Fixed positions and exact scoring were used in the replacement retrieval test.}
\end{codebox}

\section{21.4 How to use recall in an effective-context evaluation}

Long-context evaluation should include the hardest operation required by the deployment workload. Retrieval is a necessary floor when exact lookup matters. Aggregation should be tested when the task combines evidence across a document. Multi-step tracking or workload-specific reasoning should be included when later decisions depend on earlier facts. Each task requires its own metric and acceptance threshold.

\begin{figure}
\centering
\includegraphics[width=\linewidth]{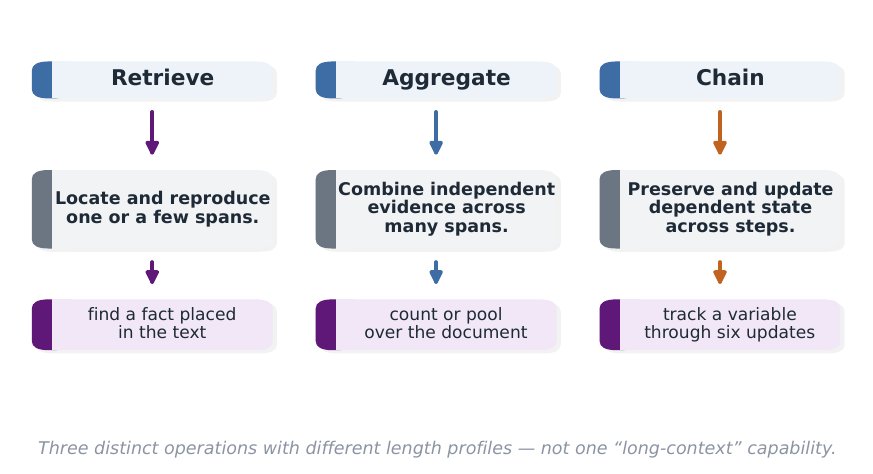}
\caption{\textbf{Three long-context operations.} Retrieve locates and reproduces one or a few spans; aggregate combines independent evidence across many spans; chain preserves and updates dependent state across steps. A conceptual taxonomy, not a claim about internal mechanism.}
\end{figure}

\begin{findingbox}Retrieval remained perfect through 28k in the corrected four-needle test, aggregation remained at 90\% or above, and six-step tracking fell to approximately 30\% by 16k. The checkpoint therefore had different usable lengths for different operations. Report context performance by task, not as one model-wide number, and reject synthetic probes whose generator or scorer produces internally inconsistent difficulty.
\end{findingbox}

Reporting context performance by task raises the next question: at what length does each task cross an acceptance threshold?

\chapter{Effective Context Depends on Task and Model Scale}

\begin{figure}
\centering
\includegraphics[width=\linewidth]{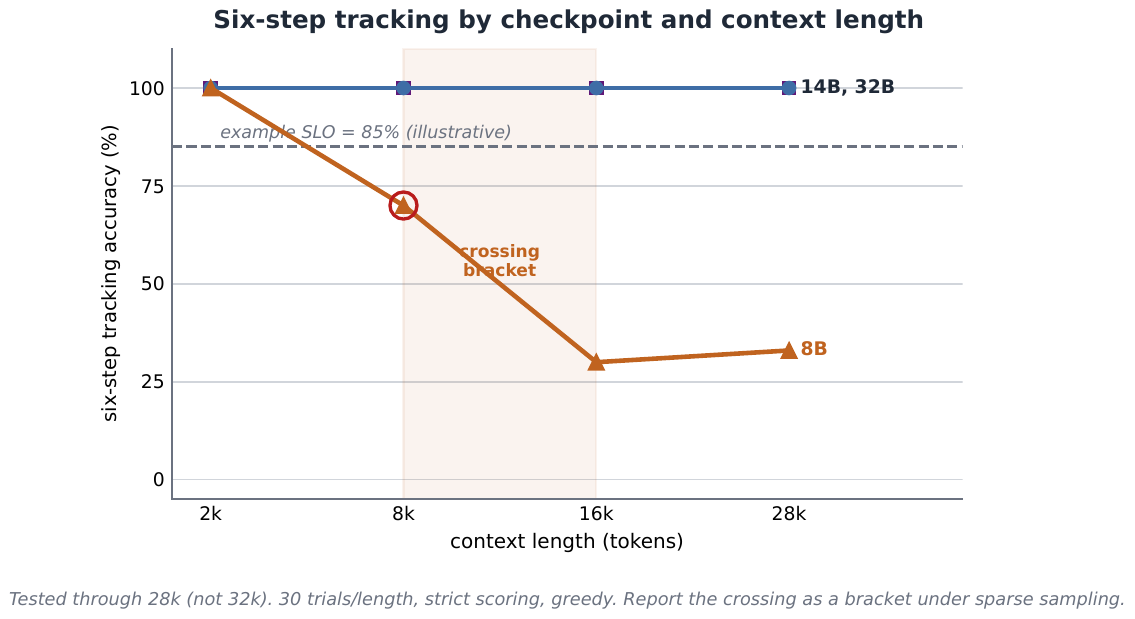}
\caption{\textbf{Six-step tracking by checkpoint and context length.} Qwen3-8B declined by 16k, while Qwen3-14B and Qwen3-32B remained correct through the tested 28k point. Effective context is obtained only after an acceptance threshold is specified; with sparse length sampling, the crossing should be reported as a bracket. Thirty trials per length, strict scoring, greedy decoding. The 8k tracking point is omitted. Source: E23c.}
\end{figure}

\medskip{}

A model's supported context length is not the same as its useful operating length. The first is an input and implementation limit. The second depends on the task, metric, acceptance threshold, checkpoint, hardware, engine, and serving configuration.

This chapter applies that definition to six-step variable tracking across three Qwen3 checkpoints. The experiment used approximately 2k, 8k, 16k, and 28k contexts, 30 trials per length, greedy decoding, and strict extraction of the answer after the designated marker. The comparison asks how the task-length curve changes across checkpoints; it does not isolate parameter count as the only causal factor.

\section{22.1 Define effective context before reporting it}

For a task $T$, operating condition $C$, metric $M$, and acceptance threshold $\tau$, define effective context as the longest evaluated length at which $M(T, C)$ remains at or above $\tau$. When only a small number of lengths are tested, report the threshold crossing as a bracket.

For example, if a checkpoint scores above the target at 8k and below it at 16k, the evidence supports ``the crossing lies between 8k and 16k.'' It does not support an exact 10k or 12k value unless intermediate lengths are measured or a justified interpolation is used. This definition also means that one checkpoint can have several effective-context values: retrieval, aggregation, and chained reasoning can cross the same threshold at different lengths, as Chapter 21 showed.

\section{22.2 The tested Qwen3 checkpoints produced different curves}

Under the six-step variable-tracking protocol, Qwen3-8B scored 30/30 at 2k, 9/30 at 16k, and 10/30 at 28k (with the 8k point omitted). Qwen3-14B and Qwen3-32B both scored 30/30 at every tested length through 28k.

The result supports a checkpoint-specific statement: in this Qwen3 family and on this task, the 8B checkpoint's score declined inside the tested window, while the 14B and 32B checkpoints remained correct through 28k. The comparison does not establish a universal parameter-count law, and it does not show performance at the untested 32k endpoint.

\begin{table}[htb]\centering\TableFontNine\TableRowAir
\TableTitle{Six-step tracking by checkpoint and context length}
\begin{tabularx}{\linewidth}{@{}C C C C@{}}
\arrayrulecolor{accent}\toprule
\thd{Context} & \thd{Qwen3-8B} & \thd{Qwen3-14B} & \thd{Qwen3-32B}\\[-4pt]
\thu{tokens} & \thu{trials correct} & \thu{trials correct} & \thu{trials correct}\\
\arrayrulecolor{rule}\midrule
2k  & 30/30 & 30/30 & 30/30\\
\zrow 16k & 9/30  & 30/30 & 30/30\\
28k & 10/30 & 30/30 & 30/30\\
\arrayrulecolor{accent}\bottomrule
\end{tabularx}
\numcaptab{Six-step tracking by checkpoint and context length. Only the 8B checkpoint declined inside the tested window; the 14B and 32B held the tested range.}
\TableNote{RULER-style \autocite{hsieh2024ruler} six-step variable tracking; strict match; greedy decoding; 30 trials per length; vLLM 0.11 for the dense runs. The 8k point is omitted; the 8B decline is bracketed between 2k and 16k. Evidence: E23c.}
\end{table}

\section{22.3 Thresholds convert curves into deployment decisions}

The same curve supports different operating ranges under different SLOs. A system that requires at least 85\% accuracy will use a shorter range than one that can tolerate 50\%. The threshold belongs in the result. The reporting form below distinguishes an observed range from a claimed absolute limit: ``at least 28k'' means the test did not fail through 28k, and ``crossing bracketed between'' means the exact crossing was not measured.

\begin{table}[htb]\centering\TableFontNine\TableRowAir
\TableTitle{Required reporting form for a threshold crossing}
\begin{tabularx}{\linewidth}{@{}Q{1.24in} C >{\RaggedRight\arraybackslash}X@{}}
\arrayrulecolor{accent}\toprule
\thd{Task and checkpoint} & \thd{Last pass / first fail} & \thd{Reported range}\\
\arrayrulecolor{rule}\midrule
Tracking, Qwen3-8B & 2k / 16k & Crossing bracketed between the 2k pass and the 16k decline\\
\zrow Tracking, Qwen3-14B & 28k / not observed & At least 28k under the tested protocol\\
Tracking, Qwen3-32B & 28k / not observed & At least 28k under the tested protocol\\
\arrayrulecolor{accent}\bottomrule
\end{tabularx}
\numcaptab{Required reporting form for a threshold crossing, at an example acceptance threshold of 85\%. Report a measured bracket, not a point estimate that was never measured.}
\TableNote{All three rows are six-step variable tracking. Evidence: E23c.}
\end{table}

\section{22.4 Memory feasibility is a separate gate}

The first Qwen3-32B launch at a configured 32k maximum failed because the engine could not reserve the required KV memory under that allocation. The engine estimated a maximum of approximately 12{,}208 tokens for that configuration. The run completed only after two serving settings were relaxed. The context-length limit came down from 32k to 30k, and the share of device memory the engine was allowed to hold rose from the default to 0.92. Both are vLLM engine arguments \autocite{vllm2025engineargs}; they appear in a run record as \code{max\_model\_len} and \code{gpu\_memory\_utilization}.

The initial estimate was therefore a configuration-specific allocation result, not a final model limit. This episode illustrates a separate deployment gate: a checkpoint may have adequate task performance at a length that a particular serving configuration cannot support at the required concurrency. Capability and memory feasibility should be reported separately.

\begin{figure}
\centering
\includegraphics[width=\linewidth]{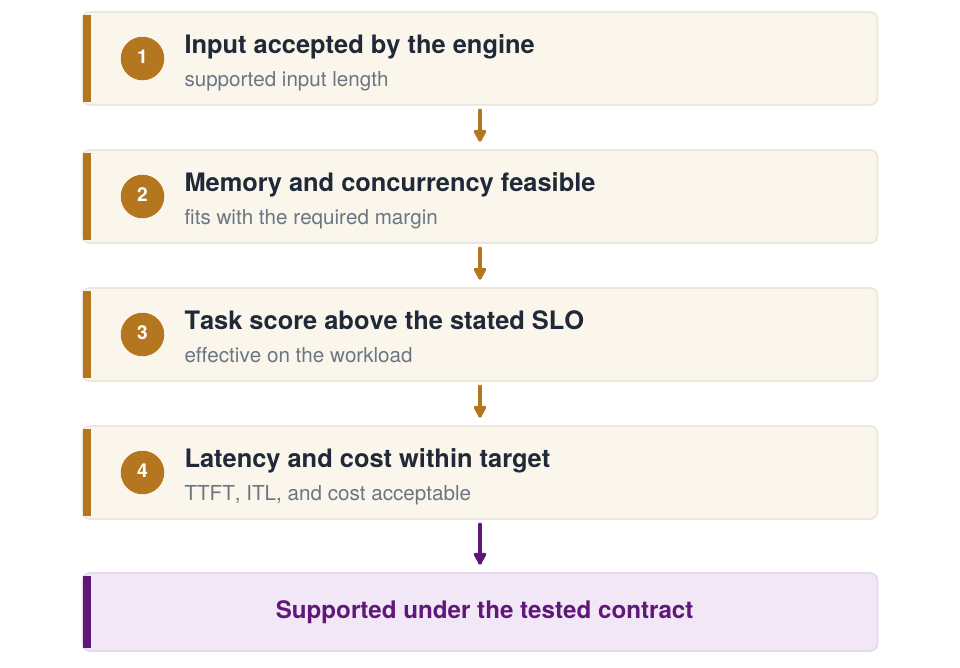}
\caption{\textbf{Four deployment gates for a target context length.} Input acceptance, memory and concurrency feasibility, task score above the SLO, and latency/cost within target are separate gates. A failure at any gate calls for a different response. A decision workflow, not a measured result.}
\end{figure}

\section{22.5 Select the response that matches the failed gate}

When retrieval fails, revise retrieval, indexing, or prompt construction. When aggregation or reasoning fails, options include a stronger checkpoint, retrieval that moves relevant evidence into a shorter span, task decomposition, context reduction, structured external state, or a different architecture with a better measured profile. When memory is the limit, adjust concurrency, precision, memory reservation, parallelism, or model choice. When latency is the limit, evaluate chunking and scheduling. No single remedy applies to every context failure; the first task is to identify which gate failed.

\begin{findingbox}Effective context is a task--metric--threshold result, not a property printed once for a model. On six-step variable tracking, Qwen3-8B declined sharply by 16k, while Qwen3-14B and Qwen3-32B remained correct through the tested 28k range. Report the threshold crossing as a measured bracket, and evaluate task performance, memory feasibility, and latency as separate gates.
\end{findingbox}

A different architecture can shift several of these gates at once, which is where the next chapter begins.

\chapter{Reasoning, KV Capacity, and Throughput in a Hybrid-Attention Checkpoint}

\begin{figure}
\centering
\includegraphics[width=\linewidth]{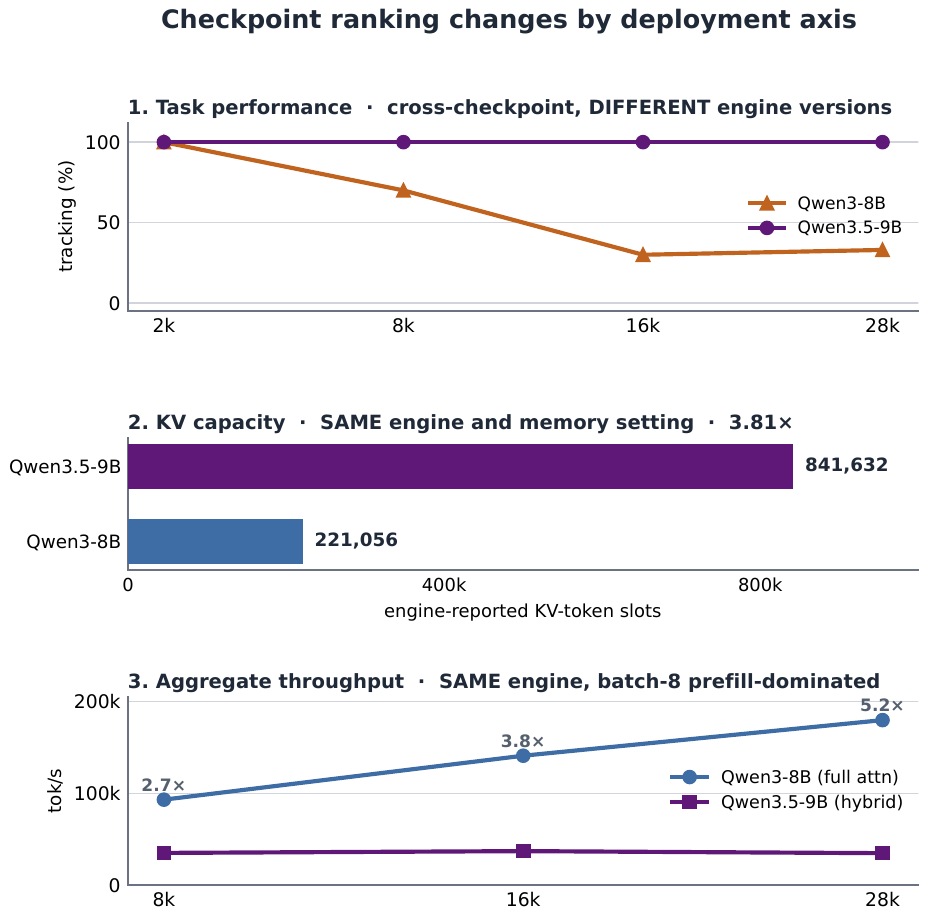}
\caption{\textbf{The checkpoint ranking changes by deployment axis.} Qwen3.5-9B maintained the tested six-step tracking task through 28k and exposed 3.81$\times$ more KV-token capacity under the same vLLM version and memory setting. Qwen3-8B delivered higher aggregate batch-8 prefill throughput at 8k, 16k, and 28k. The reasoning comparison used different engine versions and is an operational checkpoint comparison, not an architecture ablation. Source: E28 and hardening.}
\end{figure}

\medskip{}

The tested hybrid checkpoint is Qwen3.5-9B, documented by the Qwen team as part of the Qwen3.5 family \autocite{qwen2026qwen35}. Its linear-recurrent component draws on Gated Delta Networks \autocite{yang2025gateddelta}. Qwen3.5-9B and Qwen3-8B were compared on four deployment axes: six-step tracking across long contexts, engine-reported KV capacity, aggregate batch-8 throughput, and prefix-cache impact. The results do not form a single ranking. The Qwen3.5 checkpoint had substantially larger KV capacity and a stronger long-context tracking curve, while the Qwen3 checkpoint had higher aggregate prefill throughput in the same-engine micro-benchmark.

These are operational checkpoint comparisons. The models differ in architecture, generation, training, parameter count, tokenizer behavior, and implementation support. The reasoning runs also used different vLLM versions. The chapter therefore reports the measured profile and identifies which comparisons are controlled, rather than attributing every difference to hybrid attention alone.

\section{23.1 Long-context tracking differed between the checkpoints}

On six-step variable tracking, Qwen3.5-9B scored 30/30 at 2k, 8k, 16k, and 28k. Qwen3-8B scored 30/30 at 2k, 9/30 at 16k, and 10/30 at 28k (with the 8k point omitted). This result is relevant to checkpoint selection: under the tested protocols, Qwen3.5-9B maintained performance through 28k while Qwen3-8B did not. It is not an architecture ablation. The dense and hybrid runs used different engine versions, and the checkpoints differ in more than their attention layers.

\section{23.2 Same-engine KV-capacity comparison}

Both checkpoints were loaded on one H100 under vLLM 0.23 with the same 0.60 GPU-memory setting. The engine reported 221{,}056 KV-token slots for Qwen3-8B and 841{,}632 for Qwen3.5-9B, a ratio of approximately 3.81$\times$. Dividing those pools by a 16k sequence length gives derived equal-length bounds of 13 and 52 sequences. These are capacity bounds rather than a measured service-level result: production concurrency also requires memory margin and must satisfy latency and preemption targets.

\begin{table}[htb]\centering\TableFontNine\TableRowAir
\TableTitle{Same-engine KV capacity}
\begin{tabularx}{\linewidth}{@{}Q{1.08in} C C C@{}}
\arrayrulecolor{accent}\toprule
\thd{Checkpoint} & \thd{Attention} & \thd{KV capacity} & \thd{Derived bound}\\[-4pt]
 & & \thu{token slots} & \thu{16k sequences}\\
\arrayrulecolor{rule}\midrule
Qwen3-8B & full & 221{,}056 & 13\\
\zrow Qwen3.5-9B & hybrid & 841{,}632 & 52\\
\arrayrulecolor{accent}\bottomrule
\end{tabularx}
\numcaptab{Same-engine KV capacity, the most tightly controlled comparison in the chapter. Sequence counts are floor-divided equal-length bounds, not SLO-qualified production concurrency.}
\TableNote{One H100; vLLM 0.23; identical GPU-memory setting of 0.60; engine-reported cache pool. Evidence: E28 hardening.}
\end{table}

The larger pool is consistent with the hybrid checkpoint's use of recurrent state in many layers rather than a full per-token KV entry in every layer. Because the checkpoints are not an architecture ablation, the mechanism is presented as interpretation supported by the model design and engine accounting, not as a fully isolated causal estimate.

\section{23.3 Same-engine aggregate prefill throughput}

The same-engine throughput audit used vLLM 0.23, batch 8, long prompts, and short generation. It was dominated by prefill work. Qwen3-8B processed 93{,}115 tokens per second at 8k, 140{,}975 at 16k, and 179{,}737 at 28k. Qwen3.5-9B processed 35{,}114, 37{,}055, and 34{,}835 tokens per second at the same lengths.

\begin{table}[htb]\centering\TableFontNine\TableRowAir
\TableTitle{Same-engine aggregate throughput}
\begin{tabularx}{\linewidth}{@{}C C C C@{}}
\arrayrulecolor{accent}\toprule
\thd{Context} & \thd{Qwen3-8B} & \thd{Qwen3.5-9B} & \thd{Ratio}\\[-4pt]
\thu{tokens} & \thu{tok/s} & \thu{tok/s} & \thu{full / hybrid}\\
\arrayrulecolor{rule}\midrule
8k  & 93{,}115  & 35{,}114 & 2.65$\times$\\
\zrow 16k & 140{,}975 & 37{,}055 & 3.80$\times$\\
28k & 179{,}737 & 34{,}835 & 5.16$\times$\\
\arrayrulecolor{accent}\bottomrule
\end{tabularx}
\numcaptab{Same-engine aggregate throughput at batch 8 on a prefill-dominated workload; the full-attention checkpoint was faster throughout. These are aggregate processed tokens per second, not per-request latency or decode-only throughput.}
\TableNote{vLLM 0.23 for both checkpoints; batch 8; prefill-dominated workload. Evidence: E28 hardening.}
\end{table}

The result does not support a hybrid throughput advantage under this workload, and it should not be generalized to all serving conditions. Separate measurements are required for TTFT, decode throughput, request rate, and mixed-length online traffic.

\section{23.4 Prefix caching produced the lowest drift on the hybrid checkpoint, not zero drift}

In the hardened five-slice cache comparison, Qwen3.5-9B showed \pmv{33.9}{2.7}\% text divergence between cache-off and cache-on generations, an accuracy delta of \pmv{0.00}{0.55} percentage points, and a \pmv{1.1}{0.5}\% correct-to-wrong transition rate. These were the lowest divergence and transition estimates among the three tested checkpoints. The result supports enabling prefix caching when its workload benefit is material and task-level impact remains within the deployment tolerance; it does not support a claim of bitwise equivalence or zero item-level risk.

\section{23.5 Deployment decision: select by workload axis}

For long-context workloads, compare at least four axes separately: task performance across the actual context distribution; KV capacity and SLO-qualified concurrency; TTFT, decode latency, and aggregate throughput under the real arrival process; and cache and reproducibility requirements. Under the tested conditions, Qwen3.5-9B was favorable on six-step tracking and KV capacity, while Qwen3-8B was favorable on aggregate batch-8 prefill throughput. The correct choice depends on which constraint dominates the workload.

\begin{table}[htb]\centering\TableFontEightSix\TableRowAir
\TableTitle{Evidence status by axis}
\begin{tabularx}{\linewidth}{@{}Q{1.10in} >{\hsize=0.93\hsize\RaggedRight\arraybackslash}X >{\hsize=1.07\hsize\RaggedRight\arraybackslash}X@{}}
\arrayrulecolor{accent}\toprule
\thd{Axis} & \thd{Comparison control} & \thd{Allowed conclusion}\\
\arrayrulecolor{rule}\midrule
Six-step tracking & Different checkpoints and engine versions & Qwen3.5-9B performed better under the tested protocols\\
\zrow KV capacity & Same engine, hardware, memory setting & Qwen3.5-9B exposed 3.81$\times$ more KV capacity\\
Prefill throughput & Same engine and workload & Qwen3-8B was faster in this \mbox{micro-benchmark}\\
\zrow Prefix-cache impact & Same framework, five disjoint slices & Qwen3.5-9B drifted least; nonzero flips remained\\
\arrayrulecolor{accent}\bottomrule
\end{tabularx}
\numcaptab{Evidence status by axis. Only the KV-capacity and throughput rows are same-engine controlled comparisons.}
\TableNote{Evidence: E28, E41, hardening.}
\end{table}

\begin{findingbox}The hybrid checkpoint exposed 3.81$\times$ the engine-reported KV-token capacity and a derived 4$\times$ 16k sequence bound under the same vLLM version and memory setting, and maintained six-step tracking through 28k. The full-attention checkpoint delivered 2.65$\times$ to 5.16$\times$ higher aggregate prefill throughput in the same-engine batch-8 audit. Report these axes separately, and avoid attributing checkpoint-level differences to architecture without a matched ablation.
\end{findingbox}

Mixture-of-experts models raise the same separation of axes, with memory and routing added to the list.

\chapter{Mixture-of-Experts Models: Memory, Quantization, and Reproducibility}

\begin{figure}
\centering
\includegraphics[width=\linewidth]{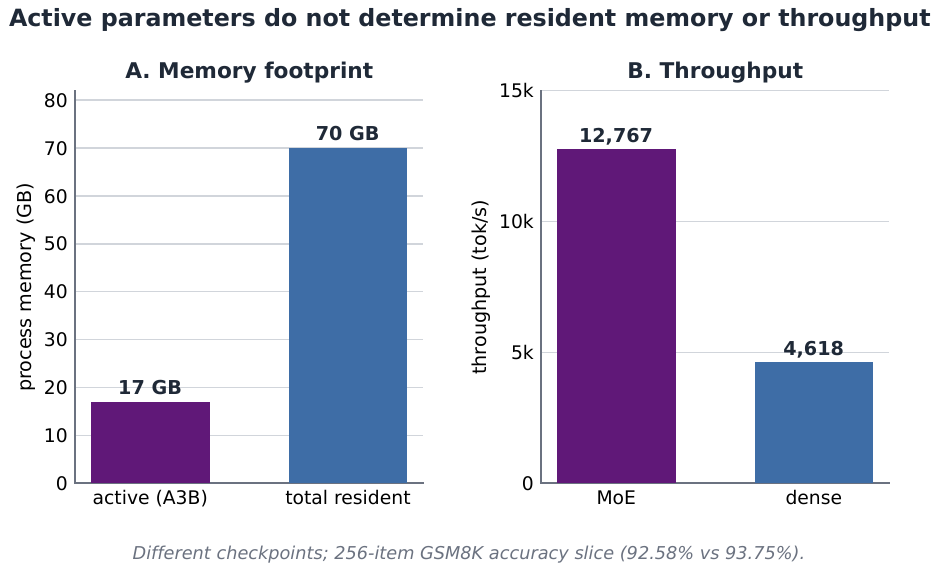}
\caption{\textbf{Active parameters do not set memory or throughput.} Qwen3-30B-A3B (3B active, 30B total) used roughly 70\,GB of observed process memory and 4{,}618 tok/s, versus roughly 17\,GB and 12{,}767 tok/s for Qwen3-4B, while scoring 93.75\% vs 92.58\% on a 256-item GSM8K slice. Different checkpoints; not a matched ablation. Source: E-MoE.}
\end{figure}

\medskip{}

Mixture-of-experts models report both total and active parameter counts. The active count describes the subset of expert computation selected for a token. It does not determine resident memory, achieved throughput, quantization behavior, or exact reproducibility.

This chapter combines three studies. The first compares Qwen3-30B-A3B with Qwen3-4B on a GSM8K slice, throughput, and observed process memory. The second compares BF16 and AWQ across knowledge, math, and code tasks. The third uses direct router instrumentation to measure how an AWQ checkpoint changes expert selection relative to BF16. The results are checkpoint- and implementation-specific.

\section{24.1 Active parameters do not determine resident memory or throughput}

Qwen3-30B-A3B activates approximately 3B parameters per token but contains approximately 30B total parameters. Under the tested vLLM configuration, the process used roughly 70\,GB of GPU memory, compared with roughly 17\,GB for Qwen3-4B. These figures include engine allocations and should be read as observed process residency, not pure weight size. On a 256-item GSM8K slice, the MoE checkpoint scored 93.75\% and the dense checkpoint scored 92.58\%, a difference of 1.17 percentage points. Throughput moved in the opposite direction: 4{,}618 tokens per second for the MoE checkpoint and 12{,}767 for the dense checkpoint.

\begin{table}[htb]\centering\TableFontNine\TableRowAir
\TableTitle{Operational comparison at similar active-parameter scale}
\begin{tabularx}{\linewidth}{@{}Q{1.08in} C C C C@{}}
\arrayrulecolor{accent}\toprule
\thd{Checkpoint} & \thd{Parameters} & \thd{GSM8K} & \thd{Throughput} & \thd{Memory}\\[-4pt]
 & \thu{total / active} & \thu{percent} & \thu{tok/s} & \thu{GB resident}\\
\arrayrulecolor{rule}\midrule
Qwen3-4B & 4B dense & 92.58 & 12{,}767 & 17\\
\zrow Qwen3-30B-A3B & 30B / 3B & 93.75 & 4{,}618 & 70\\
\arrayrulecolor{accent}\bottomrule
\end{tabularx}
\numcaptab{Operational comparison at similar active-parameter scale. The MoE checkpoint used roughly four times the memory and delivered lower throughput than a dense model of similar active size. A checkpoint comparison, not a matched architecture ablation.}
\TableNote{Memory figures are approximate process residency read from the GPU management tool, and include engine allocations as well as weights. vLLM 0.11; greedy decoding; GSM8K-256; one H100 per checkpoint. Evidence: E-MoE.}
\end{table}

Size model residency from total stored parameters and measured engine overhead. Active parameters are useful for describing selected arithmetic, but they do not provide a memory budget or a throughput prediction.

\section{24.2 Quantization effects were checkpoint- and task-dependent}

Qwen3-14B AWQ was approximately unchanged on MMLU and GSM8K but 3.04 percentage points lower on HumanEval. Qwen3-30B-A3B AWQ was lower on all three tasks by 1.8 to 2.6 points. A Mistral-7B AWQ replication showed small positive differences on all three tasks and did not reproduce the Qwen dense pattern.

\begin{table}[htb]\centering\TableFontNine\TableRowAir
\TableTitle{BF16-to-AWQ task deltas}
\begin{tabularx}{\linewidth}{@{}Q{1.04in} C C C >{\RaggedRight\arraybackslash}X@{}}
\arrayrulecolor{accent}\toprule
\thd{Checkpoint} & \thd{MMLU} & \thd{GSM8K} & \thd{HumanEval} & \thd{Pattern}\\[-4pt]
 & \thu{pp} & \thu{pp} & \thu{pp} & \\
\arrayrulecolor{rule}\midrule
Qwen3-14B dense & \textminus0.70 & +0.61 & \textminus3.04 & Code-only loss\\
\zrow Qwen3-30B-A3B MoE & \textminus2.6 & \textminus1.9 & \textminus1.8 & Broad degradation\\
Mistral-7B dense & +4.9 & +2.6 & +3.0 & Did not replicate\\
\arrayrulecolor{accent}\bottomrule
\end{tabularx}
\numcaptab{BF16-to-AWQ task deltas in percentage points. A pattern audit across three checkpoints, not a controlled three-way ranking: the Mistral non-replication is visible, so the Qwen dense pattern should not be treated as the general baseline.}
\TableNote{Conditions and sample sizes differ across studies; the Mistral deltas are small and within noise. Validate quantization for the exact checkpoint, quantizer, calibration, engine, and served tasks. Evidence: E8, E-MoE, H2-E8.}
\end{table}

\section{24.3 Routing sensitivity under quantization}

The Qwen3-30B-A3B router-forensics study compared BF16 and AWQ token by token. Approximately 31.0\% of top-1 expert selections changed, and approximately 86.8\% of top-$k$ expert sets changed in at least one position. The router gate weights themselves were not quantized; the route differences arose because quantization changed upstream activations, which changed the inputs presented to later router gates. Routing can therefore change even when the gate weights remain at higher precision.

\begin{figure}
\centering
\includegraphics[width=\linewidth]{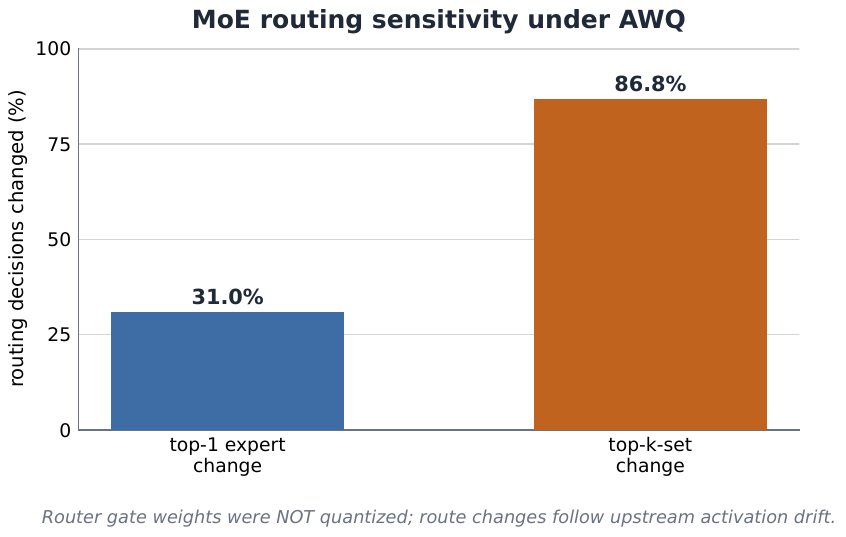}
\caption{\textbf{Routing changes under AWQ.} Direct token-by-token comparison found 31.0\% of top-1 expert selections and 86.8\% of top-$k$ expert sets changed between BF16 and AWQ, although the router gate weights were not quantized. The changes follow upstream activation drift. Source: RL-rollout router forensics.}
\end{figure}

\begin{table}[htb]\centering\TableFontNine\TableRowAir
\TableTitle{Direct router-forensics measurements}
\begin{tabularx}{\linewidth}{@{}Q{1.28in} C >{\RaggedRight\arraybackslash}X@{}}
\arrayrulecolor{accent}\toprule
\thd{Measurement} & \thd{Result} & \thd{What is directly established}\\
\arrayrulecolor{rule}\midrule
Top-1 expert change & 31.0\% & The top expert differed frequently\\
\zrow Top-$k$ set change & 86.8\% & At least one expert differed in most cases\\
Router gate quantized & no & Route changes followed activation drift\\
\zrow Routing entropy & 3.94 / 3.61 & AWQ routing was somewhat more concentrated\\
\arrayrulecolor{accent}\bottomrule
\end{tabularx}
\numcaptab{Direct router-forensics measurements. Route changes were measured directly; the task-level cause was not fully isolated. Routing entropy is shown as BF16 / AWQ.}
\TableNote{Qwen3-30B-A3B, BF16 vs AWQ, token-by-token route logging; means across tokens and layers. Evidence: RL-rollout forensics.}
\end{table}

\section{24.4 Batch-context output forking generalized qualitatively}

The repeated Qwen sweep produced one distinct output for Qwen3-8B and two for Qwen3-30B-A3B under the hardened protocol; the original two-versus-four counts did not reproduce. A later gpt-oss-20b MoE replication produced a mean of 3.8 distinct outputs, range 3--4, across five repeats. The qualitative result is that MoE checkpoints can exhibit material batch-context-dependent output variation, and the effect was not confined to Qwen.

\begin{codebox}
\noindent{\sffamily\bfseries\footnotesize\color{accent}What was and was not instrumented.}\enspace{\sffamily\footnotesize Expert-route changes were measured directly in the BF16-versus-AWQ forensics. The batch-context experiment measured output forks but did not log expert assignments. Routing is therefore a supported interpretation of the batch result, not a direct trace from that experiment.}
\end{codebox}

\section{24.5 Deployment controls for sparse-expert models}

For an MoE checkpoint: calculate resident memory from total parameters and measured engine allocations; benchmark achieved throughput under the actual batch and parallelism regime; validate each quantized checkpoint on every served task family; record expert-parallel configuration and placement; measure batch-context text and task drift when reproducibility matters; and treat deterministic or replayable routing as an additional exact-replay requirement whose implementation cost must be measured.

\begin{findingbox}Qwen3-30B-A3B used roughly four times the observed GPU memory and delivered lower throughput than Qwen3-4B despite a similar active-parameter count, while scoring 1.17 points higher on a 256-item GSM8K slice. Its AWQ checkpoint showed broad 1.8--2.6-point degradation, and direct instrumentation found 31\% top-1 and 86.8\% top-$k$ routing changes even though router gate weights were not quantized. These justify separate memory, quality, and reproducibility gates for MoE deployment.
\end{findingbox}

Each of these comparisons raises the same question: what does the comparison design actually license one to claim?

\chapter{Controlled Comparisons for Architecture Claims}

\begin{figure}
\centering
\includegraphics[width=\linewidth]{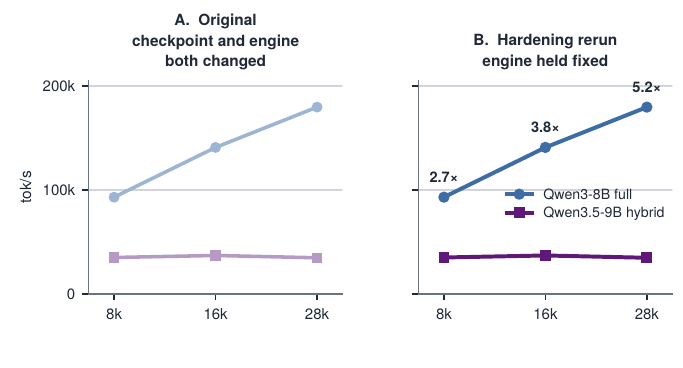}
\caption{\textbf{From boundary to controlled comparison.} The original hybrid-throughput comparison changed both checkpoint and engine version (a boundary result); the hardening rerun aligned the engine and produced a controlled operational comparison. Remaining differences: model generation, training, scale, tokenizer, and architecture. Source: E28.}
\end{figure}

\medskip{}

An architecture claim is only as strong as the comparison that produced it. Holding the inference engine fixed is necessary because kernels, schedulers, compilation, memory management, and defaults can change the result independently of the model. It is not sufficient when the checkpoints also differ in training, scale, tokenizer, model generation, or other components.

\section{25.1 Why cross-engine results are boundary evidence}

The first E28 run used vLLM 0.11 for Qwen3-8B and vLLM 0.23 for Qwen3.5-9B. Both the checkpoint and the execution stack changed. The resulting throughput difference could therefore reflect model design, engine version, kernel selection, compilation, scheduler behavior, or an interaction among them. The correct label for that result was a boundary: it showed that the two complete systems behaved differently, but it did not estimate an architecture effect.

\section{25.2 What the same-engine rerun established}

The hardening rerun placed both checkpoints on vLLM 0.23 and repeated the batch-8 prefill-dominated workload. Qwen3-8B produced 93{,}115 tok/s at 8k, 140{,}975 at 16k, and 179{,}737 at 28k; Qwen3.5-9B produced 35{,}114, 37{,}055, and 34{,}835 tok/s. The same-engine result establishes that Qwen3-8B had higher aggregate throughput than Qwen3.5-9B under this benchmark. It does not establish that full attention is universally faster than hybrid attention or that architecture alone caused the 2.65$\times$ to 5.16$\times$ ratios.

The same distinction applies to the KV-capacity result. Both checkpoints were measured under one engine and memory setting, so the 221{,}056-versus-841{,}632 token-pool difference is a controlled operational result. It is consistent with the hybrid model's state design, but it is not a paired architecture ablation.

\section{25.3 An evidence ladder for architecture comparisons}

Architecture comparisons should carry one of four labels. A same-engine test usually reaches level 2, not level 4.

\begin{figure}
\centering
\includegraphics[width=\linewidth]{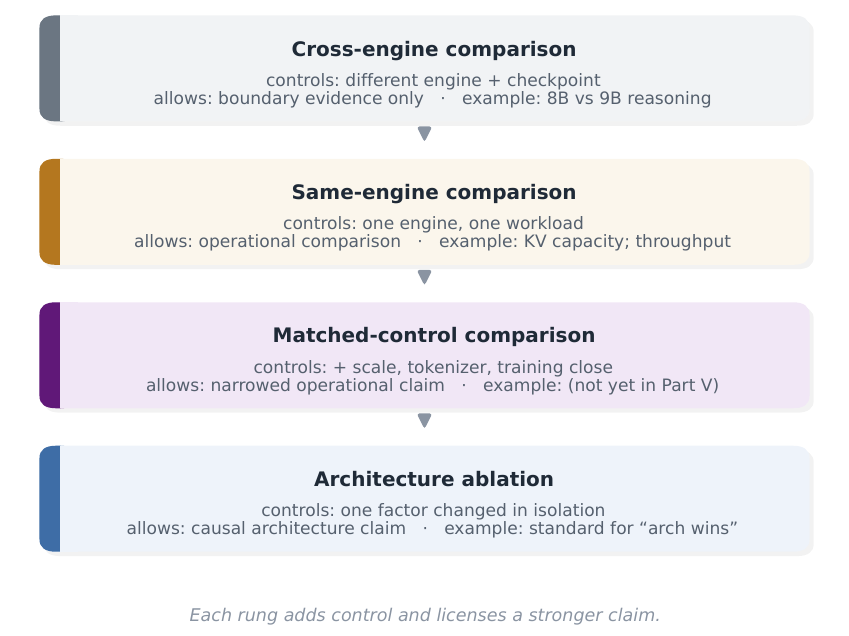}
\caption{\textbf{An evidence ladder for architecture claims.} Boundary (architecture and stack vary together), controlled operational (engine, hardware, workload, metric aligned), matched comparative (checkpoints aligned on family, scale, training, tokenizer), and causal ablation (the architectural component is varied while the rest is paired). A methodological workflow, not a measured result.}
\end{figure}

\begin{table}[htb]\centering\TableFontEightSix\TableRowAir
\TableTitle{Evidence audit for Part V}
\begin{tabularx}{\linewidth}{@{}>{\hsize=0.95\hsize\RaggedRight\arraybackslash}X Q{0.76in} >{\hsize=1.05\hsize\RaggedRight\arraybackslash}X@{}}
\arrayrulecolor{accent}\toprule
\thd{Claim} & \thd{Label} & \thd{Allowed conclusion}\\
\arrayrulecolor{rule}\midrule
Curves differ by operation & task profile & Operations differ in sensitivity\\
\zrow Larger Qwen3 held to 28k & matched family & The 14B and 32B held the range\\
Qwen3.5-9B led on tracking & boundary & Higher under the tested protocols\\
\zrow Qwen3.5-9B: 3.81$\times$ KV capacity & controlled & It exposed a larger pool\\
Qwen3-8B led on prefill & controlled & It was faster on this benchmark\\
\zrow Qwen MoE AWQ rerouted & mechanistic & Expert selections differed under AWQ\\
\arrayrulecolor{accent}\bottomrule
\end{tabularx}
\numcaptab{Evidence audit for Part V: each claim, what its comparison controls, and what it therefore permits. These are neutral evidence labels, not status tags.}
\TableNote{Evidence: E23c, E28, E-MoE, RL-rollout forensics.}
\end{table}

\section{25.4 Report the remaining confounds in the result sentence}

Use the following form: under [hardware, engine/version, workload, and metric], [checkpoint A] produced [result] and [checkpoint B] produced [result]; because the checkpoints also differ in [remaining factors], this is a [evidence label] comparison rather than an isolated architecture effect. This wording makes the result useful for deployment while preserving the scientific boundary.

\section{25.5 Part V conclusion}

Long-context deployment is a collection of separate contracts. The engine must accept and allocate the target length. The checkpoint must meet the task-specific quality threshold. The serving system must satisfy concurrency, latency, and cost targets. Architecture comparisons must state which of those quantities were measured and which factors were controlled.

\begin{table}[htb]\centering\TableFontEightSix\TableRowAir
\TableTitle{Part V in one table: axes and results}
\begin{tabularx}{\linewidth}{@{}Q{0.98in} >{\hsize=0.93\hsize\RaggedRight\arraybackslash}X >{\hsize=1.07\hsize\RaggedRight\arraybackslash}X@{}}
\arrayrulecolor{accent}\toprule
\thd{Axis} & \thd{Measurement contract} & \thd{Part V result}\\
\arrayrulecolor{rule}\midrule
Retrieval & Fixed placement, exact scoring & Correct through 28k\\
\zrow Aggregation & Distributed evidence, task metric & 90\% or above through 28k\\
Chained reasoning & Workload-matched task and SLO & 8B declined by 16k; larger held\\
\zrow KV capacity & Same engine, memory setting, concurrency & Qwen3.5-9B exposed 3.81$\times$ more\\
Throughput & Same engine, precise workload metric & Qwen3-8B led on batch-8 prefill\\
\zrow MoE quantization & Exact checkpoint, quantizer, route logging & AWQ lost broadly and rerouted\\
\arrayrulecolor{accent}\bottomrule
\end{tabularx}
\TableCaption{\textbf{Table V.1.} Part V in one table: the measurement contract for each long-context and architecture axis, and the result it produced.}
\TableNote{Synthesis of Chapters 21--25. Evidence: E23c, E28, E-MoE.}
\end{table}

\begin{findingbox}Same-engine testing removed the software-version confound from the hybrid-throughput comparison and showed that Qwen3-8B had higher aggregate throughput under the tested batch-8 workload. The result is a controlled checkpoint comparison, not a universal full-attention-versus-hybrid law. Architecture claims should carry an evidence label and state the remaining checkpoint differences in the result sentence.
\end{findingbox}

Long-context deployment requires separate contracts for input support, memory feasibility, task-level quality, latency, and cost. Architecture comparisons add a second requirement: the comparison design must support the strength of the claim. The next part converts those contracts into run manifests, checklists, formulas, and evidence labels for day-to-day use.

\part{The Field Guide}

\begin{partopener}
Turn the preceding evidence into an auditable run record, a preflight, a calculation, and a release decision.
\end{partopener}

The preceding chapters present the evidence and its boundaries. This part reorganizes the same material for operational use. Chapter 26 defines the run manifest required for audit and rerun. Chapter 27 condenses evaluation controls into a checklist. Chapter 28 collects the capacity, cost, and latency formulas. Chapter 29 records which findings are canonical, model-specific, version-specific, or still bounded by unresolved conditions.

The appendices are the reference layer for those chapters: Appendices~\ref{app:roster} and \ref{app:workloads} identify the printed checkpoint and workload roster; Appendices~\ref{app:reproduction} and \ref{app:evidence} connect reproduction and evidence status back to the chapter claims; Appendices~\ref{app:energy} through \ref{app:modalities} extend the measurement contracts to energy, speculative decoding, multimodal input, and embeddings; and Appendix~\ref{app:operations} provides the operational failure-mode index.

Table~VI.1 is a symptom index for that material. Use it after something has gone wrong and two runs need to be compared. Find the matching symptom, retrieve the records named in the middle column from both runs, and compare them field by field. The final column points to the part, chapter, or appendix that defines the relevant measurement contract.

The router narrows the search. It does not assign a cause: a difference in one of the listed records is the starting point for an investigation, not its conclusion.

\begin{statictable}\TableFontEightSix\TableRowAir
\TableTitle{Initial diagnostic router}
\begin{tabularx}{\linewidth}{@{}Q{1.08in} >{\RaggedRight\arraybackslash}X Q{0.84in}@{}}
\arrayrulecolor{accent}\toprule
\thd{Symptom} & \thd{Records to compare between the two runs} & \thd{Contract defined in}\\
\arrayrulecolor{rule}\midrule
Score changed unexpectedly & Sample IDs and order; budget and truncation; extractor and scorer; template and harness; decoding & Part I\\
\zrow Greedy output differs across runs & Model and tokenizer revisions; engine version; attention backend; batch trace; cache state; precision & Part II\\
Throughput or cost differs & Workload trace; prompt and output lengths; batch occupancy; concurrency; engine path & Part III; App.\ H\\
\zrow Tail latency exceeds the SLO & Arrival process; offered versus achieved load; chunked prefill; maximum concurrency; queue metrics & Part III; App.\ H\\
Structured output valid but wrong & Schema enforcement; parser; task correctness; argument values; abstention cases & Part IV\\
\zrow Retrieval passes, workload fails & Task-specific context curve; threshold; context distribution; memory feasibility & Part V\\
Allocation fails or preemption rises & Total model residency; KV-token pool; context $\times$ concurrency; memory margin; engine memory share & Ch.\ 14; App.\ H\\
\arrayrulecolor{accent}\bottomrule
\end{tabularx}
\TableCaption{\textbf{Table VI.1.} Initial diagnostic router. For each of seven common symptoms: the run records to pull from both runs and compare, and where in the book the measurement contract for those records is defined.}
\TableNote{Use after a difference has been observed, not as a preflight checklist; Chapter 27 covers preflight. Operational symptoms---throughput or cost anomalies, tail-latency breaches, and allocation or preemption failures---also route to Appendix H, which catalogs the runtime failure modes and the check that separated each from model behavior. Field-guide synthesis of Parts I--V, Chapter 14, and Appendix H.}
\end{statictable}

\clearpage

\chapter{Recording an Auditable Inference Run}

A reported inference result should be accompanied by a run record: a structured description of the condition that produced it. When that record is stored in machine-readable form it is often called a manifest. It preserves model identity, software and hardware environment, serving and decoding configuration, evaluation contract, and links to the raw artifacts.

The run record serves three purposes. It makes the result auditable, allows the intended condition to be rerun, and makes differences between runs explicit. It does not guarantee byte-identical replay. Exact replay may additionally require deterministic operations, request and batch traces, cache history, random states, routing traces, and an environment image.

\section{26.1 Record identity and provenance}

Use immutable model and tokenizer revisions, not only repository names. Hash the chat template and prompt template, because formatting changes can alter tokenization and model behavior. Record the dataset revision, exact sample IDs, order, few-shot examples, and generator version. Store hashes for raw prompts, outputs, per-item scores, aggregate metrics, and engine logs. A run ID, timestamp, experiment ID, code commit, and evidence-status label connect the run record to the book's evidence ledger. The status should describe the result (measured, hardened, boundary, or external), not the confidence of the author.

\section{26.2 Record the software and hardware substrate}

The engine name alone is insufficient. Record its version and commit, the container digest, PyTorch, CUDA runtime, GPU driver, compiler or graph mode, and any explicit kernel backend. Record GPU model, count, and topology, along with tensor, pipeline, data, and expert parallelism. These fields are required because a numerically identical high-level configuration can execute through different kernels, schedulers, and communication paths across versions or hardware.

\section{26.3 Record the execution path}

Record weight, activation, and KV-cache dtypes; quantization method and configuration; attention backend; the context-length limit; the engine's share of device memory; maximum sequences; the scheduler token budget; chunked prefill; prefix-cache state; speculative decoding; scheduler policy; and full decoding parameters. A scheduler label such as ``continuous'' does not reconstruct the batch shapes that actually occurred; when exact replay or a nondeterminism investigation matters, preserve the request-arrival trace, request order, batch trace, cache initial state, and routing trace if the engine exposes one.

Record the seed for every run. It is necessary for sampled decoding and may affect other stochastic components. Under greedy decoding, however, it does not control batch-, backend-, cache-, or kernel-dependent numerical divergence. It is necessary but insufficient for exact replay.

\begin{bookconfig}{yaml}
# Illustrative run-record schema. Values are examples.
run_id:         long_context_tracking_qwen3_8b_001
study_id:       long_context_tracking_v1
record_status:  illustrative
identity:
  model_revision:  Qwen/Qwen3-8B@a1b2c3d
  tokenizer_rev:   a1b2c3d
  chat_template_sha256: e3b0c44298fc…
  dataset_rev:     gsm8k@d4e5f6a
  sample_ids_sha256: 7d865e959b24…
software_hardware:
  engine:    {name: vllm, version: 0.11.0, commit: b2c3d4e}
  container: sha256:9f2b7c1a4e0d…
  torch: 2.x   cuda: 12.x   driver: 550.90.07
  graph_mode: cudagraph
  gpu: {model: H100-80GB, count: 1, tp: 1, pp: 1, ep: 1}
execution_path:
  weight_dtype: bf16   kv_dtype: auto
  quantization: none   attention_backend: flash-attn
  max_model_len: 32768        gpu_memory_utilization: 0.85
  max_num_batched_tokens: 8192   chunked_prefill: true
  prefix_cache: off   scheduler: continuous
  decoding: {mode: greedy, temperature: 0, max_new_tokens: 1024}
  seed: 0            # necessary, not sufficient under greedy
evaluation:
  extractor: last_number_regex@v2   scorer: exact_match
  aggregation: mean   uncertainty: bootstrap_1000
artifacts:
  raw_generations_sha256:  2c26b46b68ff…
  per_item_scores_sha256:  fcde2b2edba5…
\end{bookconfig}

The field names and values above are illustrative. This volume does not distribute a schema implementation, code package, run records, or referenced artifacts; the example defines the information that a future release bundle should preserve.

\section{26.4 Record the evaluation contract}

The run record must state how text became a score. Record the extractor, scorer, metric definition, correctness tolerance, exclusion rules, aggregation method, and uncertainty calculation. For LLM-judge evaluations, record the judge model, revision, prompt, decoding, and order policy. For tool calling, separate parse validity, semantic correctness, and abstention. The evaluation contract belongs in the run record because a run can be generated correctly and scored differently by a changed parser or harness.

\section{26.5 Store the run record with the raw artifacts}

Create the run record at launch, finalize it after the run with artifact hashes, and validate it against a schema in continuous integration. The run record should be stored beside the raw generations and metrics, and the reported table or figure should reference the run ID. Recording a field and choosing its value are different actions: documentation has negligible runtime cost, while restrictive execution choices (fixed batches, cache disabled, eager mode, deterministic kernels, or recorded/replayed routing) may change throughput or latency and must be benchmarked under the target workload.

\begin{figure}
\centering
\includegraphics[width=\linewidth]{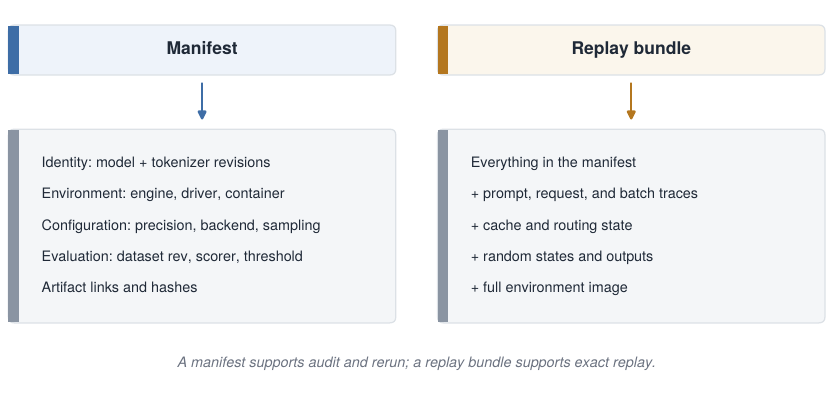}
\caption{\textbf{Manifest versus replay bundle.} A manifest records identity, environment, configuration, evaluation, and artifact links, and supports audit and rerun. A replay bundle adds prompts, request/batch/cache/routing traces, random states, outputs, and an environment image, and supports the strongest exact-replay attempt. A reference diagram, not a measured result.}
\end{figure}

\begin{table}[htb]\centering\TableFontEightSix\TableRowAir
\TableTitle{Manifest layers and purpose}
\begin{tabularx}{\linewidth}{@{}Q{1.06in} >{\hsize=0.96\hsize\RaggedRight\arraybackslash}X >{\hsize=1.04\hsize\RaggedRight\arraybackslash}X@{}}
\arrayrulecolor{accent}\toprule
\thd{Layer} & \thd{Anchor fields} & \thd{What it lets you reconstruct}\\
\arrayrulecolor{rule}\midrule
Identity & Model and tokenizer revision, prompt and dataset hashes & What was evaluated, and where the evidence sits\\
\zrow Substrate & Container digest, engine version, driver, GPU topology & The stack the run executed on\\
Execution path & Precision, backend, cache, scheduler, decoding & The numerical and scheduling path\\
\zrow Evaluation & Extractor, scorer, metric, exclusions, uncertainty & How outputs became the reported number\\
\arrayrulecolor{accent}\bottomrule
\end{tabularx}
\numcaptab{Manifest layers and purpose. Anchor fields only; the full list for each layer is in sections 26.1 to 26.4. Documentation is cheap, but restrictive execution choices are not, and the manifest should outlive this edition's particular numbers.}
\TableNote{Synthesis of Chapters 1--25. Evidence: framework.}
\end{table}

\begin{codebox}
\noindent{\sffamily\bfseries\footnotesize\color{accent}Validation rule.}\enspace{\sffamily\footnotesize Fail the run in CI when required manifest fields are missing, mutable revisions such as \code{main} are used, artifact hashes do not resolve, or the scorer version is absent.}
\end{codebox}

\begin{findingbox}A complete run record makes an inference result auditable and rerunnable under the intended condition. It does not automatically make the execution deterministic. Preserve immutable model and tokenizer revisions, the software and hardware environment, the serving and decoding path, the exact request set, the evaluation contract, and hashes for raw artifacts. Use a replay bundle when byte-identical output is required.
\end{findingbox}

\chapter{Evaluation Preflight and Reporting Requirements}

\begin{figure}
\centering
\includegraphics[width=\linewidth]{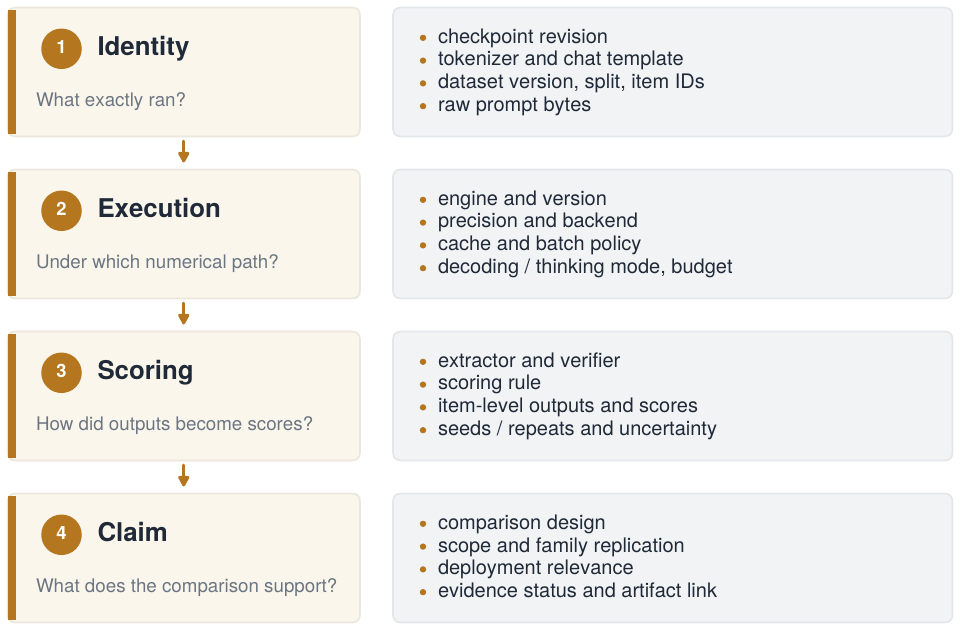}
\caption{\textbf{Four control stages in an evaluation result.} Identity records the checkpoint, prompt, and data; execution records the numerical path and decoding conditions; scoring records how outputs became item-level results; and claim review records comparability, scope, and evidence status. The preflight in Tables 27.1 and 27.2 attaches an artifact to every stage.}
\end{figure}

\medskip{}

An evaluation result is produced by four stages: identity, execution, scoring, and claim review. A failure at any stage can change the reported value without changing the underlying checkpoint. The preflight in this chapter records the controls required at each stage and the artifact that demonstrates completion.

The checklist has two uses. Before a run, it fixes the model, data, prompt, numerical path, and decoding conditions. Before a result is compared or published, it verifies the scorer, uncertainty, workload relevance, and supported scope. The compact version at the end is suitable for routine use; the full version defines the release requirement.

\section{27.1 Four stages of evaluation control}

\emph{Identity} answers what was evaluated: the exact checkpoint revision, tokenizer, chat template, dataset revision, split, item IDs, preprocessing, and rendered prompt. These fields prevent an apparent model change from being caused by a different revision, prompt wrapper, or data slice.

\emph{Execution} answers how outputs were produced: engine and version, CUDA and framework versions, attention backend, precision, quantization, KV dtype, cache state, batch policy, decoding mode, thinking-mode state, generation budget, stop rules, and request order. The preceding chapters show that several of these choices change text at temperature zero and that an insufficient generation budget can invalidate the score.

\emph{Scoring} answers how outputs became item-level judgments: extractor or verifier, scoring rule, parser version, raw outputs, and paired transitions. Aggregate accuracy is not enough when a parser can move the score or when a change causes both correct-to-wrong and wrong-to-correct transitions that cancel in the net delta.

\emph{Claim review} answers what the evidence supports: the comparison design, repeatability, family coverage, workload relevance, remaining confounds, and evidence status. A same-engine checkpoint comparison is stronger than a cross-engine comparison but does not isolate architecture. A high-confidence single-checkpoint result is valid within its scope and should be labeled accordingly.

\section{27.2 Before generation: identity and execution conditions}

The run manifest should be created before model launch rather than reconstructed afterward.

\begin{table}[htb]\centering\TableFontNine\TableRowAir
\TableTitle{Pre-generation controls}
\begin{tabularx}{\linewidth}{@{}Q{0.78in} >{\RaggedRight\arraybackslash}X@{}}
\arrayrulecolor{accent}\toprule
\thd{Stage} & \thd{Records to fix before launch}\\
\arrayrulecolor{rule}\midrule
Identity & Checkpoint revision, tokenizer, chat-template hash, dataset revision and split, item IDs, preprocessing, rendered prompt\\
\zrow Execution & Engine and version, CUDA and framework, attention backend, precision, quantization, KV dtype, cache state, batch policy, scheduler\\
Decoding & Mode, temperature, thinking-mode state, generation budget, stop rules, seed policy, request order\\
\arrayrulecolor{accent}\bottomrule
\end{tabularx}
\numcaptab{Pre-generation controls: the identity, execution, and decoding records fixed before any output is produced.}
\TableNote{Create the manifest before launch; a manifest reconstructed afterwards can miss the condition that moved the number.}
\end{table}

The generation budget requires an explicit validity test. Report the fraction of outputs that reached the configured limit; a capability score should not be interpreted when truncation remains material. Thinking mode, total output limit, reasoning-block behavior, and stop conditions should be recorded separately, because a prompt-level thinking request is not equivalent to a hard output cap. Templates require a comparison-design label: a controlled study may hold prompt bytes fixed across checkpoints, while a deployment study may use each checkpoint's native template. These answer different questions and should not be presented as the same experiment.

\section{27.3 During scoring: extraction, item-level records, and uncertainty}

Raw outputs and item IDs should be retained before any extractor or scorer is applied. The scorer should then produce an item-level record containing the extracted answer, correctness judgment, failure reason, and scorer version. A small stratified sample of correct, incorrect, and unparsed cases should be inspected manually before the aggregate number is accepted.

Sampled decoding requires repeated seeds and a stated summary statistic. Greedy decoding still requires repeatability controls when the numerical path can vary with backend, batch, cache state, or scheduler history; when reseeding has no effect, use disjoint data slices or repeated execution conditions to estimate variation. A paired comparison should report transitions as well as the net change: for correctness, the useful counts are correct-to-correct, correct-to-wrong, wrong-to-correct, and wrong-to-wrong. Changed text is not equivalent to degraded task performance and should be reported separately.

\section{27.4 Before comparison: comparability and scope}

A comparison is interpretable only when the shared and differing conditions are explicit. Hold item IDs, scoring, output budget, and decoding policy constant unless the purpose is to study one of those variables, and record engine, backend, precision, cache, and batching conditions for both arms.

Model-native prompt templates may differ. If native templates are used, report the rendered prompt and describe the result as an operational checkpoint comparison; if identical prompt bytes are used, describe it as a controlled common-prompt comparison. Neither design alone establishes a causal architecture effect. Generalization should follow the replication depth: backend and prefix-cache divergence were tested across three model families; the MoE completion-diversity pattern was extended to two MoE families; and the MMLU harness gap and the Qwen AWQ task split did not have uniform magnitudes across families. The claim wording should preserve those differences.

\section{27.5 Before release: the minimum evidence bundle}

The evaluation must represent the operation used by the deployment. Long-context capability should be reported separately for retrieval, distributed aggregation, and chained state tracking. Tool-use evaluation should separate format validity, semantic call correctness, and abstention. Serving evaluation should state the request-length distribution, arrival process, offered and achieved rate, and absolute p50/p95/p99 metrics for TTFT, inter-token latency, and end-to-end latency as applicable.

\begin{table}[htb]\centering\TableFontNine\TableRowAir
\TableTitle{Scoring, comparison, and release controls}
\begin{tabularx}{\linewidth}{@{}Q{0.86in} >{\RaggedRight\arraybackslash}X@{}}
\arrayrulecolor{accent}\toprule
\thd{Stage} & \thd{Records and artifacts}\\
\arrayrulecolor{rule}\midrule
Scoring & Raw outputs and item IDs; extractor or verifier and its version; item-level judgments; spot-check sample; uncertainty from repeats, slices, or seeds\\
\zrow Comparison & Shared and differing conditions per arm; paired transitions; comparison-design label, common-prompt or native-template\\
Release bundle & Manifest; item IDs and prompt-render hashes; raw outputs; item-level scores and scorer revision; aggregates with uncertainty; engine logs; canonical data; artifact hashes; evidence-status record\\
\arrayrulecolor{accent}\bottomrule
\end{tabularx}
\numcaptab{Scoring, comparison, and release controls. Every aggregate value should trace to an item-level record and a canonical artifact.}
\TableNote{The bundle establishes an auditable result; exact replay requires the additional controls of Chapter 26.}
\end{table}

\section{27.6 The one-page preflight}

\begin{codebox}
{\sffamily\bfseries\footnotesize\color{accent}Before generation}\par\vspace{2pt}
{\sffamily\footnotesize
\begin{itemize}[leftmargin=1.3em,itemsep=1pt,topsep=1pt,label={\framebox[0.62em]{\rule{0pt}{0.5em}}}]
\item Exact checkpoint, tokenizer, template, and prompt rendering recorded.
\item Dataset revision, split, item IDs, and preprocessing recorded.
\item Engine, version, backend, precision, cache, batch, and scheduler recorded.
\item Decoding, thinking mode, output budget, stop rules, and seed policy recorded.
\end{itemize}}
\vspace{3pt}{\sffamily\bfseries\footnotesize\color{accent}Before accepting the score}\par\vspace{2pt}
{\sffamily\footnotesize
\begin{itemize}[leftmargin=1.3em,itemsep=1pt,topsep=1pt,label={\framebox[0.62em]{\rule{0pt}{0.5em}}}]
\item Truncation rate meets the task-specific acceptance threshold.
\item Raw outputs and item-level results are retained.
\item Extractor or verifier is versioned and manually spot-checked.
\item Repeats, slices, or seeds support the stated uncertainty.
\end{itemize}}
\vspace{3pt}{\sffamily\bfseries\footnotesize\color{accent}Before comparison or release}\par\vspace{2pt}
{\sffamily\footnotesize
\begin{itemize}[leftmargin=1.3em,itemsep=1pt,topsep=1pt,label={\framebox[0.62em]{\rule{0pt}{0.5em}}}]
\item Shared and differing conditions are stated for both arms.
\item Paired transitions are reported where aggregate deltas can cancel.
\item The task operation and serving workload match the intended use.
\item Scope, confounds, evidence status, and canonical artifact are recorded.
\end{itemize}}
\end{codebox}

A result is ready for interpretation when the checklist is complete and every aggregate value traces to an item-level record. The checklist does not guarantee identical replay; it establishes an auditable result and identifies the additional controls required when exact replay is the objective.

\nextlesson{\hyperlink{chapter.28}{Chapter 28} defines the capacity and cost calculations that can be performed before launch, and the empirical load test required to establish a sustainable latency operating point.}

\chapter{Capacity, Cost, and Tail-Latency Calculations}

\medskip{}

Several deployment questions can be addressed before a production incident, but they require different methods. KV capacity can be read from the initialized engine and checked against an analytical estimate. Equal-length concurrency can then be calculated as a conservative planning bound. GPU-compute cost can be derived from provisioned cost and measured throughput. A sustainable latency operating point, by contrast, must be established empirically with the intended arrival and length distribution. This chapter separates those four objects and states the assumptions required for each.

\section{28.1 Assumptions and units}

Every calculation should begin with a conditions block: checkpoint and architecture; engine and version; GPU model and count; tensor and pipeline parallel configuration; weight and KV precision; cache block size; the share of device memory reserved for the engine; prompt and maximum output length; throughput denominator; and provisioned cost basis.

The same symbol can otherwise refer to materially different quantities. ``Context length'' may mean prompt length, current sequence length, or the maximum prompt-plus-output reservation. ``Tokens per second'' may mean generated output tokens or total processed tokens. ``GPU cost'' may mean accelerator rental only or a complete provisioned instance.

\section{28.2 Engine-reported KV capacity and a conservative concurrency bound}

When the engine reports the number and size of GPU KV blocks, use that value as the primary capacity input:
\[
C_{\mathrm{KV,engine}} = N_{\mathrm{blocks}}\, B_{\mathrm{block}}.
\]
For a conventional full-attention model, an analytical estimate can be used before initialization:
\[
M_{\mathrm{KV,usable}} = M_{\mathrm{reserved}} - M_{\mathrm{weights}} - M_{\mathrm{runtime}} - M_{\mathrm{margin}},
\]
\[
m_{\mathrm{KV/token/GPU}} \approx 2\, L_{\mathrm{local}}\, H_{kv,\mathrm{local}}\, d_{head}\, b + m_{\mathrm{metadata}},
\]
\[
C_{\mathrm{KV,estimate}} = \left\lfloor M_{\mathrm{KV,usable}} / m_{\mathrm{KV/token/GPU}} \right\rfloor .
\]
The estimate is not architecture-general. Hybrid recurrent layers, multi-head latent attention (MLA), sliding-window attention, cache offload, prefix sharing, and implementation-specific scale metadata require a different memory model. Tensor and pipeline parallelism also change the local number of layers or KV heads. The initialized engine report should therefore replace the estimate whenever available.

A conservative equal-length bound uses prompt plus reserved output tokens:
\[
S_{\mathrm{reserve}} = S_{\mathrm{prompt}} + S_{\mathrm{max\ output}},\qquad
N_{\mathrm{equal}} \le \left\lfloor \frac{N_{\mathrm{blocks}}}{\left\lceil S_{\mathrm{reserve}}/B_{\mathrm{block}} \right\rceil} \right\rfloor .
\]

\begin{notationtable}
$C_{\mathrm{KV,engine}}$ & Engine-reported KV-cache capacity, in tokens\\
$N_{\mathrm{blocks}}$ & Number of GPU KV-cache blocks reported by the engine\\
$B_{\mathrm{block}}$ & Tokens stored in one KV-cache block\\
$M_{\mathrm{reserved}}, M_{\mathrm{weights}}$ & Engine-reserved memory and model-weight residency, in the same memory unit\\
$M_{\mathrm{runtime}}, M_{\mathrm{margin}}$ & Runtime overhead and explicit safety margin, in that memory unit\\
$m_{\mathrm{KV/token/GPU}}$ & Approximate KV-cache memory per active token on one GPU\\
$L_{\mathrm{local}}, H_{kv,\mathrm{local}}$ & Locally stored attention layers and local KV heads\\
$d_{head}, b$ & Attention-head dimension and bytes per cache element\\
$m_{\mathrm{metadata}}$ & Per-token scale or layout metadata included by the cache implementation\\
$C_{\mathrm{KV,estimate}}$ & Analytical KV-token estimate for a conventional full-attention cache\\
$S_{\mathrm{prompt}}, S_{\mathrm{max\ output}}$ & Prompt tokens and reserved maximum output tokens per sequence\\
$S_{\mathrm{reserve}}$ & Sum of prompt and reserved output tokens per sequence\\
$N_{\mathrm{equal}}$ & Conservative equal-length sequence bound\\
\end{notationtable}

For Qwen3-14B on one H100-80GB under vLLM 0.11, with the engine allowed to hold 85\% of device memory, the engine reported 245{,}088 KV token slots. The corresponding equal-length bounds are 119 sequences at 2{,}048 reserved tokens, 29 at 8{,}192, 14 at 16{,}384, and 7 at 32{,}768.

These are planning bounds, not direct observations of the first request above each value failing. Depending on engine policy, capacity pressure may produce queuing, preemption and recomputation, reduced admission, or allocation failure. A representative mixed-length load test remains necessary, and production settings should retain headroom for runtime and workload variation.

\begin{figure}
\centering
\includegraphics[width=\linewidth]{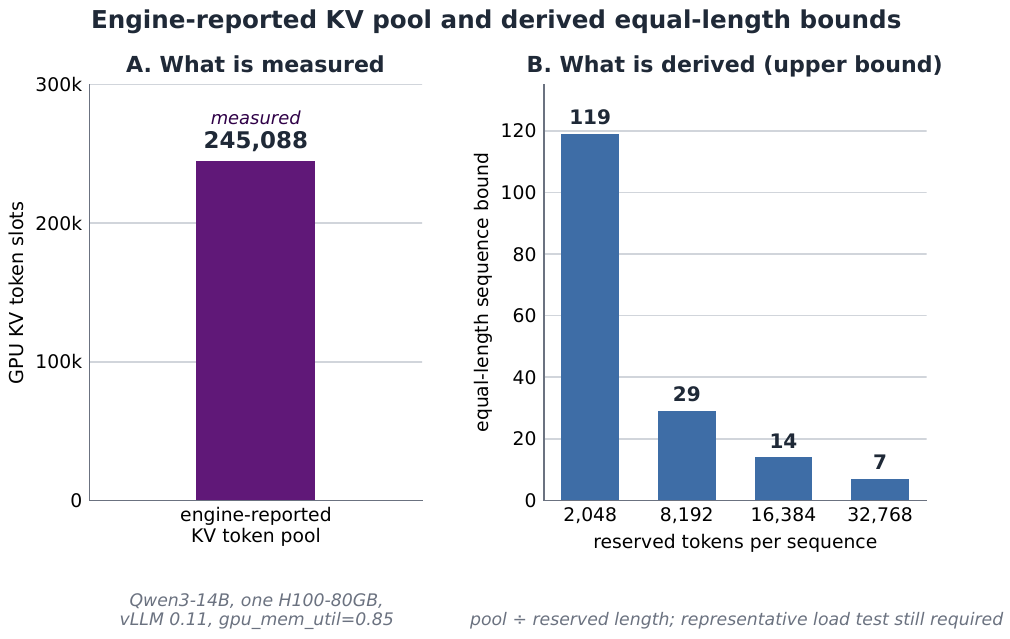}
\caption{\textbf{Engine-reported KV pool and derived equal-length concurrency bounds.} For Qwen3-14B on one H100-80GB under vLLM 0.11, with the engine holding 85\% of device memory, the engine reported 245{,}088 GPU KV token slots. Dividing that pool by reserved lengths of 2{,}048, 8{,}192, 16{,}384, and 32{,}768 tokens gives conservative equal-length bounds of 119, 29, 14, and 7 sequences. The pool is measured; the sequence counts are derived and should be validated under the production length mix and scheduler.}
\end{figure}

\begin{table}[htb]\centering\TableFontEightSix\TableRowAir
\TableTitle{Capacity and cost calculations}
\begin{tabularx}{\linewidth}{@{}>{\hsize=1.14\hsize\RaggedRight\arraybackslash}X >{\hsize=0.86\hsize\RaggedRight\arraybackslash}X Q{0.86in}@{}}
\arrayrulecolor{accent}\toprule
\thd{Calculation} & \thd{Required inputs} & \thd{Status}\\
\arrayrulecolor{rule}\midrule
$C_{\mathrm{KV,engine}} = N_{\mathrm{blocks}} B_{\mathrm{block}}$ & Engine block report & Measured\\
\zrow $C_{\mathrm{KV,estimate}}$, full attention only & Reserved memory, layers, KV heads, head dim, precision & Analytical estimate\\
$N_{\mathrm{equal}} \le \lfloor N_{\mathrm{blocks}} / \lceil S_{\mathrm{reserve}}/B_{\mathrm{block}} \rceil \rfloor$ & Pool, prompt plus reserved output & Derived bound\\
\zrow $\$/\text{M tok} = (r_{\$}/T_{out})\times 10^6$ & GPU count, price, output throughput & Derived estimate\\
$\$/\text{success} = \text{cost} / N_{\mathrm{successful}}$ & Total measured cost, successful items & Derived estimate\\
\arrayrulecolor{accent}\bottomrule
\end{tabularx}
\numcaptab{Capacity and cost calculations: the expression, the inputs it needs, and how far the result can be trusted. The engine-reported KV pool is preferred over an analytical estimate whenever it is available.}
\TableNote{The concurrency and cost rows are planning bounds and estimates, not measured infrastructure results.}
\end{table}

\section{28.3 GPU-compute cost per token}

Let the provisioned accelerator cost rate be $r_{\$} = N_{\mathrm{GPU}}\, P_{\mathrm{GPU\text{-}hour}} / 3600$. For measured output-token throughput $T_{out}$:
\[
\$/\text{M output tokens} = \frac{r_{\$}}{T_{out}} \times 10^6.
\]
The denominator must be named; an output-token estimate should not be compared with a total-token estimate. If the provider prices the full node, use the full allocated-node rate. The result remains a compute estimate unless host resources, storage, networking, orchestration, idle reserve, retries, and autoscaling overhead are included.

In E14, one H100 was priced at an assumed \$4 per hour. Under the tested offline configurations, the derived GPU-compute estimate ranged from \$7.96 per million output tokens at static batch one to \$0.063 under the continuous-batching configuration. The 126.7$\times$ range demonstrates the importance of occupancy under those conditions. It is not a complete bill and should not be transferred to another workload without a new throughput measurement.

\section{28.4 Cost per successful task}

Token cost becomes more useful when connected to task outcome:
\[
\$/\text{success} = \frac{\text{total measured compute cost}}{N_{\mathrm{successful\ items}}}.
\]
\begin{notationtable}
$r_{\$}$ & Provisioned accelerator cost rate, in currency units per second\\
$N_{\mathrm{GPU}}$ & Number of provisioned accelerators\\
$P_{\mathrm{GPU\text{-}hour}}$ & Provisioned price per accelerator-hour\\
$T_{out}$ & Measured generated-output-token throughput, in tokens per second\\
$N_{\mathrm{successful\ items}}$ & Number of items meeting the fixed success criterion\\
\end{notationtable}

For a binary task with comparable attempt cost, this is approximately cost per attempt divided by accuracy. For code, tool use, or open-ended tasks, ``success'' must be defined by a fixed test or acceptance rubric. Cost per success is an optimization metric only after minimum quality, latency, safety, and capacity requirements are met; it should not replace the underlying task score or the declared latency objective.

\section{28.5 Empirical sizing against a latency SLO}

Tail-latency capacity is established through a load sweep rather than a closed-form equation:

\begin{enumerate}[leftmargin=1.6em,itemsep=2pt,topsep=3pt]
\item Record the request-length distribution, arrival process, cache-reuse pattern, and request mix.
\item Define absolute TTFT, inter-token, and end-to-end p95/p99 targets.
\item Sweep offered load or configured concurrency.
\item At each point, record offered and achieved rate, errors, queueing, and absolute p50/p95/p99 metrics.
\item Identify the first point where achieved rate no longer tracks offered rate or any SLO is exceeded.
\item Select the preceding operating point and retain workload-specific headroom.
\end{enumerate}

The H5 real-arrival study used Qwen3-8B, vLLM 0.23, random 256-input/128-output requests, Poisson arrivals, and 200 prompts per rate. Median and p99 TTFT were 67.2 and 79.7 ms at 4 requests per second, 73.8 and 88.2 ms at 16, and 84.0 and 123.8 ms at 64; at the highest condition, only 31.46 requests per second were achieved. The combination of absolute p99 growth and offered-versus-achieved divergence identifies saturation for that workload. This load test and the E14 cost sweep used different workloads and engine versions; they establish two related operational principles but do not form one directly measured cost--tail curve.

\section{28.6 What the calculations do not establish}

The capacity calculation does not establish task quality at long context. The cost calculation does not establish a complete infrastructure bill. The cost-per-success metric does not replace quality or safety constraints. The load sweep does not transfer across model, engine, scheduler, or workload changes without verification. The practical rule is therefore to read the initialized capacity when possible; use analytical estimates for planning, not as a substitute for engine evidence; state the cost denominator and provisioned resources; connect cost to a defined task outcome; and establish the latency operating point with the actual request distribution.

\nextlesson{\hyperlink{chapter.29}{Chapter 29} records how the study's principal claims changed after repetition, cross-family replication, confound controls, and methodology extensions, and defines the evidence metadata used in this edition. Appendix~\ref{app:energy} extends the cost discussion with measured gross GPU energy per output token.}

\chapter{Replication and Evidence Status}

\begin{figure}
\centering
\includegraphics[width=\linewidth]{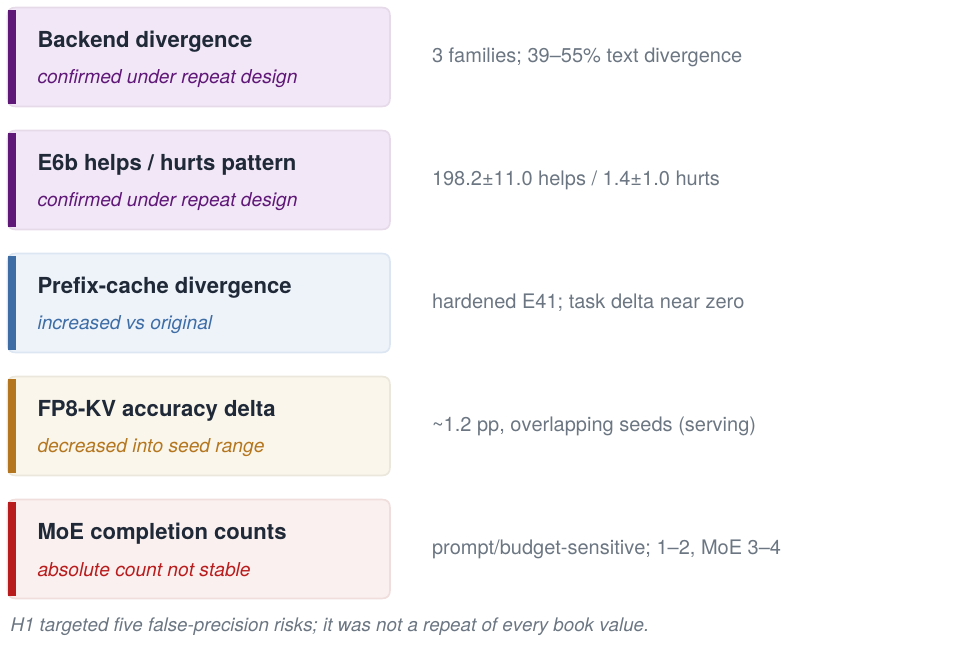}
\caption{\textbf{Repeatability outcomes for the five H1 headline quantities.} Backend divergence and the E6b helps/hurts pattern were confirmed under the repeat design. Prefix-cache divergence increased, the FP8-KV serving-accuracy delta decreased into overlapping seed ranges, and the absolute MoE completion counts were not stable. Conditions and canonical artifacts are listed in Table 29.2.}
\end{figure}

\medskip{}

This edition records how its principal claims changed after repetition, cross-family replication, confound controls, and methodology extensions. The purpose is not to present an unbroken narrative. It is to identify the canonical value, the supported scope, the remaining boundary, and the evidence record from which each claim is generated.

The hardening work addressed five different risks. H1 tested repeatability and false precision. H2 tested cross-family scope. H3 removed a quote-extraction hazard caused by an insufficient generation budget. H4 reran a key architecture comparison under the same engine and locked boundary statements to the result. H5 added real-arrival latency and a harder long-context integration task. The E33--E48 program then repeated under-seeded cells and corrected several parser, thinking-mode, and preliminary-value confounds.

\section{29.1 What the hardening program tested}

A result can fail in different ways. Its magnitude may not repeat. A pattern may be specific to one checkpoint. A standalone number may be misleading outside its setup paragraph. A comparison may contain an engine or harness confound. A method may omit the traffic or task operation required by the claim. The hardening stages were designed around those distinct risks rather than around one generic ``rerun'' label.

\begin{table}[htb]\centering\TableFontNine\TableRowAir
\TableTitle{The five hardening stages}
\begin{tabularx}{\linewidth}{@{}Q{0.42in} >{\hsize=0.92\hsize\RaggedRight\arraybackslash}X >{\hsize=1.08\hsize\RaggedRight\arraybackslash}X@{}}
\arrayrulecolor{accent}\toprule
\thd{Stage} & \thd{Risk tested} & \thd{What it settled}\\
\arrayrulecolor{rule}\midrule
H1 & Repeatability and false precision & Which headline magnitudes were stable under repeats\\
\zrow H2 & Cross-family scope & Which patterns held beyond one model family\\
H3 & Quote safety under truncation & Backend divergence persisted above the truncation knee\\
\zrow H4 & Engine confound closure & The hybrid speed win did not survive a same-engine rerun\\
H5 & Methodology extensions & Real-arrival latency and a harder \mbox{long-context} task\\
\arrayrulecolor{accent}\bottomrule
\end{tabularx}
\numcaptab{The five hardening stages. Each stage tested a different failure mode rather than one generic rerun; the E33--E48 program later repeated under-seeded cells and closed parser and thinking-mode confounds.}
\TableNote{Evidence: H1--H5, E33--E48.}
\end{table}

\section{29.2 Repeatability corrections}

H1 selected five headline quantities with high false-precision risk and repeated them under the original Qwen and vLLM 0.11 stack. Backend text divergence remained approximately 52\%; the important control was same-backend run-to-run divergence of \pmv{0.86}{0.16}\%, which shows that the material change came from the backend switch rather than ordinary rerun noise. Prefix-cache text divergence increased from the preliminary 40\% to \pmv{51.5}{0.9}\%; this strengthens the text-divergence finding but does not establish correctness loss, and the later E41 paired study found accuracy deltas centered near zero with correct-to-wrong transitions of approximately 1.1 to 2.0\% on the tested task.

The E6b sampled short-versus-long generation-budget comparison remained stable: \pmv{198.2}{11.0} items were helped and \pmv{1.4}{1.0} were hurt when moving from 256 to 1024 tokens at temperature 0.7. This should not be merged with the deterministic 1024-versus-4096 plateau comparison, where almost all answers agreed and the larger cap helped none. The short-context FP8-KV serving-accuracy delta decreased from $-1.8$ to approximately $-1.2$ percentage points, with overlapping seed ranges; that conclusion is specific to ordinary task accuracy on the tested short-context workload, and the RL-rollout study used a stricter distribution-preservation contract and rejected FP8-KV under it. The original Qwen dense and MoE completion counts did not repeat in magnitude: the stable conclusion is qualitative and scoped, namely that the tested MoE checkpoints showed at least as much completion diversity across batch contexts as the tested dense reference.

\begin{table}[htb]\centering\TableFontNine\TableRowAir
\TableTitle{H1 repeatability outcomes}
\begin{tabularx}{\linewidth}{@{}>{\RaggedRight\arraybackslash}X C C Q{0.62in}@{}}
\arrayrulecolor{accent}\toprule
\thd{Quantity} & \thd{Preliminary} & \thd{Hardened} & \thd{Status}\\
\arrayrulecolor{rule}\midrule
Backend text divergence & \textasciitilde52\% & \textasciitilde52\% & Confirmed\\
\zrow Prefix-cache divergence & 40\% & \pmv{51.5}{0.9}\% & Increased\\
E6b helps / hurts & 197 / 0 & 198 / 1.4 & Stable\\
\zrow FP8-KV serving delta & $-1.8$ pp & $-1.2$ pp & Reduced\\
MoE completion counts & 2 vs 4 & 1 vs 2 & Not stable\\
\arrayrulecolor{accent}\bottomrule
\end{tabularx}
\numcaptab{H1 repeatability outcomes: five headline quantities, their preliminary and hardened values, and status. The absolute MoE counts did not repeat; the qualitative direction did.}
\TableNote{Same-backend control was \pmv{0.86}{0.16}\%. E6b helps/hurts are \pmv{198.2}{11.0} and \pmv{1.4}{1.0} over the repeat design; the FP8-KV seed ranges overlap; the MoE counts are prompt-dependent. Qwen and vLLM 0.11 stack. Evidence: H1, E41.}
\end{table}

\section{29.3 Replication depth and supported scope}

Backend and prefix-cache text divergence were measured on Qwen3-8B, Llama-3.1-8B, and Mistral-7B; the magnitudes differed, but the pattern was material in all three, so the claims may be stated across the tested families while preserving the model-specific ranges. The MoE completion-diversity result was extended from Qwen3-30B-A3B to gpt-oss-20b, supporting a two-family MoE pattern rather than a universal fixed ratio, and not direct proof that routing assignments changed in the batch experiment; the mechanistic routing evidence came from a separate quantization-forensics study.

Two findings remained model- or toolchain-dependent: the MMLU harness gap was large on Qwen, smaller on Llama, and nearly absent on Mistral, and the Qwen3-14B AWQ task split did not reproduce on the tested Mistral AWQ build. These support per-model validation rather than one correction factor or one quantization rule. License statements should be exact: the Qwen, Mistral, and gpt-oss checkpoints used for the permissively licensed replication claims were Apache-2.0, and Llama was an additional open-weight result under the Llama community license.

\begin{table}[htb]\centering\TableFontEightSix\TableRowAir
\TableTitle{Replication depth and supported scope}
\begin{tabularx}{\linewidth}{@{}Q{1.02in} >{\hsize=0.94\hsize\RaggedRight\arraybackslash}X >{\hsize=1.06\hsize\RaggedRight\arraybackslash}X@{}}
\arrayrulecolor{accent}\toprule
\thd{Finding} & \thd{Replication depth} & \thd{Supported claim}\\
\arrayrulecolor{rule}\midrule
Text divergence & Qwen, Llama, Mistral & Material in all three; state \mbox{per-model} ranges\\
\zrow MoE completion diversity & Qwen3-30B-A3B, gpt-oss-20b & A two-family pattern, not a fixed ratio\\
MMLU harness gap & Large on Qwen, smaller on Llama, near zero on Mistral & Model-dependent; validate per model\\
\zrow Qwen AWQ task split & Not reproduced on Mistral AWQ & Checkpoint- and toolchain-specific\\
\arrayrulecolor{accent}\bottomrule
\end{tabularx}
\numcaptab{Replication depth and supported scope: how far each finding was replicated, and the claim it therefore supports. Cross-family text divergence generalized; the harness gap and the AWQ split did not.}
\TableNote{Text divergence covers both the backend switch and the prefix cache. Licenses: Qwen, Mistral, and gpt-oss are Apache-2.0; Llama is under the Llama community license. Evidence: H2, E8, E-MoE.}
\end{table}

\Needspace{10\baselineskip}
\section{29.4 Confound closure and quote safety}

\begin{readerlist}
\item \textbf{Generation-budget control (H3).} Above the truncation knee, backend text divergence remained 52 to 55\%, while both backend conditions scored above 90\%. The low approximately 53\% accuracies from the earlier 256-token run should not appear as standalone backend results.
\item \textbf{Same-engine control (H4).} Under vLLM 0.23 on both checkpoints, Qwen3-8B full attention produced approximately 2.65$\times$, 3.80$\times$, and 5.16$\times$ the aggregate throughput of Qwen3.5-9B hybrid at 8k, 16k, and 28k in the tested batch-8, prefill-dominated benchmark. The earlier cross-engine implication that the hybrid checkpoint was faster was not supported. The measured hybrid advantages were KV capacity and long-context task performance; these remain checkpoint-level operational comparisons, not a paired architecture ablation.
\end{readerlist}

The editorial rule is to remove superseded figures, retain the audit history in the ledger, and state the confound boundary in the sentence that carries the result.

\Needspace{10\baselineskip}
\section{29.5 Methodology extensions}

\begin{readerlist}
\item \textbf{Real-arrival latency.} The serving-cost sweep used an offline all-at-once workload and could not characterize a stochastic arrival tail. A separate Poisson-arrival experiment showed increasing p99 TTFT and offered-versus-achieved saturation at higher load. The studies support related operational conclusions, but they do not form one jointly measured cost--latency curve.
\item \textbf{A harder context operation.} The original long-context work separated retrieval from chained tracking; H5 added common-words aggregation. On Qwen3-8B, retrieval remained intact through 28k, aggregation declined gently, and six-step state tracking collapsed by 16k. Usable context is therefore operation-specific and requires an explicit acceptance threshold.
\end{readerlist}

\section{29.6 Later corrections from E33--E48}

The E33--E48 program applied a hardening-by-construction rule and later repeated under-seeded cells. Its corrections are clearer as a ledger:

\begin{readerlist}
\item \textbf{Structured output.} The preliminary Qwen result appeared to show a 27-point prompt-request tax, but the freeform condition had default thinking enabled and was truncated before the answer. With thinking disabled and the correct vLLM 0.23 structured-output API, the prompt request was small or beneficial; hard grammar enforcement produced the model-dependent accuracy cost.
\item \textbf{Hybrid-cache risk (E41).} One preliminary slice showed zero correct-to-wrong transitions. Five disjoint slices produced \pmv{1.1}{0.5}\%, so the canonical result does not claim zero risk.
\item \textbf{Under-seeded cells (E36 and E46).} Both were repeated to the five-run or five-seed bar. The E36 Qwen3-32B/512 output-throughput cell was inconsistent between the preliminary sweep and the hardened set, so the released table reports only verified TTFT and ITL for that setting.
\item \textbf{Parser-sensitive claims (E35, E43, and E48).} A scorer mismatch with harmony-format output invalidated the affected gpt-oss cells and the E48 accuracy component. Measurements independent of that parser, including the energy grid, remain with an explicit scope note.
\item \textbf{Cold-start storage label.} Warm runtime initialization was approximately 11 seconds on both tiers. With host-level cache dropping, cold initialization was approximately 201 seconds from the shared network file system and 139 seconds from local block-backed storage. The latter was not true local NVMe.
\end{readerlist}

\section{29.7 Evidence metadata for release}

A single evidence ladder is insufficient because its dimensions are independent. Every primary claim should record:

\begin{readerlist}
\item \textbf{Provenance:} measured in this program or external.
\item \textbf{Validation:} single run, repeated, cross-family replicated, or hardened.
\item \textbf{Scope:} checkpoint, family, task, workload, hardware, and software conditions.
\item \textbf{Status:} canonical, boundary, superseded, or blocked.
\item \textbf{Contract:} task accuracy, text reproducibility, RL distribution preservation, latency SLO, capacity, cost, or energy.
\item \textbf{Canonical artifact:} the raw data object used to generate the final table or figure.
\end{readerlist}

\begin{figure}[H]
\centering
\includegraphics[width=0.84\linewidth]{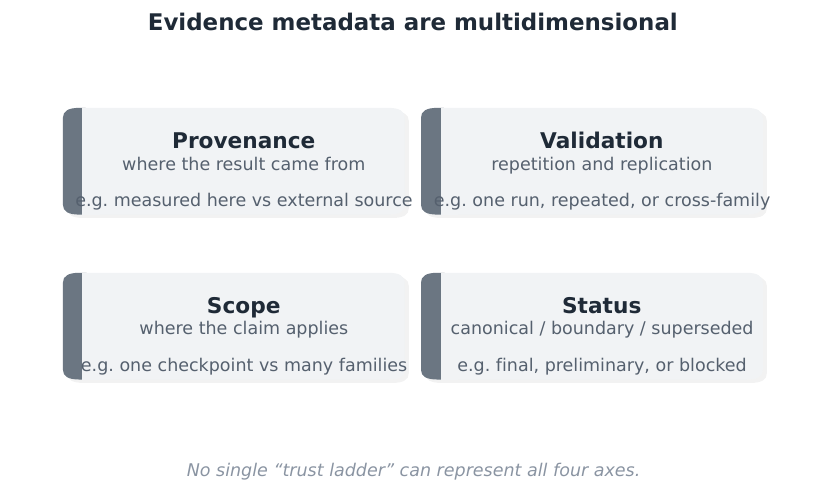}
\caption{\textbf{Evidence metadata has independent dimensions.} Provenance, validation, scope, status, contract, and canonical artifact cannot be collapsed into one ladder. ``Canonical'' identifies the selected artifact; it is not a validation tier.}
\end{figure}

\Needspace{10\baselineskip}
\section{29.8 Remaining boundaries}

\begin{readerlist}
\item \textbf{Precision coverage.} The Hopper-versus-Ampere matrix still requires an A100-class run.
\item \textbf{Offload threshold.} The native prefix-cache sweep did not establish a CPU or NVMe KV-offload threshold.
\item \textbf{Storage tier.} The local cold-start tier was block-backed overlay storage, not instance-attached NVMe.
\item \textbf{Architecture attribution.} A paired ablation is still required.
\item \textbf{Tracking boundary.} The inconsistent E23c Qwen3-8B 8k point is omitted; the released tables report a 2k--16k crossing bracket.
\end{readerlist}

\begin{codebox}
\noindent{\sffamily\bfseries\footnotesize\color{accent}Chapter finding.}\enspace{\sffamily\footnotesize Hardening separates repeatability, cross-family scope, quote safety, engine confounds, and methodology gaps. Record provenance, validation, scope, status, contract, and canonical artifact. Publish canonical values; keep superseded values in the audit ledger and state boundaries with the claim.}
\end{codebox}

\chapter*{Conclusion: Inference Results Are Conditional Measurements}
\addcontentsline{toc}{chapter}{Conclusion: Inference Results Are Conditional Measurements}
\markboth{Conclusion}{Conclusion}

Across six parts, this book examines evaluation, reproducibility, serving, interface controls, architecture comparisons, and operational evidence. Together they support one practical claim: an inference result is a conditional measurement, defined as much by its execution and evaluation path as by the checkpoint that produced it. The sections below restate what each part established.

\section{A score is a pipeline result}

An evaluation number is produced by a chain of choices, not read directly from a model. Generation length, answer extraction, prompt template, scoring path, decoding mode, and task coverage each move the reported value, sometimes by more than the model does. An evaluation claim is interpretable only when those conditions accompany it.

\section{Changed text and changed task outcome are different measurements}

Batch context, attention backend, and prefix caching can each alter greedy-decoded text on a large fraction of prompts. That is a reproducibility fact, not automatically a quality fact. Whether the change affects the task outcome requires a paired evaluation and explicit correct-to-wrong and wrong-to-correct transition counts. Text divergence and task impact answer different questions and must be measured separately.

\section{Throughput and cost belong to a workload}

Model identity alone does not determine throughput, latency, cost, or energy. The arrival process, batch occupancy, prefix reuse, output length, and scheduler policy are part of the result. A serving number is a property of a complete stack under a stated workload, and a derived per-token cost is an estimate over a named denominator, not a complete infrastructure bill.

\section{Capacity is an active constraint}

KV-cache capacity, context reservation, output growth, fragmentation, and admitted concurrency determine what a service can actually hold. Analytical bounds and engine-reported pools are planning aids; they require a representative load test before they can be treated as a production limit.

\section{Interface controls have distinct contracts}

A format request, a grammar constraint, a schema validator, and a prompted thinking target act on different properties. Parseability does not establish semantic correctness; a prompt-level request is not a hard control; and a constraint that guarantees valid structure does not guarantee the right value. Each control should be evaluated against the property it actually governs.

\section{Architecture claims require a matching comparison design}

Cross-engine comparisons establish that two complete systems differ. Same-engine checkpoint comparisons establish an operational difference between the exact tested artifacts. Attributing a difference to an architectural component requires matched checkpoints or a paired ablation. The strength of an architecture claim is bounded by the design of the comparison that produced it.

\section{Reproducibility begins with naming the required equality}

Statistical reproducibility, task-level equivalence, and byte-identical replay are different contracts with different records and acceptance tests. ``Reproducible'' is incomplete until the required level is named. The run record makes a result auditable and rerunnable; exact replay additionally requires a deterministic path and a complete execution trace.

\section{The inference layer is part of the experiment}

The inference layer should not be treated as a transparent channel between a checkpoint and a result. It is part of the experiment and part of the deployed system. The practical response is not to demand one universally deterministic or universally optimal configuration. It is to state the required contract, measure the complete condition, preserve the evidence, and keep the conclusion within the scope that the comparison can support.

\clearpage
\phantomsection
\addcontentsline{toc}{part}{\protect\numberline{}Appendices}
\appendix
\markboth{Appendices}{Appendices}

\chapter{Checkpoint and Runtime Roster}
\label{app:roster}

Reach for this appendix when you need to identify the model, quantized artifact, or runtime behind a result---a family name is not enough. The printed tables are the complete reader-facing roster for this edition and record the most specific checkpoint, artifact, and runtime identities preserved in the retained research notes. Repository revision, tokenizer, chat template, weight format, auxiliary draft head, engine version, and thinking-mode defaults can all change the measured numerical path. Immutable repository revisions, file checksums, raw run records, and executable code are not distributed with this volume; where those fields are absent, the result is auditable at the printed-condition level but not reproducible byte for byte.

\section{A.1 Base checkpoints used in canonical or supporting evidence}

\begin{statictable}\TableFontEightSix\TableRowAir
\TableTitle{Base and target checkpoint registry}
\begin{tabularx}{\linewidth}{@{}Q{0.70in} >{\hsize=0.86\hsize\RaggedRight\arraybackslash}X Q{0.72in} >{\hsize=1.14\hsize\RaggedRight\arraybackslash}X@{}}
\arrayrulecolor{accent}\toprule
\thd{Key} & \thd{Checkpoint} & \thd{Architecture} & \thd{Principal evidence role}\\
\arrayrulecolor{rule}\midrule
M-Q3-4B & Qwen3-4B & dense & Dense comparator for E27\\
\zrow M-Q3-8B & Qwen3-8B & dense & Primary eval, numerical, cache, long context, structured output, serving, energy\\
M-Q3-14B & Qwen3-14B & dense & Quantization, KV capacity, serving, n-gram spec., EAGLE3 target\\
\zrow M-Q3-32B & Qwen3-32B & dense & Scaling, chunked prefill, EAGLE3, quant regime, energy\\
M-Q3-30A3 & Qwen3-30B-A3B & MoE, 30B / 3B active & MoE nondeterminism, memory, quant, reasoning budget, energy\\
\zrow M-Q35-9B & Qwen3.5-9B & hybrid attn. & Hybrid KV capacity, \mbox{long-context} comparison, cache impact\\
M-Q3N-80A3 & Qwen3-Next-80B-A3B-Instruct & hybrid MoE, MTP & RL-rollout speculative-decoding study\\
\zrow M-MIS-7B & Mistral-7B-Instruct-v0.3 & dense & Cross-family replication and head-of-line null\\
M-LL31-8B & Llama-3.1-8B-Instruct & dense & Replication, structured output, BFCL, FP4 pair, energy\\
\zrow M-GPTOSS-20 & gpt-oss-20b & MoE, reasoning & Second-MoE replication; parser-confounded boundary cells\\
M-Q25VL-7B & Qwen2.5-VL-7B-Instruct & multimodal & E45 supplemental visual-token study\\
\zrow M-Q3E-06 & Qwen3-Embedding-0.6B & embedding & E47 embedding throughput and occupancy\\
M-BGEM3 & BGE-M3 & multiling. embed. & E47 second-family embedding study\\
\arrayrulecolor{accent}\bottomrule
\end{tabularx}
\numcaptab{Base and target checkpoint roster. Each key identifies the named repository or recorded mirror used in the retained evidence.}
\TableNote{Licenses: Qwen, Mistral, gpt-oss, and Qwen3-Next are Apache-2.0; Llama is under the Llama 3.1 Community License; embedding and vision-language licenses follow their model cards. Immutable repository revisions and tokenizer and template hashes are not included in this volume.}
\end{statictable}

Each checkpoint family resolves to an official report or model card: the Qwen3 checkpoints to the Qwen3 technical report \autocite{qwen2025qwen3}, Qwen3.5-9B to the Qwen3.5 family source \autocite{qwen2026qwen35}, Llama-3.1-8B to the Llama 3 herd report \autocite{grattafiori2024llama3}, Mistral-7B to the Mistral 7B report \autocite{jiang2023mistral}, gpt-oss-20b to the gpt-oss model card \autocite{openai2025gptoss}, Qwen2.5-VL-7B to the Qwen2.5-VL technical report \autocite{bai2025qwen25vl}, Qwen3-Embedding-0.6B to the Qwen3 Embedding report \autocite{zhang2025qwen3embedding}, and BGE-M3 to the BGE M3-Embedding report \autocite{chen2024bgem3}. A model report identifies the family; an exact rerun would additionally require the immutable revision and, for a quantized derivative, its parent checkpoint, quantizer version, calibration source, group size, excluded layers, tensor format, and artifact digest. Those exact-byte fields are not supplied here.

\section{A.2 Quantized checkpoints and auxiliary artifacts}

Quantized derivatives and draft heads are separate measured artifacts; they must not inherit the base checkpoint's identity without an explicit parent relation.

\begin{statictable}\TableFontEightSix\TableRowAir
\TableTitle{Quantized and auxiliary artifact registry}
\begin{tabularx}{\linewidth}{@{}Q{0.86in} >{\hsize=1.10\hsize\RaggedRight\arraybackslash}X >{\hsize=0.90\hsize\RaggedRight\arraybackslash}X Q{0.62in}@{}}
\arrayrulecolor{accent}\toprule
\thd{Key} & \thd{Artifact $\to$ parent} & \thd{Format or function} & \thd{Used in}\\
\arrayrulecolor{rule}\midrule
Q-Q3-14-AWQ & Qwen3-14B-AWQ $\to$ M-Q3-14B & W4A16, AWQ & E8\\
\zrow Q-Q3-32-AWQ & Qwen3-32B-AWQ $\to$ M-Q3-32B & AWQ/Marlin, FP16 runtime & E9, E48\\
Q-Q3-32-W8A8 & Qwen3-32B-W8A8 $\to$ M-Q3-32B & W8A8 & E9\\
\zrow Q-Q3-30A3-AWQ & Qwen3-30B-A3B-AWQ $\to$ M-Q3-30A3 & MoE AWQ & E29, A3\\
Q-MIS-7-AWQ & Mistral-7B AWQ $\to$ M-MIS-7B & AWQ & H2 quant\\
\zrow Q-LL31-NVFP4 & Llama-3.1-8B NVFP4A16 $\to$ M-LL31-8B & compressed-tensors NVFP4A16 & E43\\
D-Q3-14-E3 & Qwen3-14B EAGLE3 head $\to$ M-Q3-14B & one-layer draft head & E44\\
\zrow D-Q3-32-E3 & Qwen3-32B EAGLE3 head $\to$ M-Q3-32B & one-layer draft head & E44\\
D-Q3N-MTP & MTP weights with M-Q3N-80A3 & native multi-token-prediction head & I-1\\
\arrayrulecolor{accent}\bottomrule
\end{tabularx}
\numcaptab{Quantized and auxiliary artifact registry. Weight format, draft method, and engine compatibility travel with each artifact. The Mistral quantization result did not replicate in the tested toolchain.}
\TableNote{The retained notes identify the named artifacts and compatibility requirements shown here; immutable repository revisions and checksums are not included in this volume.}
\end{statictable}

\section{A.3 Runtime and tokenizer compatibility}

\begin{statictable}\TableFontEightSix\TableRowAir
\TableTitle{Runtime compatibility and recorded overrides}
\begin{tabularx}{\linewidth}{@{}Q{1.06in} Q{0.94in} >{\RaggedRight\arraybackslash}X@{}}
\arrayrulecolor{accent}\toprule
\thd{Artifact class} & \thd{Canonical runtime} & \thd{Load-critical condition or override}\\
\arrayrulecolor{rule}\midrule
Original E1--E32 Qwen serving & vLLM 0.11.0 & Exact backend, budget, tokenizer, thinking state; main-guard required in offline scripts\\
\zrow Hybrid and E33--E48 runs & vLLM 0.23 & API-specific structured-output and hybrid support; do not transplant 0.11 API calls\\
Qwen3-Next MTP & SGLang 0.5.12, CUDA 13 & Four-way tensor parallelism and MTP scheduler flags; separate from the vLLM CUDA 12.8 environment\\
\zrow Llama mirror & vLLM 0.11 or 0.23 per claim & Mirror must match official weights and tokenizer; the official Llama license applies\\
Llama NVFP4 & vLLM 0.23 & Quant format supported; base Llama chat template supplied and recorded\\
\zrow Qwen3-32B AWQ, energy & vLLM 0.23 & FP16 with the AWQ/Marlin path; a BF16 load is invalid for this artifact\\
EAGLE3 target and draft pairs & vLLM 0.23 & Fresh process per mode; sufficient KV headroom; batch-1 preliminary cells kept separate\\
\arrayrulecolor{accent}\bottomrule
\end{tabularx}
\numcaptab{Runtime compatibility and recorded overrides. Engine version, template source, and required dtype are part of the reproduction contract.}
\TableNote{The main-guard note records a runtime requirement observed in the offline vLLM path; no script is distributed with this volume. Artifact licenses are summarized above; ``open,'' ``open source,'' and ``open weight'' are not license identifiers.}
\end{statictable}

\chapter{Dataset and Workload Registry}
\label{app:workloads}

Reach for this appendix when you need the dataset selection, generated workload, or scorer behind a printed number. This appendix is the complete reader-facing roster of the public datasets, generated diagnostic tasks, serving-load definitions, and scoring rules used in printed claims. A dataset name alone is insufficient. The tables therefore state the selection and scoring contract preserved in the retained notes. Generator source, immutable item lists, item hashes, and executable scorers are not distributed, so the designs can be reconstructed from the printed definitions but not replayed byte for byte.

\section{B.1 Public datasets used in printed claims}

\begin{statictable}\TableFontEightSix\TableRowAir
\TableTitle{Public datasets used in printed evidence}
\begin{tabularx}{\linewidth}{@{}Q{0.62in} >{\hsize=1.20\hsize\RaggedRight\arraybackslash}X >{\hsize=0.86\hsize\RaggedRight\arraybackslash}X >{\hsize=0.94\hsize\RaggedRight\arraybackslash}X@{}}
\arrayrulecolor{accent}\toprule
\thd{Key} & \thd{Dataset and selection} & \thd{Scorer} & \thd{Used in}\\
\arrayrulecolor{rule}\midrule
D-GSM8K & GSM8K; repository ID \code{openai/gsm8k}; test split, n = 1{,}319; named 100/200/256/512 slices retain item IDs & Exact scorer key, Table B.4 & E1, E2, E4, E6, E8--E11, E16, E33, E41, E43, E46, E48\\
\zrow D-MMLU & MMLU; exact tasks, split, few-shot policy; 200/500/1{,}000-item selections & Generation or log-likelihood & E3, E5, E8, E29, E43\\
D-HE & HumanEval; repository ID \code{openai\_humaneval}; 164-item test & Unit-test harness and version & E8, E29\\
\zrow D-BFCL-S & BFCL v4; category \mbox{\code{simple\_python}}; 40 per category per seed & Official AST checker version & E35\\
D-BFCL-I & BFCL v4; category \mbox{\code{irrelevance}}; 40 per category per seed & Abstention classification & E35\\
\arrayrulecolor{accent}\bottomrule
\end{tabularx}
\numcaptab{Public datasets used in printed evidence. The roster includes only datasets supporting a printed claim; a release bundle for exact rerun would also need repository revision, retained item IDs, and scorer identity.}
\TableNote{``Repository ID'' is the source package name, not an error or status label. Licenses: GSM8K, MMLU, and HumanEval are MIT; BFCL follows its package license. Exact dataset revisions and retained item lists are not included in this volume.}
\end{statictable}

Each public dataset resolves to its defining source: GSM8K \autocite{cobbe2021gsm8k}, MMLU \autocite{hendrycks2021mmlu}, HumanEval \autocite{chen2021humaneval}, and BFCL \autocite{patil2025bfcl}. The generated diagnostic tasks in the next section are not public benchmarks; they are identified by generator version, seed, item artifact, and checksum rather than by a literature citation, and the RULER task family that inspired the tracing and aggregation diagnostics is cited at its first use in the \mbox{long-context} chapters.

\section{B.2 Generated diagnostic and long-context workloads}

\begin{statictable}\TableFontEightSix\TableRowAir
\TableTitle{Generated diagnostic and long-context workloads}
\begin{tabularx}{\linewidth}{@{}Q{0.90in} >{\hsize=1.10\hsize\RaggedRight\arraybackslash}X >{\hsize=0.90\hsize\RaggedRight\arraybackslash}X Q{0.62in}@{}}
\arrayrulecolor{accent}\toprule
\thd{Key} & \thd{Generator or fixed artifact} & \thd{Metric or matcher} & \thd{Used in}\\
\arrayrulecolor{rule}\midrule
W-FORK & Fixed prompt bytes for the batch-context study & Byte divergence and first fork & E7\\
\zrow W-CACHE-GSM & Long shared-prefix GSM8K construction & Divergence, accuracy delta, C $\to$ W & E16, E41\\
W-NIAH & Custom needle-recall generator & Exact recall & E22, E23\\
\zrow W-TRACK & Six-step chained variable tracking & Strict-match task accuracy & E23c\\
W-AGG & Common-words aggregation generator & At least 2 of the top 3 words & H5\\
\zrow W-PREFIX-PRESSURE & \textasciitilde3k-token shared prefix, 40 queries & Mean ms per query and speedup & E42\\
\arrayrulecolor{accent}\bottomrule
\end{tabularx}
\numcaptab{Generated diagnostic and long-context workloads. A suite label is not a substitute for the printed construction, sample count, and scoring contract.}
\TableNote{Generator source, hashes, and produced item artifacts are not distributed; claims are scoped to the workload definitions printed here.}
\end{statictable}

\section{B.3 Serving, multimodal, and embedding workloads}

\begin{statictable}\TableFontEightSix\TableRowAir
\TableTitle{Serving and supplemental workload registry}
\begin{tabularx}{\linewidth}{@{}Q{0.90in} >{\hsize=1.16\hsize\RaggedRight\arraybackslash}X >{\hsize=0.84\hsize\RaggedRight\arraybackslash}X Q{0.60in}@{}}
\arrayrulecolor{accent}\toprule
\thd{Key} & \thd{Definition} & \thd{Measured outputs} & \thd{Used in}\\
\arrayrulecolor{rule}\midrule
W-H5-POISSON & \code{vllm bench serve}; 256 in, 128 out; 200 requests; Poisson at 4, 16, 64 req/s & Achieved rate, median and p99 TTFT & H5, Ch.\ 13\\
\zrow W-E14-OFFLINE & 512-prompt offline batching; static and continuous grouping & Output tok/s, derived cost, offline latency & E14\\
W-E36-CHUNK & Random 512 in, 128 out; rate 16; token budget 512, 2{,}048, 8{,}192 & TTFT, ITL, throughput & E36\\
\zrow W-VQA-DIGIT & 1024$\times$1024 digit-and-shape images; 30 items; five seeds & Prompt tokens, accuracy, request time & E45\\
W-EMB-SHORT & Synthetic text, about 5 tokens & Embeddings/s, GPU-busy fraction & E47\\
\zrow W-EMB-LONG & Synthetic text, about 512 tokens & Embeddings/s, GPU-busy fraction & E47\\
\arrayrulecolor{accent}\bottomrule
\end{tabularx}
\numcaptab{Serving and supplemental workload roster. Arrival process, request count, and length distribution are part of the workload identity and travel with the claim.}
\TableNote{The H5 tail study used generated Poisson arrivals, not a public trace.}
\end{statictable}

\section{B.4 Scorers and matchers}

\begin{statictable}\TableFontEightSix\TableRowAir
\TableTitle{Scorer registry}
\begin{tabularx}{\linewidth}{@{}Q{0.90in} >{\hsize=0.88\hsize\RaggedRight\arraybackslash}X >{\hsize=1.12\hsize\RaggedRight\arraybackslash}X@{}}
\arrayrulecolor{accent}\toprule
\thd{Key} & \thd{Purpose} & \thd{Known boundary}\\
\arrayrulecolor{rule}\midrule
S-GSM-STRICT & Number after the strict marker & Undercounts alternate formatting\\
\zrow S-GSM-FINAL & Final-number or flexible extraction & Must not select intermediate arithmetic; the E48 version was confounded\\
\zrow S-MMLU-GEN & Generated letter scoring & Output-format sensitive\\
S-MMLU-LL & Log-likelihood choice scoring & First-token and chat-template interaction is model-dependent\\
\zrow S-BFCL-AST & BFCL AST correctness & Does not execute an external tool\\
S-LONG-* & Task-specific long-context matchers & Substring and placement biases must be audited\\
\arrayrulecolor{accent}\bottomrule
\end{tabularx}
\numcaptab{Scorer roster. A printed score can be reproduced exactly only when the scorer implementation is identified alongside the data and output bytes; this volume instead records the operational scoring rule available in the retained notes.}
\TableNote{Proposed-but-unused suites such as MATH-500, AIME, MMLU-Pro, and LongBench are omitted because they do not support a canonical printed result. Scorer code hashes and library versions are not distributed.}
\end{statictable}

\chapter{Reproduction Sequence and Acceptance Criteria}
\label{app:reproduction}

Reach for this appendix after the reproducibility contracts in Chapter~\ref{chapter:repro-contract} or the release checklist in Chapter~27, when you need the order of work and what ``reproduced'' means for a specific claim. The recommended sequence is ordered by dependency and diagnostic value. It begins by validating the runtime and measurement path, then moves from inexpensive evaluation controls to numerical-path sensitivity, serving systems, long-context studies, and specialized inference. A successful reproduction does not always mean reproducing an identical decimal; the target may be an exact value, a statistical effect, a curve, a boundary, or a null.

\section{C.1 Reproduction stages}

\begin{statictable}\TableFontEightSix\TableRowAir
\TableTitle{Reproduction sequence}
\begin{tabularx}{\linewidth}{@{}Q{0.34in} >{\hsize=0.90\hsize\RaggedRight\arraybackslash}X >{\hsize=0.98\hsize\RaggedRight\arraybackslash}X >{\hsize=1.12\hsize\RaggedRight\arraybackslash}X@{}}
\arrayrulecolor{accent}\toprule
\thd{Stage} & \thd{Purpose} & \thd{Representative experiments} & \thd{Completion condition}\\
\arrayrulecolor{rule}\midrule
0 & Validate runtime, artifact path, and evidence capture & Completion probe; mount, process, and scorer checks & Manifest complete; non-empty raw outputs; GPU and mount verified\\
\zrow 1 & Validate evaluation controls & E1, E2, scoped E5 & Truncation, extraction, and prompt-order effects measured on retained items\\
2 & Numerical-path sensitivity with controls & E7, E11/H1/H2/H3, E16/E41 & Switch effect against same-path repeatability; task impact separated from text drift\\
\zrow 3 & Serving economics and capacity & E13, E14/H5, E15, E36 & Throughput, achieved load, p50 and p99, and capacity reported together\\
4 & Context and architecture & E23/H5, E23c, E25/E28/H4 & Retrieval, aggregation, and tracking reported separately; same-engine boundary respected\\
\zrow 5 & Specialized control surfaces & E33--E35, E43--E48, E19/E44 & Format, semantics, energy, modality, and speculative results carry full method and uncertainty\\
\arrayrulecolor{accent}\bottomrule
\end{tabularx}
\numcaptab{Reproduction sequence: dependency-ordered stages from runtime validation to specialized surfaces. A Stage 0 failure is operational, not model behavior.}
\TableNote{Resource needs vary by stage and target hardware; this dependency sequence does not specify a minimum machine allocation.}
\end{statictable}

\section{C.2 What should reproduce}

Exact numerical equality is required only where the same bytes and deterministic path are being replayed. Most deployment findings are curves, scoped comparisons, or statistical effects.

A future reproduction release is complete only when it retains the run record and
image digest; exact checkpoint, tokenizer, and template revisions; data item IDs or
the generated artifact; raw prompts and outputs; scorer version and item-level scores;
logs and GPU and process state; summary metrics with uncertainty; a canonical artifact
key and checksum; and the outcome relative to the printed claim. These files do not
accompany this volume.

\begingroup
\renewcommand{\arraystretch}{0.94}
\begin{longtable}{@{}>{\RaggedRight\arraybackslash}p{0.80in} >{\RaggedRight\arraybackslash}p{1.02in} >{\RaggedRight\arraybackslash}p{2.28in}@{}}
\caption*{\usekomafont{caption}\textbf{Table C.2.} Acceptance criteria for representative reproductions. The acceptance type states what has to match; the third column states what may legitimately differ.}\\[2pt]
\arrayrulecolor{accent}\toprule
\thd{Experiment} & \thd{Acceptance type} & \thd{Replication target, and what may differ}\\
\arrayrulecolor{rule}\midrule\endfirsthead
\arrayrulecolor{accent}\toprule
\thd{Experiment} & \thd{Acceptance type} & \thd{Replication target, and what may differ}\\
\arrayrulecolor{rule}\midrule\endhead
E1 truncation & curve & Accuracy rises as truncation falls to a plateau; the exact knee may differ\\
E2 extractor & exact on frozen bytes & Identical bytes receive scorer-dependent results; the spread differs on new outputs\\
E5 exemplar order & directional, scoped & Order materially affects the tested prompt; the exact 34-point range may differ\\
E7 batch context & directional & A batch context may fork the greedy path; the exact fork token differs\\
E11 backend & statistical, control & A/B divergence materially exceeds the same-backend repeat; the percentage differs\\
E16/E41 cache & statistical & Cache can change text; task impact measured separately; magnitude differs\\
E13 execution path & boundary, full stack & Three stacks reported as distinct; the factor differs by hardware and workload\\
E14/H5 serving load & curve, boundary & Cost and energy improve with occupancy; the tail is measured under arrivals\\
E15 KV capacity & calculated and empirical & Engine pool and derived bound reproduce under a matching configuration\\
E23/H5 context & curve & Task types have different context-length curves; the threshold differs\\
E33 structured output & statistical & Request and grammar effects decomposed per model\\
E34 guided decoding & null, statistical & Overhead below the preregistered 2$\times$ threshold\\
E35 tool calls & statistical & Format validity distinct from AST correctness; abstention reported\\
E43 NVFP4 & paired, scoped & The paired Llama comparison reproduces within uncertainty\\
E44 EAGLE3 & statistical & Paired speedup above baseline through the tested batches\\
E45 visual tokens & scoped curve & Visual-token count rises with the processor setting; the accuracy knee is measured\\
E47 embeddings & statistical, method-limited & Throughput and sampled occupancy reproduce under the same host and runtime\\
E48 energy & physical, statistical & The gross joules-per-output-token grid reproduces within instrumentation uncertainty\\
\arrayrulecolor{accent}\bottomrule
\end{longtable}
\endgroup

\chapter{Claim and Evidence Index}
\label{app:evidence}

Reach for this appendix when you have a chapter claim in hand and need its evidence set, status, and scope. This appendix maps each chapter and supplemental result to a stable claim key and is the complete reader-facing evidence index for this edition. The evidence IDs are provenance labels, not download links. Immutable artifact checksums, raw run conditions, and executable materials are not distributed. No single evidence tag can represent provenance, validation, status, and scope at once.

\section{D.1 Chapter-to-evidence map}

\begingroup
\renewcommand{\arraystretch}{0.94}
\begin{longtable}{@{}>{\RaggedRight\arraybackslash}p{0.60in} >{\RaggedRight\arraybackslash}p{2.44in} >{\RaggedRight\arraybackslash}p{1.16in}@{}}
\caption*{\usekomafont{caption}\textbf{Table D.1.} Chapter-to-evidence map. Each location resolves to a canonical evidence set and a status.}\\[2pt]
\arrayrulecolor{accent}\toprule
\thd{Location} & \thd{Primary claim or function} & \thd{Status}\\
\arrayrulecolor{rule}\midrule\endfirsthead
\arrayrulecolor{accent}\toprule
\thd{Location} & \thd{Primary claim or function} & \thd{Status}\\
\arrayrulecolor{rule}\midrule\endhead
Ch.\ 1 & Generation-budget truncation curve (E1) & canonical measured\\
Ch.\ 2 & Scorer and extractor spread on frozen outputs (E2) & canonical; verifier wording guarded\\
Ch.\ 3 & Harness and model interaction, exemplar order (E3/H2; E5) & harness hardened; order scoped\\
Ch.\ 4 & Checkpoint- and task-dependent quantization (E8; H2; E9) & model and toolchain scoped\\
Ch.\ 5 & Budget plateau and sampled short/long (E6a; H1-E6b) & separate deterministic and sampled contracts\\
Ch.\ 6 & Batch-context sensitivity of greedy decoding (E7/H1) & qualitative canonical; exact count fragile\\
Ch.\ 7 & MoE amplification across two families (E7-MoE; H2) & general pattern, scoped magnitudes\\
Ch.\ 8 & Backend-switch divergence and controls (E11/H1/H2/H3) & hardened, canonical\\
Ch.\ 9 & Prefix-cache divergence and pressure (E16/H1/H2; E42) & hardened; GPU-resident scope\\
Ch.\ 10 & Cache-on/off task impact (E41) & canonical, \mbox{task-specific}\\
Ch.\ 11 & Reproducibility contracts (synthesis) & synthesis; full-stack bitwise contract\\
Ch.\ 12 & Execution-stack throughput (E13) & canonical stack comparison\\
Ch.\ 13 & Batching cost and real-arrival tail (E14; H5) & two experiments; not one frontier\\
Ch.\ 14 & KV pool and derived concurrency (E15; I-6) & measured pool, derived bound\\
Ch.\ 15 & Chunk-size model-scale inversion (E36) & canonical; one raw cell guarded\\
Ch.\ 16 & P/D boundary, head-of-line null, cold and warm start (E37; E39; E40) & topology and version scoped\\
Ch.\ 17 & Format request against grammar constraint (E33) & canonical per model\\
Ch.\ 18 & Guided-decoding overhead (E34) & canonical small overhead\\
Ch.\ 19 & BFCL validity, AST, abstention (E35) & standard-format canonical; gpt-oss excluded\\
Ch.\ 20 & Thinking-budget adherence (E46) & canonical, two Qwen families; parser boundary\\
Ch.\ 21 & Retrieval, aggregation, tracking (E23; H5; E23c) & task-specific curves\\
Ch.\ 22 & Scale and effective-context bracket (E23c) & canonical; \mbox{raw-cell} guard where applicable\\
Ch.\ 23 & Hybrid reasoning, KV, throughput, cache (E25; E28/H4; E41) & operational comparison\\
Ch.\ 24 & MoE routing, memory, quantization, forensics (E7/H2; E27; E29; A3) & mixed contracts; non-replication inline\\
Ch.\ 25 & Requirements for architecture claims (E28/H4) & synthesis of control levels\\
Ch.\ 26 & Manifest and artifact bundle & manifest framework; no new benchmark\\
Ch.\ 27 & Evaluation preflight & synthesis; no new benchmark\\
Ch.\ 28 & Capacity, cost, and empirical SLO (E15/E14/H5) & calculation and procedure\\
Ch.\ 29 & Repeatability, replication, confounds (H1--H5; V4) & evidence audit\\
App.\ E & Energy and paired NVFP4 (E48; E43) & physical energy canonical; FP4 scoped\\
App.\ F & N-gram and EAGLE3 speculative (E19; E44) & separate engine and version studies\\
App.\ G & Visual-token and embedding (E45; E47) & supplemental, scoped\\
App.\ H & Runtime and operational incidents & operational evidence\\
\arrayrulecolor{accent}\bottomrule
\end{longtable}
\endgroup

\chapter{Supplemental Energy and Low-Precision Measurements}
\label{app:energy}

Reach for this appendix when you need the measured energy-per-token figures or the paired FP4 accuracy result that the chapters reference but do not tabulate. This appendix records two supplemental studies with different evidence contracts. E48 measures physical GPU energy over a generation workload; its accuracy-normalized metric is excluded because the answer extractor was confounded. E43 is a paired checkpoint comparison between Llama-3.1-8B BF16 and a pre-quantized NVFP4 artifact. Cost calculations remain in Chapter 28 and cold-start measurements in Chapter 16, so each canonical result has one home.

\section{E.1 Gross GPU energy per generated token}

Gross GPU energy per generated token is
\[
E_{\mathrm{gross,out}} = \frac{\int_{t_0}^{t_1} P_{\mathrm{GPU}}(t)\,dt}{N_{\mathrm{output}}}.
\]
Here, $E_{\mathrm{gross,out}}$ is gross GPU energy per generated output token; $t_0$ and $t_1$ bound the measurement interval; $P_{\mathrm{GPU}}(t)$ is GPU power in watts at time $t$; and $N_{\mathrm{output}}$ is the number of generated output tokens in that interval. The value is workload-dependent: prompt length, output length, batch occupancy, measurement window, and idle-power treatment all affect it. The table reports the exact workload and does not treat joules per output token as a checkpoint constant. Energy was measured through the NVIDIA Management Library (NVML) cumulative-energy interface \autocite{nvidia2026nvml}. Driver version, sampling interval, integration method, timed interval, prefill inclusion, idle-subtraction policy, and device support status belong in a reproduction run record; they are not supplied as code or raw telemetry with this volume.

\begin{staticfigure}
\includegraphics[width=\linewidth]{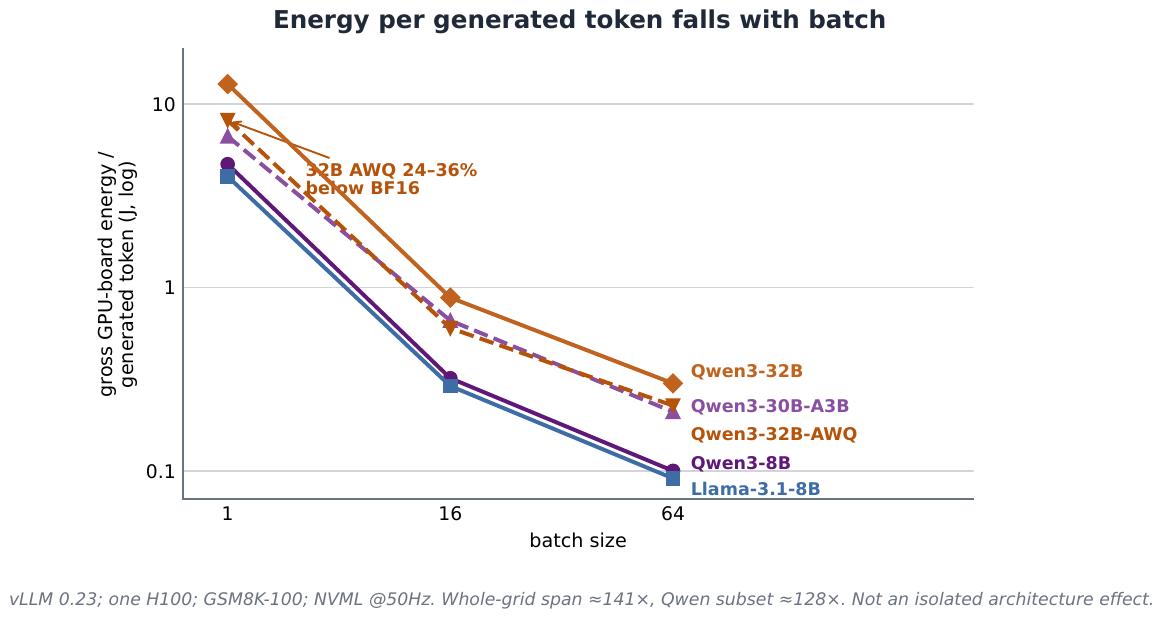}
\StaticCaption{\textbf{Figure E.1. Gross GPU energy per output token by batch occupancy.} Across the tested checkpoints, batch 1 to batch 64 reduced gross joules per output token by roughly 43 to 47 times on one H100. The 32B checkpoint used about 2.7 to 3.0 times the 8B energy per token. Source: E48.}
\end{staticfigure}

\begin{statictable}\TableFontNine\TableRowAir
\TableTitle{Gross GPU energy per generated output token}
\begin{tabularx}{\linewidth}{@{}>{\RaggedRight\arraybackslash}X C C C C@{}}
\arrayrulecolor{accent}\toprule
\thd{Model and config} & \thd{Batch 1} & \thd{Batch 16} & \thd{Batch 64} & \thd{Power at 64}\\[-4pt]
 & \thu{J / out tok} & \thu{J / out tok} & \thu{J / out tok} & \thu{watts}\\
\arrayrulecolor{rule}\midrule
Qwen3-8B BF16 & 4.70 & 0.32 & 0.10 & 259\\
\zrow Llama-3.1-8B BF16 & 4.03 & 0.29 & 0.091 & 288\\
Qwen3-30B-A3B BF16 & 6.67 & 0.66 & 0.21 & 263\\
\zrow Qwen3-32B BF16 & 12.85 & 0.88 & 0.30 & 446\\
Qwen3-32B AWQ & 8.16 & 0.60 & 0.227 & 357\\
\arrayrulecolor{accent}\bottomrule
\end{tabularx}
\numcaptab{Gross GPU energy per generated output token. Batch occupancy produced the largest within-checkpoint change in this one-H100 workload; the printed grid spans about 141 times from Llama at batch 64 to Qwen3-32B at batch 1.}
\TableNote{vLLM 0.23; one H100; GSM8K-100 generation; NVML sampled at 50 Hz with trapezoidal integration; generated-output-token denominator. The timer wraps the logged generation call, and reported energy is gross: measured idle draw of about 69 W is not subtracted. Accuracy and watt-hours per correct answer are excluded because the scorer was confounded. Evidence: E48.}
\end{statictable}

\section{E.2 Paired Llama BF16 and NVFP4 accuracy}

NVFP4 is the low-precision format described by NVIDIA as vendor format documentation \autocite{nvidia2025nvfp4}; the gpt-oss native format is the microscaling MXFP4 format \autocite{rouhani2023mx,ocp2023mx}. The paired accuracy difference below is the E43 result for one Llama checkpoint pair, not a claim made by those format sources.

\begin{statictable}\TableFontNine\TableRowAir
\TableTitle{Paired Llama BF16 and NVFP4 accuracy}
\begin{tabularx}{\linewidth}{@{}>{\RaggedRight\arraybackslash}X C C C C@{}}
\arrayrulecolor{accent}\toprule
\thd{Checkpoint} & \thd{GSM8K} & \thd{MMLU} & \thd{GSM8K $\Delta$} & \thd{MMLU $\Delta$}\\[-4pt]
 & \thu{accuracy} & \thu{accuracy} & \thu{pp} & \thu{pp}\\
\arrayrulecolor{rule}\midrule
Llama-3.1-8B BF16 & \pmv{0.858}{0.017} & \pmv{0.577}{0.014} & baseline & baseline\\
\zrow Llama-3.1-8B NVFP4 & \pmv{0.820}{0.015} & \pmv{0.564}{0.026} & \textminus3.8 & \textminus1.3\\
\arrayrulecolor{accent}\bottomrule
\end{tabularx}
\numcaptab{Paired Llama BF16 and NVFP4 accuracy on GSM8K and MMLU. The observed loss was larger on GSM8K than on MMLU in this checkpoint pair, and the result is checkpoint- and toolchain-specific.}
\TableNote{200 items per task; five seeds; vLLM 0.23; identical prompts, decoding, and scorers; base Llama chat template supplied for the quantized artifact. The gpt-oss native-MXFP4 MMLU cell is excluded because of a harmony-format parse failure. Evidence: E43.}
\end{statictable}

This comparison supports a narrow rule: a small loss on one knowledge benchmark is not sufficient to approve an FP4 artifact for math or code. It does not establish a universal FP4 math-versus-knowledge law.

\chapter{Supplemental Speculative-Decoding Measurements}
\label{app:speculative}

Reach for this appendix after the serving analysis in Chapters 12--16 when you are deciding whether to enable speculative decoding and need the measured, workload-specific speedups. Speculative decoding and speculative sampling preserve the target distribution through verification and correction under their stated assumptions \autocite{leviathan2023specdec,chen2023specsampling}. EAGLE and EAGLE-3 use learned feature-level draft models \autocite{li2024eagle,li2025eagle3}. Prompt Lookup is a repository implementation of repeated-span drafting, while LLMA is a related reference-based lossless acceleration method \autocite{saxena2023promptlookup,yang2023llma}. Multi-token prediction is a separate training objective and acceleration path \autocite{gloeckle2024mtp}. E19 and E44 are separate studies with different engines and cannot be used as a same-engine method ranking.

Speculative-decoding speedup depends on the target checkpoint, draft method, engine version, workload, batch, temperature, context length, and output length. This appendix reports two separate studies; they are not a same-engine method ablation. For a paired enabled and baseline measurement,
\[
S_{\mathrm{spec}}(b,T,d,w,e) = \frac{r_{\mathrm{enabled}}(b,T,d,w,e)}{r_{\mathrm{baseline}}(b,T,w,e)},
\]
where $S_{\mathrm{spec}}$ is the enabled-to-baseline throughput ratio; $b$ is batch size; $T$ is sampling temperature; $d$ is the draft method and compatible artifact; $w$ is the workload definition; $e$ is the engine and version; and $r_{\mathrm{enabled}}$ and $r_{\mathrm{baseline}}$ are the paired end-to-end generated-output rates. Acceptance rate and accepted draft length are separate diagnostics.

\section{F.1 N-gram prompt lookup in vLLM 0.11}

\begin{statictable}\TableFontNine\TableRowAir
\TableTitle{N-gram prompt-lookup throughput in vLLM 0.11}
\begin{tabularx}{\linewidth}{@{}C C C C >{\RaggedRight\arraybackslash}X@{}}
\arrayrulecolor{accent}\toprule
\thd{Batch} & \thd{Enabled} & \thd{Baseline} & \thd{Speedup} & \thd{Interpretation}\\[-4pt]
 & \thu{tok/s} & \thu{tok/s} & \thu{ratio} & \\
\arrayrulecolor{rule}\midrule
1  & 38.5    & 85.7    & 0.45 & Slower than baseline\\
\zrow 4  & 408.2   & 314.2   & 1.30 & Largest measured gain\\
16 & 1{,}408.1 & 1{,}247.7 & 1.13 & Modest gain\\
\zrow 64 & 4{,}103.2 & 4{,}053.3 & 1.01 & Approximately break-even\\
\arrayrulecolor{accent}\bottomrule
\end{tabularx}
\numcaptab{N-gram prompt-lookup throughput in vLLM 0.11. The measured benefit was confined to an intermediate batch range in this GSM8K workload, and the batch-1 regression is specific to short, weakly repetitive prompts.}
\TableNote{Qwen3-14B; vLLM 0.11.0; four speculative tokens; prompt-lookup maximum 3; GSM8K; greedy decoding; generation cap 128 tokens. This is a single measurement sweep; repeat counts and error bars were not logged, so the 1.01 cell at batch 64 is reported as approximately break-even rather than a definitive gain or null. The throughput shape reproduced across engine versions 0.11 and 0.23. Evidence: E19.}
\end{statictable}

\section{F.2 EAGLE3 in vLLM 0.23}

\begin{staticfigure}
\includegraphics[width=\linewidth,trim=0 14pt 0 0,clip]{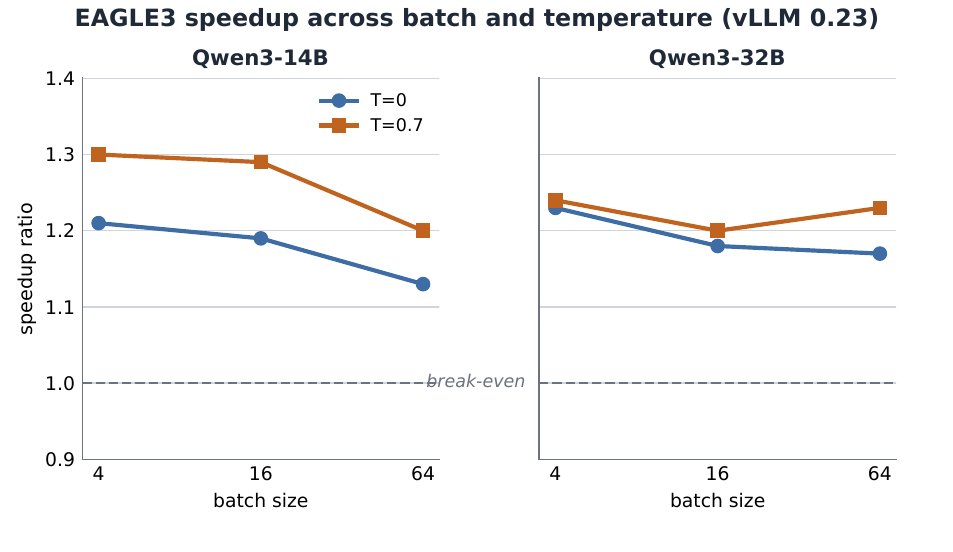}
\StaticCaption{\textbf{Figure F.1. Hardened EAGLE3 enabled-to-baseline throughput ratios.} The five-seed mean remained above 1.0 through batch 64 on both tested targets, at both temperatures. Source: E44 hardened.}
\end{staticfigure}

\begin{statictable}\TableFontNine\TableRowAir
\TableTitle{Hardened EAGLE3 enabled-to-baseline throughput ratios}
\begin{tabularx}{\linewidth}{@{}C C C C C@{}}
\arrayrulecolor{accent}\toprule
\thd{Batch} & \thd{14B, T = 0} & \thd{14B, T = 0.7} & \thd{32B, T = 0} & \thd{32B, T = 0.7}\\[-4pt]
 & \thu{ratio} & \thu{ratio} & \thu{ratio} & \thu{ratio}\\
\arrayrulecolor{rule}\midrule
4  & 1.21 & 1.30 & 1.23 & 1.24\\
\zrow 16 & 1.19 & 1.29 & 1.18 & 1.20\\
64 & 1.13 & 1.20 & 1.17 & 1.23\\
\arrayrulecolor{accent}\bottomrule
\end{tabularx}
\numcaptab{Hardened EAGLE3 enabled-to-baseline throughput ratios in vLLM 0.23. All five-seed means remained above 1.0 through batch 64.}
\TableNote{Qwen3-14B and Qwen3-32B targets with matched EAGLE3 draft heads; vLLM 0.23; three speculative tokens; five seeds; paired throughput. The retained summary provides five-seed point means but not per-cell standard deviations; preliminary batch-1 cells are excluded. Remeasure a deployment cell when the margin over 1.0 is decision-critical. Evidence: E44 hardened.}
\end{statictable}

The studies use different engines and do not rank draft methods. Enable speculative
decoding only when a paired deployment test clears 1.0 with uncertainty, passes the
correctness contract, uses a compatible draft and tokenizer, fits the KV budget, and
gives enabled and baseline runs the same measurement window.

\chapter{Supplemental Multimodal and Embedding Measurements}
\label{app:modalities}

Reach for this appendix as a modality-specific extension of the workload and occupancy arguments in Chapters 12 and 13 when you are serving a vision-language or embedding model. This appendix records two scoped studies outside ordinary text generation. The first measures image-token scaling on a simple synthetic visual task; the second measures embedding throughput and sampled GPU occupancy in vLLM pooling mode. Neither is a complete quality or runtime comparison for its modality.

\section{G.1 Visual-token scaling on a high-contrast synthetic task}

E45 used Qwen2.5-VL-7B-Instruct \autocite{bai2025qwen25vl} on 30 generated 1024$\times$1024 images containing a large digit and a colored shape, sweeping the image-resolution setting across four levels over five seeds. Because the features were large and high-contrast, this is a lower-bound resolution case rather than a fine-detail benchmark.

\begin{statictable}\TableFontNine\TableRowAir
\TableTitle{Visual-token scaling on a high-contrast synthetic VQA task}
\begin{tabularx}{\linewidth}{@{}C C C C@{}}
\arrayrulecolor{accent}\toprule
\thd{Resolution setting} & \thd{Prompt tokens} & \thd{Accuracy} & \thd{Mean request time}\\[-4pt]
\thu{visual-token cap} & \thu{per image} & \thu{five seeds} & \thu{seconds}\\
\arrayrulecolor{rule}\midrule
256  & 294      & \pmv{1.000}{0.000} & 0.062\\
\zrow 512  & 522      & \pmv{1.000}{0.000} & 0.063\\
1024 & 1{,}062  & \pmv{1.000}{0.000} & 0.068\\
\zrow 2048 & 1{,}407  & \pmv{1.000}{0.000} & 0.069\\
\arrayrulecolor{accent}\bottomrule
\end{tabularx}
\numcaptab{Visual-token scaling on a high-contrast synthetic VQA task. Prompt-token count rose 4.8 times while five-seed accuracy stayed at 1.000. The result does not establish the setting for fine text, documents, charts, or interfaces.}
\TableNote{The resolution setting is the maximum-pixels knob passed through the multimodal processor arguments; the four levels are visual-token caps. Qwen2.5-VL-7B-Instruct; vLLM 0.23 multimodal; 30 items; five seeds; generated digit-and-shape images. Evidence: E45.}
\end{statictable}

The supported rule is to locate a content-specific resolution knee on representative inputs; the lowest setting that preserves required detail is preferable to an unconditional maximum-resolution default.

\section{G.2 Embedding throughput and occupancy in vLLM pooling mode}

E47 measured embedding throughput and sampled GPU occupancy for Qwen3-Embedding-0.6B \autocite{zhang2025qwen3embedding} and BGE-M3 \autocite{chen2024bgem3} in vLLM pooling mode. These sources identify the model families; the throughput and occupancy values below are the E47 result.

\begin{statictable}\TableFontNine\TableRowAir
\TableTitle{Embedding throughput}
\begin{tabularx}{\linewidth}{@{}>{\RaggedRight\arraybackslash}X Q{0.50in} C C C C@{}}
\arrayrulecolor{accent}\toprule
\thd{Model} & \thd{Input} & \thd{B1} & \thd{B32} & \thd{B256} & \thd{B512}\\[-4pt]
 & & \thu{emb/s} & \thu{emb/s} & \thu{emb/s} & \thu{emb/s}\\
\arrayrulecolor{rule}\midrule
Qwen3-Embedding-0.6B & short & 50 & 673 & 2{,}804 & 3{,}346\\
\zrow Qwen3-Embedding-0.6B & long & 48 & 474 & 922 & 988\\
BGE-M3 & short & 72 & 1{,}018 & 3{,}655 & 4{,}158\\
\zrow BGE-M3 & long & 68 & 437 & 561 & 564\\
\arrayrulecolor{accent}\bottomrule
\end{tabularx}
\numcaptab{Embedding throughput in vLLM 0.23 pooling mode, in embeddings per second. Throughput rose with batch for short inputs; long BGE-M3 inputs plateaued.}
\TableNote{Short inputs are about 5 tokens, long inputs about 512. The retained summary provides five-seed point estimates but not the per-cell ranges; close capacity decisions require remeasurement. Evidence: E47.}
\end{statictable}

\begin{staticfigure}
\includegraphics[width=\linewidth]{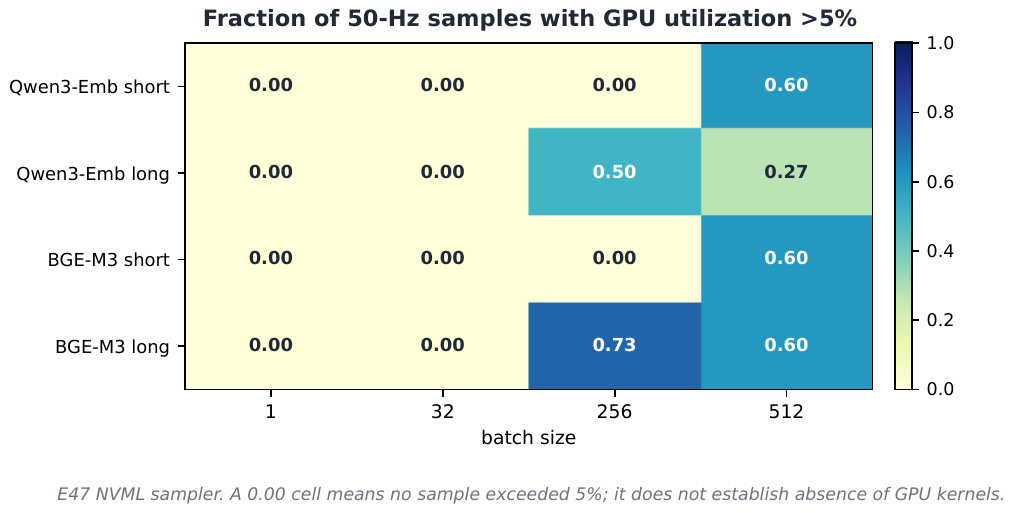}
\StaticCaption{\textbf{Figure G.1. Observed GPU-busy fraction during the embedding runs.} Few 50 Hz samples exceeded 5\% before high batch, suggesting host or scheduler overhead. A 0.00 cell does not mean no kernel ran; confirming a CPU bottleneck requires CPU profiling or a pretokenized control. Embedding quality and a dedicated-runtime comparison were not measured. Source: E47.}
\end{staticfigure}

\chapter{Operational and Runtime Failure Modes}
\label{app:operations}

Use this appendix alongside the serving boundaries in Chapter 16 and the preflight in Chapter 27 when a run failed, stalled, or produced an implausible result. A failed or anomalous run is not evidence about model capability until the artifact, runtime, readiness, process state, storage path, and scoring output have been validated. This appendix records operational failures that occurred during the program and the checks that distinguished them from model behavior.

\section{H.1 Failed-run classification}

\begin{staticfigure}
\includegraphics[width=\linewidth]{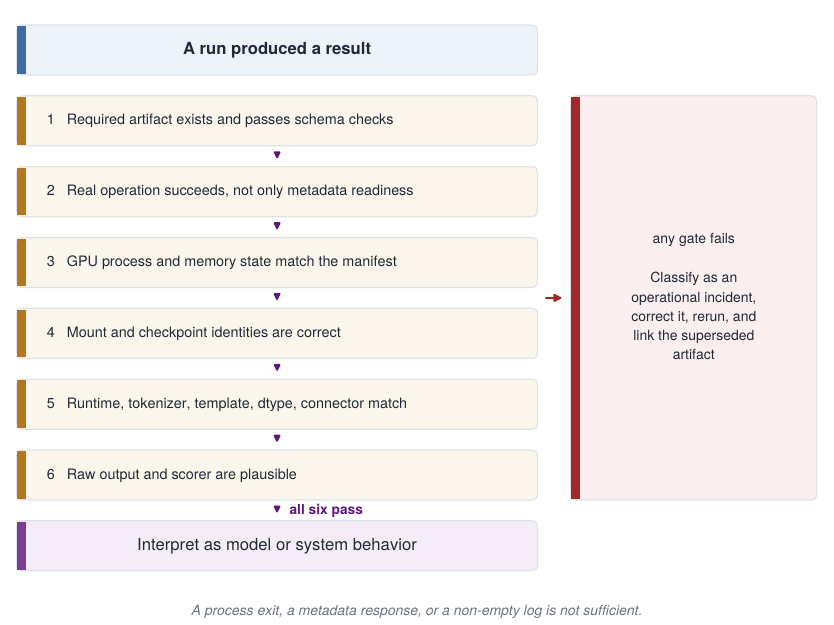}
\StaticCaption{\textbf{Figure H.1. Failed-run classification order.} Confirm a non-empty metrics artifact, require a real generation result, verify GPU allocation and release, verify mounts and checkpoint visibility, verify engine, CUDA, and tokenizer versions, and inspect raw outputs before interpreting a result as model or system behavior. A diagnostic workflow, not a measured result.}
\end{staticfigure}

The order is: confirm a non-empty metrics artifact matching the expected fields; require a real generation or embedding result, since a metadata endpoint is not a readiness test; verify intended GPU allocation before launch and complete memory release after; verify mounted filesystem and checkpoint visibility from every container; verify engine, CUDA, PyTorch, tokenizer, template, quantization, and API versions against the run record; inspect raw outputs and scorer behavior for plausibility; and only then interpret the result as model or system behavior.

\section{H.2 Environment and artifact failures}

\begin{statictable}\TableFontEightSix\TableRowAir
\TableTitle{Environment, mount, and artifact-identity failures}
\begin{tabularx}{\linewidth}{@{}Q{1.10in} >{\hsize=1.00\hsize\RaggedRight\arraybackslash}X >{\hsize=1.00\hsize\RaggedRight\arraybackslash}X@{}}
\arrayrulecolor{accent}\toprule
\thd{Failure mode} & \thd{Diagnostic evidence} & \thd{Corrective action}\\
\arrayrulecolor{rule}\midrule
Two engines in one environment & The resolver pulled CUDA 13 into a CUDA 12.8 environment & Separate pinned containers; preserve the environment lock\\
\zrow Shared-storage path resolved to the overlay & The mount table showed the container overlay, not the network mount & Compare mount tables, filesystem type, and a known checksum before a large download\\
Bare dataset name & The namespaced repository ID loaded successfully & Pin the canonical dataset repository ID and revision\\
\zrow Gated checkpoint unavailable & The official repository was inaccessible; a mirror was downloaded & Record mirror, base identity, revision, byte relation, and license\\
Derived checkpoint missing template & Chat-template application failed, or output was implausible & Use a compatible base template and record the override\\
\zrow Quantized checkpoint needs a specific dtype & The AWQ path rejected BF16 and required FP16 with Marlin & Record dtype and quantization backend in the roster\\
\arrayrulecolor{accent}\bottomrule
\end{tabularx}
\numcaptab{Environment, mount, and artifact-identity failures. These incidents prevent a valid model comparison until the runtime identity is corrected.}
\TableNote{Scope: SGLang and vLLM setup, Qwen3-Next and EAGLE, the nightly dataset stack, Llama replication, NVFP4, and the AWQ energy runs.}
\end{statictable}

\section{H.3 Process and benchmark-validation failures}

\begin{statictable}\TableFontEightSix\TableRowAir
\TableTitle{Process lifecycle, readiness, and benchmark-validation failures}
\begin{tabularx}{\linewidth}{@{}Q{1.10in} >{\hsize=1.00\hsize\RaggedRight\arraybackslash}X >{\hsize=1.00\hsize\RaggedRight\arraybackslash}X@{}}
\arrayrulecolor{accent}\toprule
\thd{Failure mode} & \thd{Diagnostic evidence} & \thd{Corrective action}\\
\arrayrulecolor{rule}\midrule
Missing main-process guard & The engine was constructed during module re-import & Wrap engine creation in the Python main-process guard\\
\zrow Orphan engine-core process & A child process ID remained visible to the GPU management tool & Kill the child process; verify that memory was released\\
Metadata-only readiness probe & The models endpoint responded but generation failed & Readiness requires a real completion with the expected schema\\
\zrow Exit code zero but no metrics & A wrong command-line path exited cleanly & Assert required metric keys and a plausible duration\\
Enabled and baseline engines in one process & Deleting the object does not terminate the worker & One fresh process per mode and model; explicit cleanup\\
\arrayrulecolor{accent}\bottomrule
\end{tabularx}
\numcaptab{Process lifecycle, readiness, and benchmark-validation failures. A successful exit code or a metadata endpoint is insufficient; validate the operation the experiment actually requires.}
\TableNote{Scope: vLLM 0.11 offline scripts, repeated runs, H5 serving, and E44.}
\end{statictable}

\section{H.4 Connector, topology, and storage-pressure failures}

\begin{statictable}\TableFontEightSix\TableRowAir
\TableTitle{Connector, topology, and cold-storage failure modes}
\begin{tabularx}{\linewidth}{@{}Q{1.10in} >{\hsize=1.00\hsize\RaggedRight\arraybackslash}X >{\hsize=1.00\hsize\RaggedRight\arraybackslash}X@{}}
\arrayrulecolor{accent}\toprule
\thd{Failure mode} & \thd{Diagnostic evidence} & \thd{Corrective action}\\
\arrayrulecolor{rule}\midrule
Connector ABI mismatch & Moved utilities and an incompatible compiled extension & Use a compatible pinned image or a supported native connector\\
\zrow Transfer port collision & Producer and consumer shared a default port & Assign distinct role ports; validate registration\\
Single-node P/D host spill & Thousands of host-spill events; the benchmark did not complete & Keep serving co-located, or use a proven fast KV fabric\\
\zrow Concurrent cold reads from the shared filesystem & Serialized reads were stable; concurrent reads thrashed & Serialize cold first touch, pre-stage, or pre-warm\\
Warm cache mistaken for storage performance & Shared and local paths both read at memory speed & Separate cold first touch from warm restart; record cache state\\
\arrayrulecolor{accent}\bottomrule
\end{tabularx}
\numcaptab{Connector, topology, and cold-storage failure modes. These are infrastructure boundaries; they do not establish model-quality conclusions.}
\TableNote{The connector rows refer to the LMCache and NIXL path and the peer-to-peer NCCL transport. Scope: the E37 and E42 environment, E37 setup and boundary, and the E40 cold-start work.}
\end{statictable}

For every failed run, retain the incident ID, timestamp, launch specification, run-record key, container digest, standard output and error, exit code, required-metric validation, GPU process and memory state, mount state, checkpoint identity, root cause, corrective action, rerun artifact, and whether any model-level claim remains valid. These are recommended incident fields; no command file or incident bundle accompanies this volume.

\input{WhatBrokeWhy_VolumeII_RL_Style_References_GPTCodex.tex}

\end{document}

%% file: why_this_book_the_gap.tex
%
\addchap{Why This Book: The Gap}
\markboth{Why This Book: The Gap}{Why This Book: The Gap}

The field has settled conventions for datasets, prompt templates, decoding
parameters, and seeds. It has none for the software that executes the model.

A 2026 survey catalogued 200 inference engines and screened roughly 35,000
machine-learning papers, isolating 9{,}018 that ran language models locally.
Among those releasing no code, about 36 percent named the engine anywhere in
the text, and a third of the released repositories specified no dependencies at
all \autocite{pape2026backends}.

The unreported layer is not small. Holding weights, prompts, decoding,
precision, and hardware fixed and changing only the engine, that same study
found GSM8K accuracy on one distilled reasoning checkpoint spanning 16.60
percentage points across five engines, and jailbreak attack success moving 8.9
points on identical weights. The causes split between correctable engine
defaults and numerical drift from optimizations that high-throughput serving
requires. This book reports the same phenomenon one level lower: inside a
single engine, changing only the attention backend altered generated text on 39
to 55 percent of prompts.

Each surrounding literature is rigorous inside its own layer. Evaluation
frameworks document setup sensitivity \autocite{biderman2024lmeval}; systems
research improves kernels, memory, and scheduling; reproducibility guidance
lists the fields a study should disclose \autocite{pineau2021repro}. None
answers the question a practitioner is holding: given this checkpoint, harness,
engine, and workload, is the score a capability result or a property of the
path that produced it?

That joint condition is what this book measures. Each chapter changes one
element of the inference path, holds the rest fixed, and reports what moved.

\clearpage

%% file: WhatBrokeWhy_VolumeII_RL_Style_References_GPTCodex.tex
\backmatter
\chapter*{References}
\addcontentsline{toc}{chapter}{References}

\begingroup
\sloppy
The methods, systems, models, benchmarks, and reproducibility guidance used by this volume are cited at their first technical use in the main text. This section collects those external sources by the role they play in the book. It is a reading list and provenance record, not evidence for this volume's measured E-, H-, or C-series results. Those results remain tied to the evidence identifiers, conditions, and scope printed in Appendix~\ref{app:evidence}; no raw run registry or executable artifact bundle is distributed with this volume.

\section*{Inference systems and kernels}
\begin{itemize}[leftmargin=1.3em,itemsep=0.45em,topsep=0.3em]
\item \fullcite{kwon2023paged}.
\item \fullcite{yu2022orca}.
\item \fullcite{dao2022flash}.
\item \fullcite{dao2023flash2}.
\item \fullcite{ye2025flashinfer}.
\item \fullcite{zheng2024sglang}.
\item \fullcite{vllm2025apc011}.
\item \fullcite{vllm2025chunked011}.
\item \fullcite{vllm2025engineargs}.
\item \fullcite{cheng2025lmcache}.
\item \fullcite{nvidia2025nixl}.
\item \fullcite{dong2025xgrammar}.
\item \fullcite{nvidia2026nvml}.
\end{itemize}

\section*{Evaluation and datasets}
\begin{itemize}[leftmargin=1.3em,itemsep=0.45em,topsep=0.3em]
\item \fullcite{cobbe2021gsm8k}.
\item \fullcite{hendrycks2021mmlu}.
\item \fullcite{gema2025mmluredux}.
\item \fullcite{wang2024mmlupro}.
\item \fullcite{chen2021humaneval}.
\item \fullcite{liu2023evalplus}.
\item \fullcite{patil2025bfcl}.
\item \fullcite{hsieh2024ruler}.
\item \fullcite{bai2024longbench}.
\end{itemize}

\section*{Quantization and numerical formats}
\begin{itemize}[leftmargin=1.3em,itemsep=0.45em,topsep=0.3em]
\item \fullcite{lin2024awq}.
\item \fullcite{frantar2023optq}.
\item \fullcite{xiao2023smoothquant}.
\item \fullcite{micikevicius2022fp8}.
\item \fullcite{rouhani2023mx}.
\item \fullcite{ocp2023mx}.
\item \fullcite{nvidia2025nvfp4}.
\item \fullcite{compressedtensors2026}.
\end{itemize}

\section*{Speculative decoding}
\begin{itemize}[leftmargin=1.3em,itemsep=0.45em,topsep=0.3em]
\item \fullcite{leviathan2023specdec}.
\item \fullcite{chen2023specsampling}.
\item \fullcite{li2024eagle}.
\item \fullcite{li2025eagle3}.
\item \fullcite{saxena2023promptlookup}.
\item \fullcite{yang2023llma}.
\item \fullcite{gloeckle2024mtp}.
\end{itemize}

\section*{Architectures and model reports}
\begin{itemize}[leftmargin=1.3em,itemsep=0.45em,topsep=0.3em]
\item \fullcite{vaswani2017attention}.
\item \fullcite{shazeer2017moe}.
\item \fullcite{fedus2022switch}.
\item \fullcite{qwen2025qwen3}.
\item \fullcite{qwen2026qwen35}.
\item \fullcite{yang2025gateddelta}.
\item \fullcite{grattafiori2024llama3}.
\item \fullcite{jiang2023mistral}.
\item \fullcite{openai2025gptoss}.
\item \fullcite{bai2025qwen25vl}.
\item \fullcite{zhang2025qwen3embedding}.
\item \fullcite{chen2024bgem3}.
\end{itemize}

\section*{Reproducibility and statistical evaluation}
\begin{itemize}[leftmargin=1.3em,itemsep=0.45em,topsep=0.3em]
\item \fullcite{pineau2021repro}.
\item \fullcite{miller2024errorbars}.
\item \fullcite{schaeffer2023mirage}.
\end{itemize}
\endgroup